\pdfoutput=1

\documentclass{article}

\usepackage{microtype}
\usepackage{graphicx}
\usepackage{subcaption}
\usepackage{booktabs}
\usepackage{hyperref}
\usepackage{multirow}
\usepackage[table]{xcolor}
\usepackage{pdflscape}

\usepackage[preprint]{icml2027}
\hypersetup{pdfsubject={}}

\usepackage{amsmath}
\usepackage{amssymb}
\usepackage{mathtools}
\usepackage{amsthm}
\usepackage[capitalize,noabbrev]{cleveref}
\usepackage{algorithm} 
\usepackage{algorithmic}
\usepackage[T1]{fontenc} 
\usepackage{xcolor}

\theoremstyle{plain}
\newtheorem{theorem}{Theorem}[section]
\newtheorem{proposition}[theorem]{Proposition}

\theoremstyle{definition}

\theoremstyle{remark}
\newtheorem{remark}[theorem]{Remark}

\icmltitlerunning{Any-Rate: Half-Bit Multi-Precision LLM Quantization with Coded Bit-Planes}

\begin{document}

\twocolumn[
\icmltitle{Any-Rate: Half-Bit Multi-Precision LLM Quantization with Coded Bit-Planes}

\icmlsetsymbol{equal}{*}

\begin{icmlauthorlist}
\icmlauthor{Hoonki Lee}{equal,add}
\icmlauthor{Haechan Shin}{equal,yonsei}
\icmlauthor{Hyuntaek Jung}{yonsei}
\icmlauthor{Yeonu Choi}{yonsei}
\icmlauthor{Sangyeob Kim}{yonsei}
\end{icmlauthorlist}

\icmlaffiliation{add}{Agency for Defense Development}
\icmlaffiliation{yonsei}{Yonsei University}

\icmlcorrespondingauthor{Sangyeob Kim}{sangyeob.kim@yonsei.ac.kr}

\icmlkeywords{Quantization, Large Language Models, Multi-Precision Quantization, On-Device Inference, Coded Bit-Planes}

\vskip 0.3in
]

\printAffiliationsAndNotice{\icmlEqualContribution}

\begin{abstract}
Nested multi-precision quantization derives multiple precisions from a single checkpoint by reading only the leading bits of each weight, which makes it attractive for on-device deployment under varying memory and energy budgets. However, existing methods offer only integer bits, leaving large accuracy gaps between adjacent precisions and severe degradation at 2 bits. To address this, we propose Any-Rate, which serves 2--8 bits from one checkpoint, in half-bit steps over 2--4 bits. Any-Rate restricts some bit-planes to extended Hamming codewords in 8-weight blocks (0.5 bits per weight), shares one LUT per row across precisions, and encodes all precisions jointly with a separate target for each, reducing the loss from sharing bit-planes. Across four LLMs, Any-Rate achieves state-of-the-art accuracy among multi-precision methods at 2 and 3 bits, keeping 69\% of FP16 accuracy at 2 bits, and is within one point of the best at 4 bits. It decodes about as fast as existing kernels, and an LPDDR energy model estimates up to 27\% less energy at equal accuracy. Code is available at \url{https://github.com/hoonki2002/Any-Rate}.
\end{abstract}

\section{Introduction}
\label{sec:intro}

Deploying large language models (LLMs) on devices requires fitting them into memory and energy budgets that differ across devices and can change at runtime~\citep{laskaridis2024melt}.
In practice, LLM inference is primarily memory-bound, so the bit width of the weights dominates latency and energy per token~\citep{lin2024awq}.
To balance the efficiency gains of lower bit widths against their accuracy loss,
runtime policies switch the precision during inference, across decoding steps~\citep{chen2025pmpd}, layers~\citep{kwon2025dpllm}, or experts~\citep{kim2026dpamoe}.

However, storing one model per precision level multiplies storage and requires reloading at every switch.
Nested multi-precision quantization instead stores a single checkpoint, from which each precision, called a \emph{mode}, reads only its first few bit-planes, as in Any-Precision LLM (AP)~\citep{park2024anyprecision}, AnyBCQ (AB)~\citep{park2026anybcq}, and MatGPTQ (MG)~\citep{kleinegger2026matgptq}.
Such checkpoints have two limits.
First, every bit-plane costs one bit per weight, so between 2 and 4 bits, where accuracy changes the most, only the 2-, 3-, and 4-bit modes are available.
Second, with error feedback, each mode compensates its own errors and moves toward a different target, but all modes must share their leading bit-planes, so each mode loses more accuracy than it would when quantized alone; we call this loss the \emph{nesting cost}.
MG optimizes all modes jointly, but its single averaged target only moves this cost between modes.
In practice, AP and AB degrade sharply at 2 bits, and MG loses accuracy at 4 bits once it includes a 2-bit mode.

We propose Any-Rate, which serves 2--8-bit modes from one checkpoint, in half-bit steps over 2--4 bits.
First, Any-Rate represents each bit-plane of a block of eight weights as a codeword of the extended Hamming $[8,4,4]$ code~\citep{conway1999splag}.
Such a coded (H) plane adds one index bit per weight but requires only 0.5 bits per weight.
One unconstrained (U) plane and six H planes (UHHHHHH) together support the 2-, 2.5-, 3-, 3.5-, and 4-bit modes,
which share a single lookup table (LUT), and residual planes add the 5--8-bit modes. 
Second, Any-Rate optimizes all modes jointly, like MG, but gives each mode its own error feedback and target, and selects the codewords of all H planes in a block jointly.
Third, we analyze the benefits of these two design choices (Section~\ref{sec:analysis}). Coded planes provide more choices for important columns, and we prove that they beat scalar planes at 2 and 3 bits even when all columns are equally important.
With error feedback, per-mode targets pay a smaller nesting cost than the averaged target of MG, and coded planes reduce it further.

Across four LLMs, Any-Rate achieves state-of-the-art average accuracy among multi-precision methods at 2 and 3 bits, keeping 69\% of FP16 accuracy at 2 bits, and is within one point of the best at 4 bits (Section~\ref{sec:main_results}).
Its 3.5-bit mode is already more accurate than AB and MG (2/3/4) at 4 bits.
It decodes about as fast as existing kernels, and an LPDDR energy model estimates up to 27\% less energy at the same accuracy (Section~\ref{sec:efficiency}).

\section{Background and Motivation}
\subsection{Nested Multi-Precision Quantization}
\label{subsec:nest_q}
To mitigate substantial memory and computational demands during large language model (LLM) inference, 
post-training quantization (PTQ) reduces the precision of pretrained model weights.
Scalar quantization operates on individual weights, ranging from naive round-to-nearest (RTN)~\citep{dettmers2022llmint8,yao2022zeroquant} to GPTQ~\citep{frantar2023gptq}, which incorporates layer-wise reconstruction errors using calibration data.
Beyond scalar quantization, structured quantization jointly encodes multiple weights, employing lattice-based codebooks (QuIP\#~\citep{tseng2024quipsharp}) or trellis coding (QTIP~\citep{tseng2024qtip}).

Nested multi-precision quantization targets multiple bit widths by sharing a common bit representation and reconstructs each mode by reading only the corresponding prefix of bit-planes.
Any-Precision LLM (AP)~\citep{park2024anyprecision} hierarchically refines quantization clusters 
while preserving bit prefixes and storing a lookup table (LUT) for each mode.
MatQuant~\citep{nair2025matryoshka} jointly optimizes multiple bit widths by sharing the most significant bits (MSBs) of quantized integer representations.
AnyBCQ (AB)~\citep{park2026anybcq} progressively expands precision by representing weights as sums of binary bases with corresponding scale factors.

Recently, MatGPTQ (MG)~\citep{kleinegger2026matgptq} extends GPTQ's Hessian-guided error feedback to nested multi-precision quantization by jointly minimizing a weighted sum of layer-wise reconstruction losses across multiple bit widths.
MG propagates the weighted average of the quantization errors of all modes to the remaining unquantized weights.

\subsection{Deployment-Driven Precision Requirements}
\label{subsec:fine_grained}
In practical deployment scenarios, varying accuracy and latency requirements motivate dynamic precision selection during LLM inference: across decoding steps~\citep{chen2025pmpd}, layers~\citep{kwon2025dpllm}, and experts~\citep{kim2026dpamoe}.
However, the effectiveness of these policies is constrained by the available modes and the accuracy of each mode.

In the 5--8-bit regime, quantized models approach full-precision performance, yielding diminishing accuracy gains despite the increased memory costs of higher bit widths. 
In contrast, accuracy degrades sharply in the 2--4-bit regime, where existing nested quantization methods either suffer severe degradation at 2 bits (AP, AB) or sacrifice higher-precision accuracy when jointly optimizing for the 2-bit mode (MG).
Since existing methods rely on scalar bit-planes, each requiring one bit per weight, they support only integer-bit increments, further limiting usable rates in the low-bit regime.

To address these limitations, we propose \textbf{Any-Rate (AR)}, which introduces block-wise Hamming-coded bit-planes that support half-bit steps across the 2--4-bit regime, while maintaining competitive accuracy even at 2 bits through joint encoding. Detailed performance results across bit-width regimes are provided in Appendix~\ref{app:full_results}.

\subsection{Optimization under Nesting Constraint}
\label{subsec:nest_cost}

While AP and AB progressively construct nested representations, MG aims to improve reconstruction quality across modes through Hessian-guided joint optimization.
Specifically, MG minimizes a weighted sum of layer-wise reconstruction losses and updates a single feedback-compensated target using the weighted average of quantization errors across modes.
However, this shared target may differ from the targets obtained through per-mode error feedback.
Even with per-mode error feedback, the shared-prefix structure imposes a \emph{nesting constraint}, restricting the quantization choices available to individual modes.
To quantify this penalty, we define the \emph{nesting cost} as the excess reconstruction loss relative to independently optimized quantization. To mitigate this cost, Any-Rate jointly optimizes reconstruction losses across modes while maintaining per-mode targets.
In Section~\ref{sec:analysis}, we analyze how error feedback contributes to nesting cost and how Any-Rate mitigates it through per-mode targets and coded bit-planes.

\subsection{Energy Constraints in On-Device Inference}
\label{subsec:energy}
Existing nested methods (AP, AB, and MG) evaluate inference efficiency through kernel latency and throughput, but not energy.
Meanwhile, for deployment on low-precision accelerators~\citep{lee2018unpu,sharify2018loom,sharma2018bitfusion,chen2025bitmod,kim2025slimllama},
particularly in battery-powered devices with limited energy budgets~\citep{laskaridis2024melt}, energy efficiency becomes a critical consideration. During small-batch autoregressive decoding, the linear layers of the transformer perform primarily memory-bound matrix-vector multiplications (GEMVs) with limited weight reuse~\citep{williams2009roofline,lin2024awq}.
Consequently, off-chip memory accesses contribute substantially to energy consumption~\citep{horowitz2014energy,moon2025trex}.

In this setting, fine-grained precision selection enables Any-Rate to reduce weight traffic at matched accuracy, while a single shared LUT limits metadata overhead across modes.
In Section~\ref{sec:efficiency}, we evaluate the resulting energy efficiency under an LPDDR-based energy model alongside GPU decoding performance.

\section{Method}
\label{sec:method}

\subsection{Fine-Grained Rates via Coded Bit-Planes}
\label{subsec:coded_planes}

\paragraph{Bit-Planes and Modes.}
For a layer weight matrix $W\in\mathbb{R}^{r\times c}$, its quantized representation assigns each weight a $b$-bit index $\iota$, where bits at the same position across indices form a \emph{bit-plane}.
Each bit-plane contributes one bit to the index, while its \emph{rate} measures the number of stored bits read per weight, excluding metadata.
We refer to each \emph{mode} by the total rate of the bit-planes used to reconstruct weights
(e.g., 2-bit or 3-bit mode) and denote its quantized weight by $\widehat w_m$, with $m$ indicating the mode.
Following Any-Precision LLM (AP)~\citep{park2024anyprecision}, $\widehat w^{(k)}$, the weight reconstructed from the first $k$ planes, is obtained from a row-wise lookup table (LUT):
\begin{equation}
\widehat w^{(k)}=T_k\left[\iota\gg(b-k)\right],
\label{eq:prefix_reconstruction}
\end{equation}
where $T_k$ is the row's $2^k$-entry LUT containing potentially non-uniform reconstruction values, so that $\widehat w_m=\widehat w^{(k)}$ for the mode that reads the first $k$ planes.

\paragraph{Coded Bit-Planes and Ladder.}
In conventional scalar bit-plane representations, adding an unconstrained (U) plane increases the rate by one bit per weight since each weight's index bit is stored independently.
To enable half-bit steps, we instead group eight adjacent weights in a row into a \emph{block} and constrain the bits of a plane within each block to a codeword of the extended Hamming $[8,4,4]$ code~\citep{conway1999splag}, and we denote such a plane by H (Hamming-coded).
Since each codeword can be specified by 4 bits per block, an H plane adds 1 index bit while increasing the rate by 0.5 bits per weight.
At the same 2-bit rate, UU provides 4 values per weight, whereas UHH provides 8 under block-wise coding constraints.
We define a \emph{layout} as a sequence of U and H planes and a \emph{ladder} as the collection of modes obtained from successive prefixes of the layout.
Any-Rate adopts the UHHHHHH layout, where reading the first 3 to 7 planes yields the 2, 2.5, 3, 3.5, and 4-bit modes, respectively.
We compare our chosen layout with the alternative configurations in Appendix~\ref{app:layout}.

\paragraph{A Single Shared LUT.}
Although coded bit-planes reduce the index storage rate, storing a separate LUT for each mode incurs additional metadata overhead.
For Llama-3.1-8B-Instruct, storing separate FP16 LUTs for the five modes spanning 2--4 bits would require about 0.8 bits per weight in metadata alone (Appendix~\ref{app:table}).
Instead, Any-Rate employs a single LUT $T$ with $2^b$ entries per row and derives each $T_k$ by averaging entries of $T$ that share the same $k$-bit prefix.

\paragraph{Residual Correction.}
In the 5--8-bit regime, extending the coded ladder with additional planes would incur substantial metadata overhead due to the exponential growth of the shared LUT.
Even with four additional U planes, the shared LUT alone would require about 3.2 bits per weight for Llama-3.1-8B-Instruct (Appendix~\ref{app:residual}).
We therefore fix the 2--4-bit ladder and encode the residual for the 5--8-bit modes using a separate 4-bit index $\rho$, consisting of four U planes (UUUU).
The $m$-bit mode reconstructs each weight by adding an offset to the 4-bit reconstruction:
\begin{equation}
\widehat w_{m}
=
\widehat w_{4}
+
\delta_m\left[\rho\gg(8-m)\right],
\quad m=5,\ldots,8,
\label{eq:residual_correction}
\end{equation}
where $\delta_m$ is a separate row-wise offset table for each mode, containing $2^{m-4}$ entries.

\subsection{Joint Encoding with Per-Mode Targets}
\label{sec:modejoint}

\paragraph{Objective and Per-Mode Targets.}
Let $\widehat W_m\in\mathbb{R}^{r\times c}$ denote the
weight matrix reconstructed at mode $m$, with quantization
error $E_m=W-\widehat W_m$.
Following GPTQ~\citep{frantar2023gptq}, we use calibration
inputs $X\in\mathbb{R}^{c\times n}$ and the input Gram
matrix $\Gamma=XX^\top$
to define the layer-wise reconstruction loss:
\begin{equation}
\mathcal{L}_m^{\mathrm{layer}}
=\|E_mX\|_F^2
=\mathrm{tr}(E_m\Gamma E_m^\top).
\label{eq:layer_loss}
\end{equation}

As in MatGPTQ (MG)~\citep{kleinegger2026matgptq}, we jointly minimize the error across modes, weighted by mode weights $\lambda_m$:
\begin{equation}
\min_{\iota,T}
\sum_{m}
\lambda_m\,\mathrm{tr}(E_m\Gamma E_m^\top).
\label{eq:joint_objective}
\end{equation}
During encoding, \emph{the error feedback} updates the targets of the remaining unquantized columns using the quantization error of each column and the Cholesky factor of $\Gamma^{-1}$.
Whereas MG updates a single shared target using the weighted average of errors between modes,
Any-Rate keeps a separate target $\tau_m$ for each mode, called \emph{per-mode targets}, and updates it with the errors of that mode.

\paragraph{Single-Mode Plane-Joint Encoding.}
\label{sec:planejoint}
Since U planes allow independent bit selection for each weight, GPTQ-style encoding selects quantization indices column by column with error feedback.
In contrast, weights cannot be encoded independently with H planes, as the eight bits of an H plane within each block are constrained to a single codeword.
Any-Rate therefore handles U and H planes differently, encoding U bits column by column while jointly selecting H-plane codewords block by block.
While greedily encoding H-plane codewords in prefix order can limit subsequent reconstruction choices, we instead consider their joint combinations as candidates $s=(s_1,\ldots,s_{n_{\mathrm{H}}})$, with one codeword from each H plane.
For each candidate, the H-plane bits are fixed and the U-plane bits are optimized with error feedback, and the combination with the lowest reconstruction loss is selected---a procedure we call \emph{plane-joint encoding}.

\paragraph{Multi-Mode Joint Encoding.}
Under the nesting constraint, we extend plane-joint encoding across modes by evaluating each candidate using the weighted loss in Eq.~\eqref{eq:joint_objective}.
For a candidate $s$, let $q_{m,i}(s,u_i)$ denote the reconstructed value at column $i$ for mode $m$ given the U-plane bit $u_i$, and let $\tilde{\tau}_{m,i}$ denote its feedback-compensated target, which includes the errors of the preceding columns. The shared U bit is selected by
\begin{equation}
u_i^*=\arg\min_{u_i\in\{0,1\}}
\sum_m\lambda_m
\left(\tilde{\tau}_{m,i}-q_{m,i}(s,u_i)\right)^2.
\label{eq:mode_joint}
\end{equation}
The resulting errors $e_{m,i}=(\tilde{\tau}_{m,i}-q_{m,i}(s,u_i^*))/R_{ii}$, where $R^\top R=\Gamma^{-1}$, are propagated separately for each mode.
Each candidate is then evaluated by its weighted loss $\sum_m\lambda_m\sum_i e_{m,i}^2$, and the lowest-loss candidate is selected.

\paragraph{Efficient Candidate Search.}
For the UHHHHHH layout used in Any-Rate, jointly selecting codewords across six H planes yields $16^6\approx1.7\times10^7$ candidates per block, making exhaustive evaluation with error feedback computationally expensive.
To reduce this cost, we exploit the prefix structure of the ladder, where each mode's reconstruction loss depends only on the planes it reads.
At each H plane, partial candidates are extended by its 16 codewords and ranked using the approximate weighted losses of the modes completed up to that plane, without within-block error feedback.
Only the best $B$ candidates are retained at each step.
After processing all H planes, the top $K$ candidates are evaluated with error feedback, and the lowest-loss candidate is selected and the errors of each mode are propagated to the remaining unquantized columns.
Algorithm~\ref{alg:block} summarizes the joint encoding procedure, where $\mathcal B$ denotes an eight-column block over which H-plane codewords are selected.
We use $B=K=64$, with further details in Appendix~\ref{app:planejoint}.

\begin{algorithm}[t]
\caption{Joint encoding of one 8-column block (rows in parallel).}
\label{alg:block}
\begin{algorithmic}[1]
\REQUIRE per-mode targets $\tau_m$, factor block $R_{\mathcal B,:}$,
mode weights $\lambda_m$, beam width $B$, candidates $K$
\STATE $\mathcal{S} \leftarrow \{\varnothing\}$
\FOR{each H plane $p = 1,\dots,n_{\mathrm{H}}$}
  \STATE extend $\mathcal{S}$ by the 16 codewords of plane $p$; keep the $B$ best by the cost (no feedback) of the modes complete up to plane $p$
\ENDFOR
\FOR{each of the $K$ best $s \in \mathcal{S}$}
  \STATE for $i = 1,\dots,8$: $\tilde\tau_{m,i} \leftarrow \tau_{m,i} - \sum_{l<i} e_{m,l}R_{li}$; choose the U-plane bit $u_i$ minimizing $\sum_m \lambda_m(\tilde\tau_{m,i} - q_{m,i})^2$; $e_{m,i} \leftarrow (\tilde\tau_{m,i} - q_{m,i})/R_{ii}$
  \STATE $\mathrm{cost}(s) \leftarrow \sum_m \lambda_m \sum_i e_{m,i}^2$
\ENDFOR
\STATE take $s^\ast = \arg\min_s \mathrm{cost}(s)$;
$\tau_{m,>\mathcal B} \leftarrow \tau_{m,>\mathcal B}
- \mathbf e_{m,\mathcal B}(s^\ast)R_{\mathcal B,>\mathcal B}$
\end{algorithmic}
\end{algorithm}

\paragraph{LUT and Residual Fitting.}
Since each mode LUT $T_k$ is linear in the shared LUT $T$, optimizing Eq.~\eqref{eq:joint_objective} with fixed indices reduces to a weighted least-squares problem with a closed-form solution (Appendix~\ref{app:table}).
Any-Rate therefore alternates between joint encoding of the indices and refitting $T$, retaining the solution with the lowest reconstruction loss (Appendix~\ref{app:algorithm}).
After fixing the 2--4-bit ladder, the residual indices $\rho$ are encoded column by column with per-mode error feedback, while the offset tables $\delta_m$ are refitted using the mean residuals of their assigned weights (Appendix~\ref{app:residual}).

\subsection{Decoding and Storage}
\label{sec:format}

\paragraph{Prefix Decoding.}
Any-Rate stores quantization indices plane by plane, allowing each mode to read only its required prefix.
For H planes, the index bits can be recovered directly from the stored 4-bit messages using XOR operations, without explicitly expanding the codewords (Appendix~\ref{app:coded}).
The mode LUT $T_k$ is derived from the shared LUT $T$ once when switching modes, requiring neither requantization nor a separate checkpoint.
For the 5--8-bit modes, decoding additionally reads the corresponding residual planes and applies the offsets in Eq.~\eqref{eq:residual_correction}.

\paragraph{Storage Format.}
A U plane packs eight weights into one byte, whereas an H plane stores a 4-bit message per eight-weight block, requiring 1 and 0.5 bits per weight, respectively.
The shared LUT $T$ is stored using 8-bit entries with an FP16 offset and step per row, requiring 0.21 bits per weight for UHHHHHH on Llama-3.1-8B-Instruct.
For the 5--8-bit modes, the four residual U planes are stored separately from the ladder, while the offset tables $\delta_m$ require only 30 FP16 entries per row, or 0.09 bits per weight.
The effective bits read per weight during decoding include both quantization codes and metadata, as detailed in Appendix~\ref{app:storage}.

\section{Analysis}
\label{sec:analysis}
We analyze the effects of Hamming-coded planes and per-mode targets. We first examine why coded planes improve single-mode quantization despite restricting admissible bit patterns, then investigate the gain from error feedback and the nesting cost across modes, and finally assess the net gain under nested quantization.
To quantify these effects, we normalize the layer-wise reconstruction loss in Eq.~\eqref{eq:layer_loss} by the squared norm of the original layer output, which we call the relative proxy loss:
\begin{equation}
L_m =
\frac{\mathcal{L}_m^{\mathrm{layer}}}{\|WX\|_F^2}
=
\frac{\mathrm{tr}(E_m\Gamma E_m^\top)}
{\mathrm{tr}(W\Gamma W^\top)}.
\label{eq:proxy_loss}
\end{equation}
We measure the gain of a method over its baseline in dB as $G=10\log_{10}(L_{\mathrm{base}}/L_{\mathrm{method}})$, and report mean gains over 246 linear-layer weight matrices of Qwen2.5-3B~\citep{qwen2025qwen25} (Appendix~\ref{app:an_setup}).

\subsection{Single-Mode Gain of the Coded Plane}
\label{sec:an_code}

\begin{figure}[t]
\centering
\includegraphics[width=\columnwidth]{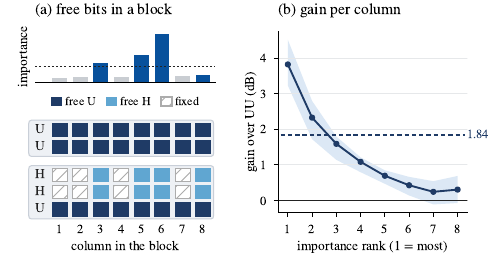}
\caption{Allocation gain of the coded plane (Qwen2.5-3B, 2-bit). (a) Free bits of UU and UHH in one block. (b) Gain of UHH over UU by in-block importance rank (mean; band: interquartile range); dashed: whole matrix.}
\label{fig:an_alloc}
\end{figure}

An H plane represents the 8 bits of a block with a 4-bit message, restricting the admissible bit patterns from $2^8=256$ to $2^4=16$.
While this constraint limits independent bit choices, structured coding schemes such as lattice and trellis codes can reduce distortion through higher-dimensional joint quantization, particularly at low rates where scalar quantization offers only a few reconstruction levels~\citep{tseng2024quipsharp, tseng2024qtip}. To examine the structural benefit of Hamming-coded planes, we compare coded layouts against Lloyd--Max scalar quantization on Gaussian inputs in the low-bit regime, and find that coded layouts can achieve lower distortion than scalar quantization (Appendix~\ref{app:an_proof}).

Coded planes can achieve additional gains by allocating representational freedom to more important columns. With error feedback, the reconstruction loss in Eq.~\eqref{eq:layer_loss} decomposes for each row as
\begin{equation}
\mathcal{L}^{\mathrm{row}} = \sum_j d_j \varepsilon_j^2,
\qquad
d_j = \Gamma_{jj} -
\Gamma_{j,>j}\Gamma_{>j,>j}^{-1}\Gamma_{>j,j},
\label{eq:importance}
\end{equation}
where $\varepsilon_j=\tilde\tau_j-\widehat w_j$ is the quantization error relative to the feedback-compensated target $\tilde\tau_j$, and $d_j$ denotes the importance of column $j$ (Appendix~\ref{app:an_decomp}). Thus, quantization errors contribute differently to the reconstruction loss depending on column importance, and coded planes can concentrate representational freedom on important columns.

Note that when $\Gamma=I$, error feedback vanishes and all columns have equal importance ($d_j=1$), isolating the structural gain. The following theorem shows that uniform importance yields the worst-case distortion for coded quantization relative to scalar quantization under exchangeable columns.

\begin{theorem}[Uniform importance is the worst case]
\label{thm:uniform}
Suppose the eight columns of a block are exchangeable,
the loss is weighted by a diagonal column importance
$d\in\mathbb{R}_{>0}^8$, and both structures use one LUT per row.
Let $D_{\mathrm{coded}}^*(d)$ and $D_{\mathrm{scalar}}^*(d)$ denote
the optimal weighted distortions of a block, and
let $D_{\mathrm{c}}$ and $D_{\mathrm{LM}}$ denote the
optimal per-column distortions of coded and Lloyd--Max
scalar quantization under uniform importance.
Then, for any $d$,
\begin{equation}
D_{\mathrm{coded}}^*(d)
\le D_{\mathrm{c}}\sum_j d_j,\; \; \; 
D_{\mathrm{scalar}}^*(d)
= D_{\mathrm{LM}}\sum_j d_j.
\end{equation}
Hence, if $D_{\mathrm{c}}<D_{\mathrm{LM}}$, the coded
structure achieves lower distortion for every $d$.
\end{theorem}

The proof follows from the concavity of the optimal weighted distortion and the transitive symmetry of the Hamming code, which together yield the bound via Jensen's inequality.
Although real matrices lie outside these assumptions, on 246 weight matrices of Qwen2.5-3B, coded structures achieve mean gains of 1.84, 1.69, and 1.31\,dB at 2, 3, and 4 bits, outperforming scalar structures on 246, 242, and 210 matrices, respectively.
At 2 bits per weight, UU assigns two independent index bits to each column, whereas UHH allows four suitable columns to independently select three index bits, while the remaining four select only their U bit. The measured gains are concentrated in columns with higher importance (Figure~\ref{fig:an_alloc}). The full proof and additional experimental results are provided in Appendix~\ref{app:an_proof}.

\subsection{Gain of Error Feedback and the Nesting Cost}
\label{sec:an_nest}

\begin{figure}[t]
\centering
\includegraphics[width=\columnwidth]{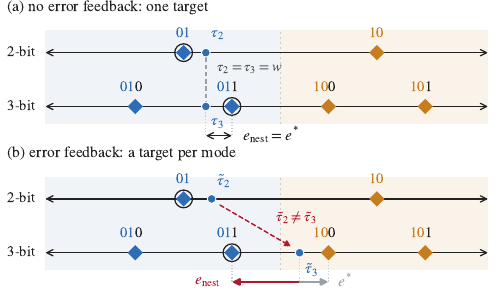}
\caption{Nesting cost on a 2/3-bit ladder (circles: chosen points; $e^*$ and $e_{\mathrm nest}$: reconstruction errors under independent quantization and nested quantization with per-mode error feedback, respectively). (a) Without error feedback, both modes share one target, so $e_{\mathrm{nest}}=e^*$. (b) With error feedback, the 3-bit target moves toward a child of another 2-bit point, so the nested choice has a larger error ($e_{\mathrm{nest}}>e^*$).}
\label{fig:nesting_cost}
\end{figure}

Error feedback reduces reconstruction loss by compensating for earlier quantization errors in subsequent columns. As shown in Eq.~\eqref{eq:importance}, this reduces the effective error weights from $\Gamma_{jj}$ to $d_j\le\Gamma_{jj}$, where the former assumes uncorrelated rounding errors without feedback.  Assuming equal per-column error variances, the feedback gain is approximated by
\begin{equation}
G^{\mathrm{FB}}\approx
10\log_{10}\frac{\sum_j\Gamma_{jj}}{\sum_jd_j}.
\label{eq:feedback_gain}
\end{equation}
Under this approximation, the feedback gain is independent of bit width, consistent with the measured gain of approximately 2.75\,dB across all three bit widths (Table~\ref{tab:an_nest}); the approximation overestimates the size but orders the matrices by their gain (Appendix~\ref{app:an_fbgain}).

\begin{table}[t]
\caption{Feedback gain and nesting cost (dB, mean over matrices). Scalar ladder: UUUU with mode weights 1-2-4. MG (2/3/4): MatGPTQ with its 2/3/4-bit recipe and mean propagation, against single-mode GPTQ; 1-1-1 is the baseline in our results.}
\label{tab:an_nest}
\centering
\small
\setlength{\tabcolsep}{4pt}
\begin{tabular}{lccc}
\toprule
 & 2-bit & 3-bit & 4-bit \\
\midrule
Feedback gain & 2.73 & 2.78 & 2.76 \\
\midrule
\multicolumn{4}{l}{Nesting cost} \\
\quad Scalar ladder, no feedback & 0.12 & 0.23 & $-0.10$ \\
\quad Scalar ladder, per-mode targets & 0.78 & 0.97 & 1.51 \\
\quad MG (2/3/4), weights 1-1-1 & 0.54 & 1.43 & 5.45 \\
\quad MG (2/3/4), weights 1-2-4 & 1.61 & 0.96 & 2.73 \\
\bottomrule
\end{tabular}
\end{table}

However, extending error feedback to nested modes introduces a cost, even with per-mode targets. Since these targets may favor incompatible reconstruction points under the nesting constraint, additional quantization error can arise (Figure~\ref{fig:nesting_cost}).
To compare shared and per-mode targets, we define the \emph{nesting cost} of mode $m$ as
\begin{equation}
c_m=10\log_{10}
\frac{L_m^{\mathrm{nest}}}{L_m^{\mathrm{single}}},
\label{eq:nesting_cost}
\end{equation}
where $L_m^{\mathrm{nest}}$ and $L_m^{\mathrm{single}}$ denote the losses of nested and single-mode encoding.
Any-Rate applies error feedback separately to each mode, whereas MatGPTQ (MG)~\citep{kleinegger2026matgptq} propagates the weighted mean of mode errors.
Let $\tilde{\tau}_{m,j}$ be the feedback-compensated target of mode $m$ at column $j$, and define its weighted mean as $\bar{\tau}_j = \sum_m \lambda_m \tilde{\tau}_{m,j}/\sum_m \lambda_m$.
\begin{proposition}[Cost of Mean Propagation]
\label{prop:meanprop}
(i) The loss under mean propagation upper-bounds that under per-mode optimization, given fixed targets and the same nesting constraint.
(ii) Assuming zero-mean quantization errors uncorrelated
with target offsets and with matching second moments,
the expected excess loss of mean propagation is
\begin{equation}
\Delta_j =
\frac{d_j}{\sum_m \lambda_m}
\sum_{m<m'} \lambda_m \lambda_{m'}
\mathbb{E}\left[
(\tilde{\tau}_{m,j}-\tilde{\tau}_{m',j})^2
\right].
\label{eq:meanprop_cost}
\end{equation}
\end{proposition}
Proposition~\ref{prop:meanprop} shows that the excess loss depends on the squared differences between per-mode targets. In the scalar ladder, replacing per-mode feedback with mean propagation increases the sum over modes of the relative excess loss $(L_m^{\mathrm{nest}}-L_m^{\mathrm{single}})/L_m^{\mathrm{single}}$ by a median factor of 1.41. The derivation and ablation results are detailed in Appendix~\ref{app:an_fbgain}.

Table~\ref{tab:an_nest} shows that the scalar ladder incurs negligible nesting cost without feedback, whereas its nesting cost increases with bit width under per-mode feedback.
MG exhibits a particularly large nesting cost at 4 bits under equal mode weights. Assigning larger weights to higher-precision modes reduces the 4-bit cost but increases the 2-bit cost, shifting the cost between modes rather than eliminating it.
Thus, the increasing nesting cost reduces the net benefit of error feedback at higher bit widths, despite its nearly constant single-mode gain.
Further analysis and experimental results are provided in Appendix~\ref{app:an_mg348}.

\subsection{Coded Planes Also Pay a Smaller Nesting Cost}
\label{sec:an_coded_nest}

Replacing the scalar ladder with the coded ladder affects both the single-mode gain and the nesting cost. We therefore decompose the resulting gain under nested encoding as
\begin{equation}
G_m^{\mathrm{nest}}
=
G_m^{\mathrm{single}}
+
\left(c_m^{\mathrm{scalar}}-c_m^{\mathrm{coded}}\right),
\label{eq:nested_gain}
\end{equation}
where $G_m^{\mathrm{single}}$ denotes the single-mode
gain from Section~\ref{sec:an_code}, and the second
term captures the reduction in nesting cost achieved
by the coded structure.
Table~\ref{tab:an_coded_gain} shows that, with the same per-mode targets and mode weights, the coded ladder further reduces the nesting cost, particularly at 2 and 4 bits, so at the same rate its gain over the scalar ladder is larger under nesting than in a single mode. While the single-mode gain accounts for most of the improvement at 2 bits, the reduction in nesting cost contributes more substantially at 4 bits.
Block-level analysis attributes this reduction primarily to smaller squared gaps between the per-mode targets (Appendix~\ref{app:an_codecost}), the quantity that also sets the cost of mean propagation in Proposition~\ref{prop:meanprop}, so the two design choices act on the nesting cost through the same quantity.
However, the coded ladder retains a structural nesting cost even without feedback, as its reconstruction cells are not perfectly nested (Appendix~\ref{app:an_mg348}).
The resulting gain remains positive across all three bit widths after accounting for LUT metadata, although it narrows at 4 bits due to the larger shared LUT.

\begin{table}[t]
\caption{Decomposition of the coded ladder's gain over
the scalar ladder (dB, mean over matrices), using the
same encoder and integer-mode weights. Matched total
bits include LUT metadata (scalar loss interpolated to the total bits of the coded ladder).}
\label{tab:an_coded_gain}
\centering
\small
\setlength{\tabcolsep}{4pt}
\begin{tabular}{lccc}
\toprule
 & 2-bit & 3-bit & 4-bit \\
\midrule
Single-mode gain & 1.84 & 1.69 & 1.31 \\
Nesting cost saved & 0.43 & 0.14 & 0.81 \\
\midrule
Nested gain, same rate & 2.26 & 1.82 & 2.12 \\
Nested gain, matched total bits & 2.19 & 1.51 & 0.64 \\
\bottomrule
\end{tabular}
\end{table}

\begin{figure*}[t]
\centering
\includegraphics[width=\textwidth]{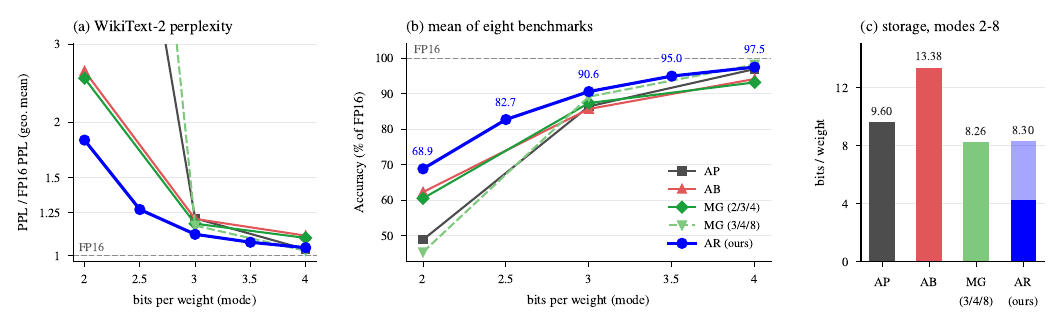}
\caption{Results averaged over the four models. (a) WikiText-2 PPL relative to FP16 (geometric mean). (b) Mean accuracy of the eight benchmarks relative to FP16. (c) Storage of one checkpoint serving all modes from 2 to 8 bits (Llama-3.1-8B-Instruct); AR splits into the 2--4-bit ladder (dark) and the 5--8-bit residual (light). Baselines support integer bits only.}
\label{fig:results}
\end{figure*}

\section{Experiments}
\label{sec:results}

\subsection{Setup}
\label{sec:setup}

We quantize all decoder linear layers of Qwen2.5-3B (base)~\citep{qwen2025qwen25}, Gemma-3-4b-it (language model only)~\citep{team2025gemma3}, Llama-3.1-8B-Instruct~\citep{grattafiori2024llama3}, and Phi-4 (14B)~\citep{abdin2024phi4}. We compare with Any-Precision LLM (AP)~\citep{park2024anyprecision}, AnyBCQ (AB, 2--4 bits only)~\citep{park2026anybcq}, and MatGPTQ (MG)~\citep{kleinegger2026matgptq} with its official 3/4/8-bit recipe, MG (3/4/8), and a 2/3/4-bit version, MG (2/3/4), all calibrated on C4~\citep{raffel2020c4}. We report WikiText-2~\citep{merity2017wikitext} PPL and the mean accuracy (Avg.) over eight benchmarks: ARC-Challenge and ARC-Easy~\citep{clark2018arc}, HellaSwag~\citep{zellers2019hellaswag}, PIQA~\citep{bisk2020piqa}, WinoGrande~\citep{sakaguchi2020winogrande}, GSM8K~\citep{cobbe2021gsm8k}, TriviaQA~\citep{joshi2017triviaqa}, and NQ-open~\citep{kwiatkowski2019nq}. Appendix~\ref{app:setup} gives the details.

\subsection{Main Results}
\label{sec:main_results}

Figure~\ref{fig:results} averages the four models (per-model values are in Appendix~\ref{app:full_results}). At 2 bits, Any-Rate reaches 69\% of the FP16 accuracy and 1.8 times the FP16 PPL, ahead of every baseline on all four models. The half-bit modes fill the gaps between integer bit widths: at 3.5 bits, Any-Rate reaches 95\% of the FP16 accuracy, higher than AB and MG (2/3/4) at 4 bits. At 3 bits, Any-Rate has the highest average accuracy, and at 4 bits it is 0.7 points below MG (3/4/8). MG (3/4/8), however, provides no usable 2-bit mode, and MG (2/3/4), which also targets the 2-bit mode, drops to 93\% at 4 bits, as the nesting cost of Section~\ref{sec:an_nest} predicts. At 6 and 8 bits, all methods, including Any-Rate, stay above 99.5\% of FP16. Serving every mode from 2 to 8 bits takes 8.30 bits per weight, close to MG (3/4/8) (8.26), and serving only 2--4 bits reads just the ladder, 4.21 bits per weight.

\subsection{Energy and Decoding Speed}
\label{sec:efficiency}

\begin{figure*}[h]
\centering
\includegraphics[width=\textwidth]{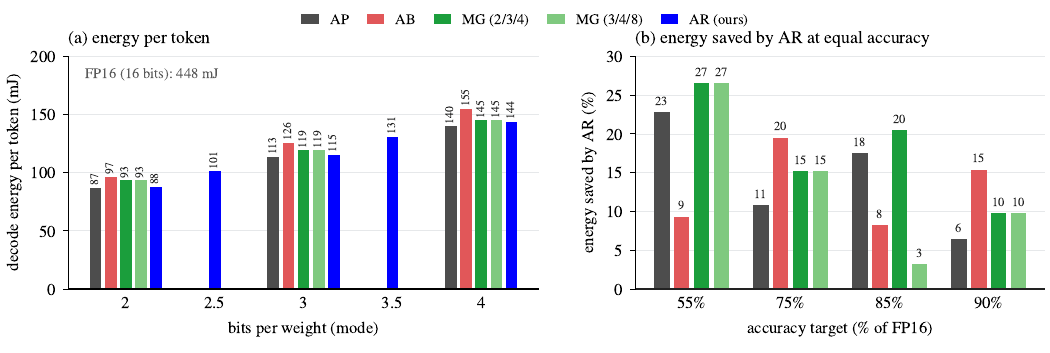}
\caption{Estimated decoding energy per token of Llama-3.1-8B-Instruct on an LPDDR-based accelerator. (a) Energy of each mode. (b) Energy AR saves at equal accuracy (mean of eight benchmarks); each method uses its cheapest mode that reaches the target.}
\label{fig:energy}
\end{figure*}

In on-device inference, most of the decoding energy goes to reading weights from external DRAM. We estimate the energy per token of Llama-3.1-8B-Instruct as the bits read per token times an LPDDR access energy of 3.7\,pJ/bit~\citep{moon2025trex}, plus the compute energy at 88.36\,TOPS/W~\citep{qin2025llmaccel} (Figure~\ref{fig:energy}). At the same bit width, the energy of Any-Rate is close to AP and equal to or lower than MG and AB: the LUT is read once per row and adds almost no bits per weight, and the compute, including table lookups, is below 1\% of the energy per token. The gap opens at equal accuracy. When each method uses its cheapest mode that reaches a target accuracy, Any-Rate needs 3--27\% less energy than the baselines to reach 55--90\% of the FP16 accuracy, because its 2-bit mode does not collapse, so low targets can be met at 2 bits, and its half-bit modes allow choosing the mode just above the target. Above 90\%, it is close to 4-bit AP (Appendix~\ref{app:efficiency}).

We measure the batch-1 throughput on an RTX 5070 Ti (Table~\ref{tab:tps}), replacing only the linear layers with the public kernel of each method and keeping the rest of the model the same. Any-Rate matches AP at 2 bits, stays within 3\% of AP at 3 and 4 bits, and is faster than AB and than MG with its \texttt{act\_order} permutation; the half-bit modes add one throughput point between each pair of integer bit widths. Appendix~\ref{app:efficiency} gives the linear-layer GEMV times and the measurement details.

\begin{table}[h]
\caption{Decoding throughput (tokens/s) of Llama-3.1-8B-Instruct, batch 1, RTX 5070 Ti (16\,GB). Higher is better. FP16 does not fit in memory.}
\label{tab:tps}
\centering
\small
\setlength{\tabcolsep}{4pt}
\begin{tabular}{lccccc}
\toprule
Method & 2 & 2.5 & 3 & 3.5 & 4 \\
\midrule
AP & \underline{141.6} & -- & \textbf{124.1} & -- & \textbf{107.6} \\
AB & 135.5 & -- & 116.0 & -- & 100.9 \\
MG (2/3/4) & 132.5 & -- & 119.3 & -- & \underline{106.9} \\
\quad + \texttt{act\_order} & 123.7 & -- & 112.1 & -- & 101.0 \\
\rowcolor[HTML]{DCEBFA} AR (ours) & \textbf{142.4} & 127.0 & \underline{121.3} & 112.7 & 104.9 \\
\bottomrule
\end{tabular}
\end{table}
\subsection{Ablation}
\label{sec:ablation}

To check whether the layer-loss results of Section~\ref{sec:analysis} carry over to the whole model, we change only the ladder or the target rule on Qwen2.5-3B and measure the PPL (Table~\ref{tab:ablation}). The scalar ladder UUUU collapses to a 2-bit PPL of 78 and is also 2.0 and 0.8 higher than Any-Rate at 3 and 4 bits. Coded planes provide both the single-mode gain of Section~\ref{sec:an_code} and the smaller nesting cost of Section~\ref{sec:an_coded_nest}, at the price of 0.33 more bits per weight for the larger LUT of each row. Encoding with MG-style mean propagation instead of per-mode targets raises the PPL in every mode, most at 2 bits (by 1.8). Supporting the half-bit modes costs little: nearly zeroing their weights improves the 2-bit PPL by only 0.14 while worsening the 3- and 4-bit modes by 0.18 and 0.05 and the half-bit modes by 1.1 and 0.4. Appendix~\ref{app:encoder} covers the choice of layout, mode weights, and beam width.

\begin{table}[h]
\caption{Ablation on Qwen2.5-3B, WikiText-2 PPL (FP16 8.03), same encoder and calibration. Scalar: ladder UUUU (weights 1-2-4); Mean prop.: MG-style mean propagation instead of per-mode targets; Half-bit off: half-bit mode weights nearly zero. b/w: bits per weight.}
\label{tab:ablation}
\centering
\small
\setlength{\tabcolsep}{4pt}
\begin{tabular}{lcccccc}
\toprule
 & b/w & 2 & 2.5 & 3 & 3.5 & 4 \\
\midrule
Scalar (UUUU) & 4.06 & 78.46 & -- & 11.32 & -- & 9.25 \\
Mean prop. & 4.39 & 17.22 & 10.95 & 9.37 & 8.81 & 8.56 \\
Half-bit off & 4.39 & \textbf{15.27} & 11.90 & 9.47 & 9.11 & 8.50 \\
\rowcolor[HTML]{DCEBFA} AR (ours) & 4.39 & 15.41 & \textbf{10.78} & \textbf{9.29} & \textbf{8.73} & \textbf{8.45} \\
\bottomrule
\end{tabular}
\end{table}

\section{Discussion}
\label{sec:discussion}

Existing nested methods tend to collapse at their lowest mode, which narrows the range of precisions that can actually be chosen; Any-Rate, by simply reading fewer planes, descends in half-bit steps down to 2 bits while keeping far higher accuracy than existing methods. The two mechanisms identified in the analysis, coded planes and per-mode targets, both keep modes readable by truncating a prefix. Both reduce the nesting cost: per-mode targets reduce it compared with the averaged target of MG (Section~\ref{sec:an_nest}), and coded planes, which also move bits to important columns without a codebook, reduce it further, mainly by narrowing the gaps between the per-mode targets (Section~\ref{sec:an_coded_nest}). The same procedure applies unchanged to MoE, vision-language, diffusion, speech recognition, and video generation models (Appendices~\ref{app:moe} and~\ref{app:beyond_lm}).

\section{Conclusion}
\label{sec:conclusion}

We proposed Any-Rate, which serves 2--8 bits from a single checkpoint and provides 2--4 bits in half-bit steps. Any-Rate consists of a coded ladder, in which some bit-planes are grouped with a Hamming code~\citep{conway1999splag}, and a joint encoding that keeps a separate target for each mode. It is more accurate than existing multi-precision methods at low bit widths, and our analysis traced this to the single-mode gain of coded planes and their smaller nesting cost. Our work has limitations. Apart from single-precision GPTQ (Appendix~\ref{app:static}), our comparison is limited to multi-precision methods, and our analysis relies on layer losses of a single model. At 4 bits, Any-Rate is slightly less accurate than the official recipe of MatGPTQ~\citep{kleinegger2026matgptq}. Its energy gains are estimates based on memory traffic, and measuring them on a dedicated accelerator is left for future work.

\clearpage
\section*{Impact Statement} 
This paper aims to reduce the cost of running neural networks on devices with limited memory and energy by serving several precisions from one checkpoint. This can broaden access to large models by running them on devices without a server, help privacy by keeping data on the device, and save both the storage of keeping one checkpoint per precision and the energy of decoding. On the other hand, output quality degrades in low-bit modes; in our experiments, responses sometimes contained wrong details, and generated images became blurred. All models and datasets used here are public.

\section*{AI Use Statement} 
Large language models (LLMs) were used in the preparation of this paper as follows. \emph{Formatting:} LLMs assisted in \LaTeX{} formatting. \emph{Retrieval:} LLMs helped locate related work, which the authors read and verified. \emph{Proofs:} the theorems and proofs were formulated by the authors; LLMs were used to check the derivations, and the authors verified every step. \emph{Code:} LLMs assisted in writing the experiment code and preparing the code for public release; the authors reviewed the code and verified every reported number against the raw experimental results.

\section*{Reproducibility Statement} 
All results in this paper can be reproduced with the code released at \url{https://github.com/hoonki2002/Any-Rate}, which contains the Any-Rate encoding and evaluation code, the scripts that run every baseline with its authors' public code, the analysis and kernel benchmarks, and a demo that switches the bit width during generation (Appendix~\ref{app:demo}). For every reported benchmark value, the release also includes the per-question model outputs and grading. The checkpoints encoded with Any-Rate are also released on Hugging Face and can be run directly in any mode without re-encoding.

\bibliography{references}
\bibliographystyle{icml2027}

\clearpage
\appendix
\onecolumn

\section{Full Benchmark Results} 
\label{app:full_results}

\begin{center}
\includegraphics[width=\textwidth,height=0.98\textheight,keepaspectratio]{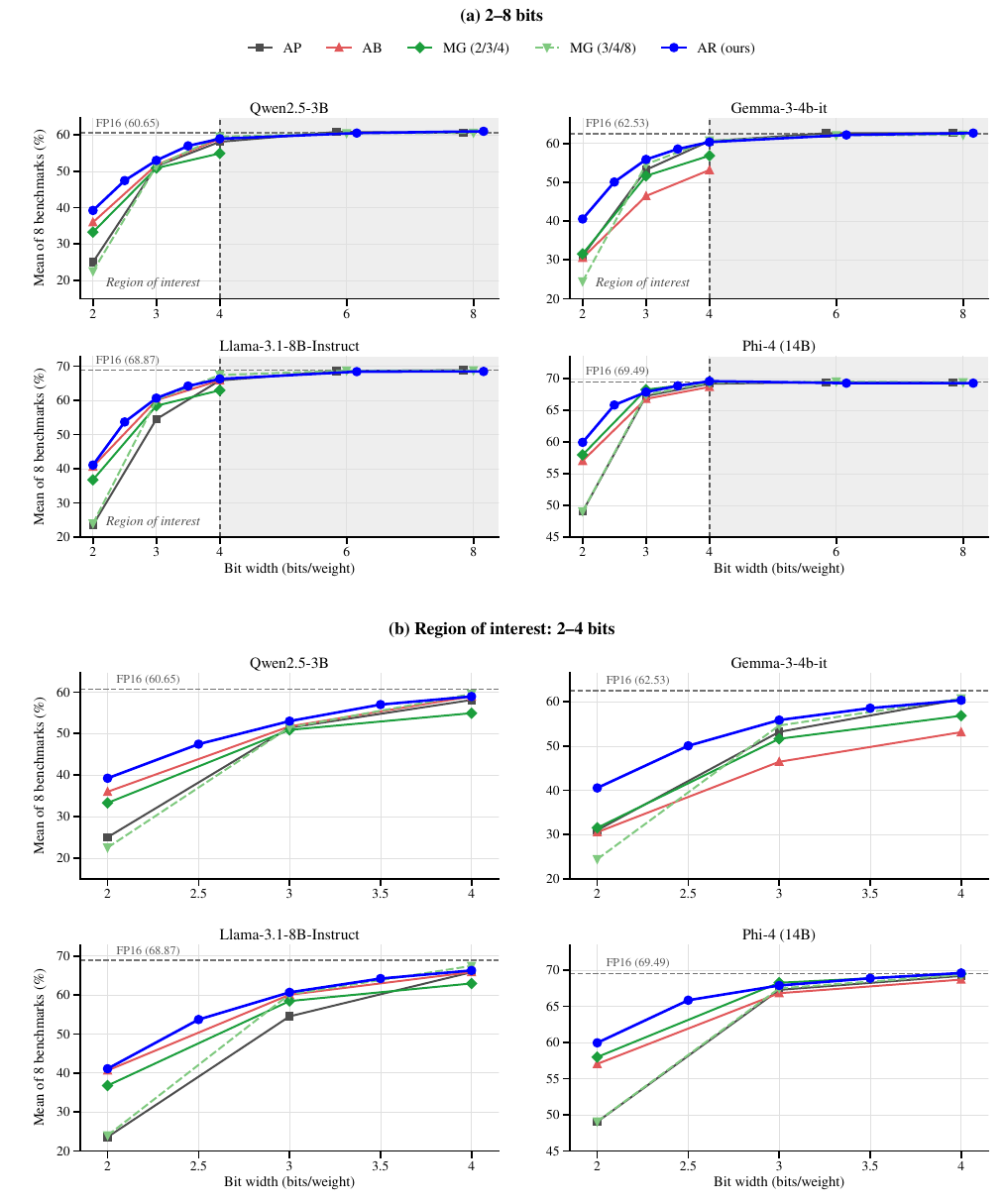}
\end{center}

\captionof{figure}{Mean accuracy over the eight benchmarks (\%) versus bit width for the four models: (a) 2--8 bits, (b) 2--4 bits. MG (3/4/8) is shown dashed.}
\label{fig:all4_bits}

\begin{table*}[p]
\centering
\caption{Qwen2.5-3B and Gemma-3-4b-it: WikiText-2 PPL and accuracy (\%) on eight benchmarks; Avg. is their mean. Best per bit width in \textbf{bold}, second best \underline{underlined}. MG (2/3/4): trained jointly for 2--4 bits; MG (3/4/8): official recipe.}
\label{tab:full_a}
\footnotesize
\setlength{\tabcolsep}{3.0pt}
\renewcommand{\arraystretch}{1.0}
\begin{tabular}{clc|cccccccc|>{\columncolor[gray]{0.92}}c}
\toprule
Bits & Method & PPL & ARC-c & ARC-e & HellaSwag & PIQA & WinoGrande & GSM8K & TriviaQA & NQ-open & Avg. \\
\midrule
\rowcolor[gray]{0.82} \multicolumn{12}{c}{\textbf{Qwen2.5-3B}} \\
\midrule
\multirow{5}{*}{2} & AP & $>$1000 & 22.01 & 37.84 & 29.97 & 57.24 & 50.99 & 1.97 & 0.20 & 0.00 & 25.03 \\
 & AB & \underline{21.88} & \underline{30.89} & \underline{56.73} & \underline{52.14} & 69.42 & \underline{59.27} & \underline{2.73} & \underline{14.10} & \underline{2.60} & \underline{35.99} \\
 & MG (2/3/4) & 25.89 & 30.80 & 46.59 & 52.07 & \underline{69.48} & 56.59 & 1.59 & 7.80 & 1.50 & 33.30 \\
 & MG (3/4/8) & $>$1000 & 23.89 & 26.89 & 26.93 & 50.38 & 52.01 & 0.15 & 0.00 & 0.00 & 22.53 \\
\rowcolor[HTML]{DCEBFA}  & AR (ours) & \textbf{15.41} & \textbf{34.04} & \textbf{58.46} & \textbf{58.82} & \textbf{73.29} & \textbf{62.43} & \textbf{3.87} & \textbf{18.00} & \textbf{5.10} & \textbf{39.25} \\
\midrule
\rowcolor[HTML]{DCEBFA} 2.5 & AR (ours) & \textbf{10.78} & \textbf{40.61} & \textbf{66.29} & \textbf{66.32} & \textbf{74.92} & \textbf{66.14} & \textbf{24.56} & \textbf{30.80} & \textbf{10.00} & \textbf{47.46} \\
\midrule
\multirow{5}{*}{3} & AP & 10.75 & \textbf{44.71} & 71.34 & 68.48 & \underline{76.22} & 66.46 & 40.86 & 34.10 & 9.30 & 51.43 \\
 & AB & \underline{9.95} & \underline{44.11} & \underline{73.78} & 67.88 & \textbf{76.82} & 65.59 & 38.29 & \underline{36.50} & 10.90 & \underline{51.73} \\
 & MG (2/3/4) & 10.75 & \textbf{44.71} & \textbf{74.03} & 67.17 & \underline{76.22} & \underline{67.48} & 35.25 & 32.10 & 10.20 & 50.90 \\
 & MG (3/4/8) & 10.19 & 42.41 & 64.69 & \underline{69.20} & 76.06 & 64.72 & \underline{46.47} & 36.20 & \underline{12.20} & 51.49 \\
\rowcolor[HTML]{DCEBFA}  & AR (ours) & \textbf{9.29} & 42.06 & 66.46 & \textbf{69.60} & 76.17 & \textbf{68.59} & \textbf{49.58} & \textbf{39.00} & \textbf{12.60} & \textbf{53.01} \\
\midrule
\rowcolor[HTML]{DCEBFA} 3.5 & AR (ours) & \textbf{8.73} & \textbf{44.45} & \textbf{72.77} & \textbf{70.80} & \textbf{76.61} & \textbf{69.46} & \textbf{62.02} & \textbf{44.50} & \textbf{15.40} & \textbf{57.00} \\
\midrule
\multirow{5}{*}{4} & AP & \underline{8.52} & 45.56 & 70.08 & \underline{72.08} & \underline{78.62} & 68.35 & 67.40 & 46.30 & \underline{16.50} & 58.11 \\
 & AB & 8.71 & \textbf{48.72} & \textbf{76.43} & 71.07 & 77.86 & 68.35 & 66.87 & \underline{46.70} & 15.80 & \underline{58.97} \\
 & MG (2/3/4) & 9.58 & 46.50 & 74.71 & 68.97 & 77.15 & \underline{68.51} & 52.62 & 39.50 & 11.50 & 54.93 \\
 & MG (3/4/8) & \textbf{8.45} & \underline{48.04} & \underline{75.13} & \textbf{72.53} & \textbf{78.73} & 67.64 & \textbf{70.05} & 46.60 & \textbf{17.60} & \textbf{59.54} \\
\rowcolor[HTML]{DCEBFA}  & AR (ours) & \textbf{8.45} & 46.33 & \underline{75.13} & 71.73 & 77.75 & \textbf{68.98} & \underline{69.07} & \textbf{46.80} & 15.80 & 58.95 \\
\midrule
\multirow{3}{*}{6} & AP & \textbf{8.06} & \underline{47.53} & 73.78 & \textbf{73.52} & 78.67 & \textbf{68.19} & \textbf{76.42} & \textbf{50.40} & \textbf{18.20} & \textbf{60.84} \\
 & MG (3/4/8) & \underline{8.13} & 46.93 & \underline{73.82} & 72.85 & \underline{78.84} & \underline{68.03} & \underline{75.13} & \underline{49.90} & \underline{18.00} & 60.44 \\
\rowcolor[HTML]{DCEBFA}  & AR (ours) & 8.19 & \textbf{47.70} & \textbf{75.63} & \underline{73.22} & \textbf{78.89} & \textbf{68.19} & 73.39 & 49.50 & 17.60 & \underline{60.52} \\
\midrule
\multirow{3}{*}{8} & AP & \textbf{8.03} & \underline{47.18} & 72.77 & \textbf{73.58} & 78.84 & \textbf{68.43} & \underline{75.36} & \textbf{50.10} & \underline{18.30} & \underline{60.57} \\
 & MG (3/4/8) & 8.11 & \underline{47.18} & \underline{74.16} & 72.96 & \underline{78.89} & \underline{67.48} & 75.21 & \underline{50.00} & \textbf{18.70} & \underline{60.57} \\
\rowcolor[HTML]{DCEBFA}  & AR (ours) & \underline{8.07} & \textbf{47.70} & \textbf{75.13} & \underline{73.43} & \textbf{79.00} & \textbf{68.43} & \textbf{75.66} & \underline{50.00} & \textbf{18.70} & \textbf{61.01} \\
\midrule
16 & FP16 & 8.03 & 47.27 & 73.11 & 73.55 & 78.89 & 68.35 & 75.51 & 50.30 & 18.20 & 60.65 \\
\midrule
\rowcolor[gray]{0.82} \multicolumn{12}{c}{\textbf{Gemma-3-4b-it}} \\
\midrule
\multirow{5}{*}{2} & AP & 145.90 & \underline{28.75} & \underline{52.06} & 42.54 & 64.36 & \underline{54.93} & 1.67 & 2.90 & 0.70 & 30.99 \\
 & AB & 69.26 & 26.79 & 44.44 & 42.01 & 64.96 & 54.22 & \underline{1.82} & 8.20 & 1.90 & 30.54 \\
 & MG (2/3/4) & \underline{41.35} & 27.22 & 45.92 & \underline{46.64} & \underline{66.16} & 53.35 & 1.67 & \underline{8.60} & \underline{2.70} & \underline{31.53} \\
 & MG (3/4/8) & $>$1000 & 24.23 & 33.04 & 30.94 & 56.64 & 49.17 & 1.52 & 0.00 & 0.00 & 24.44 \\
\rowcolor[HTML]{DCEBFA}  & AR (ours) & \textbf{27.04} & \textbf{37.88} & \textbf{64.10} & \textbf{57.33} & \textbf{70.89} & \textbf{59.04} & \textbf{8.11} & \textbf{20.70} & \textbf{6.20} & \textbf{40.53} \\
\midrule
\rowcolor[HTML]{DCEBFA} 2.5 & AR (ours) & \textbf{18.39} & \textbf{45.73} & \textbf{72.52} & \textbf{64.43} & \textbf{75.73} & \textbf{64.09} & \textbf{33.89} & \textbf{32.40} & \textbf{11.60} & \textbf{50.05} \\
\midrule
\multirow{5}{*}{3} & AP & 17.46 & \underline{50.00} & \underline{74.58} & \underline{69.25} & 75.52 & 63.22 & 44.43 & 36.70 & 11.80 & 53.19 \\
 & AB & 18.90 & 41.64 & 64.81 & 63.50 & 74.97 & 62.67 & 19.48 & 33.20 & 11.20 & 46.44 \\
 & MG (2/3/4) & \textbf{15.47} & 46.33 & 71.68 & 67.31 & \textbf{76.55} & \textbf{66.14} & 35.71 & 37.00 & 12.30 & 51.63 \\
 & MG (3/4/8) & 18.10 & 47.78 & 73.61 & 69.16 & 76.17 & 64.33 & \underline{52.08} & \textbf{41.10} & \underline{12.80} & \underline{54.63} \\
\rowcolor[HTML]{DCEBFA}  & AR (ours) & \underline{16.21} & \textbf{52.13} & \textbf{74.87} & \textbf{69.69} & \underline{76.44} & \underline{65.82} & \textbf{52.92} & \underline{40.60} & \textbf{14.40} & \textbf{55.86} \\
\midrule
\rowcolor[HTML]{DCEBFA} 3.5 & AR (ours) & \textbf{16.48} & \textbf{54.10} & \textbf{76.26} & \textbf{71.13} & \textbf{77.97} & \textbf{66.85} & \textbf{62.70} & \textbf{43.20} & \textbf{16.20} & \textbf{58.55} \\
\midrule
\multirow{5}{*}{4} & AP & 15.90 & \textbf{56.23} & \textbf{78.75} & \textbf{73.59} & 77.31 & 66.30 & \underline{71.19} & 45.70 & 15.60 & \underline{60.58} \\
 & AB & 18.34 & 48.63 & 70.24 & 68.51 & 76.71 & 64.25 & 42.76 & 40.40 & 13.50 & 53.13 \\
 & MG (2/3/4) & \textbf{15.01} & 51.02 & 75.72 & 70.17 & \underline{78.07} & \underline{67.48} & 52.54 & 45.20 & 14.40 & 56.83 \\
 & MG (3/4/8) & \underline{15.85} & \underline{55.03} & 75.29 & \underline{73.12} & 77.20 & 67.32 & \textbf{72.18} & \textbf{48.20} & \underline{17.50} & \textbf{60.73} \\
\rowcolor[HTML]{DCEBFA}  & AR (ours) & 16.16 & \underline{55.03} & \underline{76.85} & 71.94 & \textbf{78.13} & \textbf{68.75} & 68.23 & \underline{46.00} & \textbf{17.80} & 60.34 \\
\midrule
\multirow{3}{*}{6} & AP & \underline{16.08} & \textbf{57.42} & \underline{78.16} & \underline{73.96} & 76.99 & \textbf{69.38} & \textbf{77.10} & \underline{49.50} & \underline{18.40} & \textbf{62.61} \\
 & MG (3/4/8) & \textbf{14.88} & 55.20 & 77.44 & \textbf{74.21} & \underline{77.26} & 69.14 & 75.06 & \textbf{50.00} & \textbf{18.60} & 62.11 \\
\rowcolor[HTML]{DCEBFA}  & AR (ours) & 16.22 & \underline{56.06} & \textbf{78.28} & 73.67 & \textbf{78.07} & \underline{69.30} & \underline{75.13} & 48.80 & 18.10 & \underline{62.18} \\
\midrule
\multirow{3}{*}{8} & AP & \underline{16.05} & \textbf{57.00} & 77.90 & \underline{74.15} & 77.31 & \textbf{69.93} & \textbf{77.79} & \underline{49.10} & \underline{18.20} & \textbf{62.67} \\
 & MG (3/4/8) & \textbf{15.29} & \underline{55.72} & \underline{78.03} & \textbf{74.28} & \textbf{77.75} & 68.67 & 75.21 & \textbf{49.60} & \underline{18.20} & 62.18 \\
\rowcolor[HTML]{DCEBFA}  & AR (ours) & 16.17 & \textbf{57.00} & \textbf{78.37} & 74.12 & \underline{77.37} & \underline{69.46} & \underline{76.57} & \textbf{49.60} & \textbf{18.70} & \underline{62.65} \\
\midrule
16 & FP16 & 16.07 & 57.08 & 77.95 & 74.13 & 76.99 & 69.38 & 77.41 & 49.00 & 18.30 & 62.53 \\
\bottomrule
\end{tabular}
\end{table*}

\begin{table*}[p]
\centering
\caption{Llama-3.1-8B-Instruct and Phi-4 (14B): WikiText-2 PPL and accuracy (\%) on eight benchmarks; Avg. is their mean. Best per bit width in \textbf{bold}, second best \underline{underlined}. MG (2/3/4): trained jointly for 2--4 bits; MG (3/4/8): official recipe.}
\label{tab:full_b}
\footnotesize
\setlength{\tabcolsep}{3.0pt}
\renewcommand{\arraystretch}{1.0}
\begin{tabular}{clc|cccccccc|>{\columncolor[gray]{0.92}}c}
\toprule
Bits & Method & PPL & ARC-c & ARC-e & HellaSwag & PIQA & WinoGrande & GSM8K & TriviaQA & NQ-open & Avg. \\
\midrule
\rowcolor[gray]{0.82} \multicolumn{12}{c}{\textbf{Llama-3.1-8B-Instruct}} \\
\midrule
\multirow{5}{*}{2} & AP & $>$1000 & 24.40 & 31.06 & 27.33 & 54.73 & 49.49 & 0.76 & 0.60 & 0.10 & 23.56 \\
 & AB & \underline{18.95} & \textbf{37.80} & \underline{61.78} & \textbf{62.45} & \textbf{72.58} & 59.12 & \underline{3.41} & \textbf{21.20} & \textbf{6.70} & \underline{40.63} \\
 & MG (2/3/4) & 23.71 & 30.97 & 52.69 & 56.35 & 68.28 & \underline{60.62} & 3.34 & 18.30 & 3.50 & 36.76 \\
 & MG (3/4/8) & $>$1000 & 24.32 & 31.73 & 30.28 & 54.68 & 49.64 & 0.76 & 0.30 & 0.00 & 23.96 \\
\rowcolor[HTML]{DCEBFA}  & AR (ours) & \textbf{18.00} & \underline{37.12} & \textbf{62.08} & \underline{61.44} & \underline{72.47} & \textbf{63.14} & \textbf{6.07} & \underline{20.50} & \underline{5.80} & \textbf{41.08} \\
\midrule
\rowcolor[HTML]{DCEBFA} 2.5 & AR (ours) & \textbf{10.53} & \textbf{45.05} & \textbf{71.59} & \textbf{71.37} & \textbf{77.37} & \textbf{70.72} & \textbf{37.00} & \textbf{42.20} & \textbf{14.30} & \textbf{53.70} \\
\midrule
\multirow{5}{*}{3} & AP & 9.70 & 45.90 & 68.86 & 71.46 & 76.66 & 65.51 & 33.89 & 55.90 & 17.90 & 54.51 \\
 & AB & 8.99 & \textbf{51.45} & \underline{76.77} & \underline{74.66} & \underline{78.62} & 69.30 & \underline{51.02} & \underline{57.20} & \textbf{21.40} & 60.05 \\
 & MG (2/3/4) & 9.81 & \underline{50.68} & \textbf{77.19} & 74.08 & 78.51 & \underline{71.43} & 46.10 & 52.30 & 17.20 & 58.44 \\
 & MG (3/4/8) & \textbf{8.72} & 50.51 & 76.30 & \textbf{75.46} & 77.69 & \textbf{72.85} & 49.36 & \textbf{58.30} & \underline{20.90} & \underline{60.17} \\
\rowcolor[HTML]{DCEBFA}  & AR (ours) & \underline{8.91} & 50.26 & 74.87 & 74.36 & \textbf{79.11} & \textbf{72.85} & \textbf{58.00} & 55.30 & 20.60 & \textbf{60.67} \\
\midrule
\rowcolor[HTML]{DCEBFA} 3.5 & AR (ours) & \textbf{8.18} & \textbf{51.62} & \textbf{78.24} & \textbf{76.30} & \textbf{80.20} & \textbf{73.01} & \textbf{66.03} & \textbf{64.20} & \textbf{24.10} & \textbf{64.21} \\
\midrule
\multirow{5}{*}{4} & AP & \underline{7.62} & 53.75 & 76.94 & \underline{78.13} & 79.33 & 72.53 & 68.84 & \underline{70.40} & 26.80 & 65.84 \\
 & AB & 7.78 & \textbf{54.44} & 79.55 & 77.38 & \underline{80.09} & 72.30 & 67.85 & 68.20 & \underline{27.20} & 65.88 \\
 & MG (2/3/4) & 8.77 & 52.39 & \underline{79.84} & 75.78 & 80.03 & 71.43 & 59.74 & 61.30 & 23.20 & 62.96 \\
 & MG (3/4/8) & \textbf{7.55} & \underline{54.10} & \textbf{80.51} & \textbf{78.37} & \textbf{80.90} & \underline{73.24} & \textbf{72.93} & \textbf{70.90} & \textbf{28.60} & \textbf{67.45} \\
\rowcolor[HTML]{DCEBFA}  & AR (ours) & 7.82 & 53.75 & 79.50 & 77.44 & 80.03 & \textbf{74.11} & \underline{70.66} & 69.20 & 25.60 & \underline{66.29} \\
\midrule
\multirow{3}{*}{6} & AP & \textbf{7.24} & \underline{55.12} & \textbf{79.88} & \textbf{79.22} & 80.74 & \textbf{74.27} & 75.21 & \textbf{74.50} & 29.80 & \underline{68.59} \\
 & MG (3/4/8) & \underline{7.31} & \textbf{55.63} & \underline{79.59} & 78.76 & \textbf{81.23} & \underline{73.88} & \underline{75.97} & \underline{74.00} & \textbf{30.60} & \textbf{68.71} \\
\rowcolor[HTML]{DCEBFA}  & AR (ours) & 7.37 & 54.78 & 79.42 & \underline{78.85} & \underline{80.90} & 73.80 & \textbf{76.72} & 73.00 & \underline{30.00} & 68.43 \\
\midrule
\multirow{3}{*}{8} & AP & \textbf{7.22} & \underline{55.03} & \underline{79.63} & \textbf{79.27} & 81.12 & 73.88 & \textbf{76.42} & \textbf{75.40} & 30.10 & \textbf{68.86} \\
 & MG (3/4/8) & 7.30 & \textbf{56.40} & \textbf{79.67} & 78.81 & \textbf{81.28} & \underline{74.27} & \underline{75.59} & 73.90 & \textbf{30.50} & \underline{68.80} \\
\rowcolor[HTML]{DCEBFA}  & AR (ours) & \underline{7.26} & 54.52 & 78.62 & \underline{79.23} & \underline{81.23} & \textbf{74.35} & 75.13 & \underline{74.50} & \underline{30.30} & 68.49 \\
\midrule
16 & FP16 & 7.21 & 55.20 & 79.59 & 79.26 & 80.96 & 73.88 & 76.50 & 75.30 & 30.30 & 68.87 \\
\midrule
\rowcolor[gray]{0.82} \multicolumn{12}{c}{\textbf{Phi-4 (14B)}} \\
\midrule
\multirow{5}{*}{2} & AP & 13.70 & 45.90 & 68.01 & 66.94 & 76.06 & 65.51 & 26.84 & 33.50 & 9.40 & 49.02 \\
 & AB & 9.69 & 49.83 & 71.93 & \underline{73.00} & \textbf{79.33} & \textbf{72.30} & 53.75 & 41.60 & \underline{14.40} & 57.02 \\
 & MG (2/3/4) & \underline{9.43} & \underline{51.19} & \underline{72.98} & 72.95 & 78.35 & 68.03 & \underline{61.03} & \underline{45.30} & 13.90 & \underline{57.97} \\
 & MG (3/4/8) & 13.82 & 45.14 & 68.43 & 63.72 & 74.21 & 63.38 & 38.67 & 30.40 & 8.50 & 49.06 \\
\rowcolor[HTML]{DCEBFA}  & AR (ours) & \textbf{8.83} & \textbf{51.28} & \textbf{74.92} & \textbf{74.48} & \underline{79.00} & \underline{71.03} & \textbf{64.14} & \textbf{47.60} & \textbf{17.10} & \textbf{59.94} \\
\midrule
\rowcolor[HTML]{DCEBFA} 2.5 & AR (ours) & \textbf{7.50} & \textbf{55.89} & \textbf{76.60} & \textbf{78.40} & \textbf{80.36} & \textbf{74.82} & \textbf{79.61} & \textbf{58.20} & \textbf{22.80} & \textbf{65.83} \\
\midrule
\multirow{5}{*}{3} & AP & 7.16 & 54.69 & 73.78 & \textbf{80.40} & 80.96 & 74.27 & \underline{86.28} & \underline{63.70} & 24.00 & 67.26 \\
 & AB & 7.62 & 54.10 & 72.73 & 79.83 & 80.36 & 74.19 & 85.82 & 62.60 & 24.80 & 66.80 \\
 & MG (2/3/4) & 7.15 & \textbf{57.17} & \textbf{78.07} & 79.07 & \textbf{81.50} & \underline{75.53} & \underline{86.28} & 63.40 & 25.00 & \textbf{68.25} \\
 & MG (3/4/8) & \underline{6.98} & 55.03 & 75.17 & \underline{79.97} & 80.90 & 72.06 & \textbf{87.04} & 63.30 & \textbf{26.30} & 67.47 \\
\rowcolor[HTML]{DCEBFA}  & AR (ours) & \textbf{6.97} & \underline{55.29} & \underline{75.51} & 79.80 & \underline{81.07} & \textbf{75.77} & 85.60 & \textbf{64.20} & \underline{25.70} & \underline{67.87} \\
\midrule
\rowcolor[HTML]{DCEBFA} 3.5 & AR (ours) & \textbf{6.74} & \textbf{55.55} & \textbf{75.46} & \textbf{80.55} & \textbf{81.39} & \textbf{75.69} & \textbf{88.55} & \textbf{66.40} & \textbf{27.30} & \textbf{68.86} \\
\midrule
\multirow{5}{*}{4} & AP & \underline{6.63} & \underline{56.74} & 73.40 & \textbf{81.52} & 80.63 & 75.53 & \textbf{89.92} & \textbf{68.30} & 27.50 & 69.19 \\
 & AB & 7.32 & 54.95 & 73.99 & 81.22 & 80.52 & 74.59 & \underline{89.39} & 66.00 & \underline{28.80} & 68.68 \\
 & MG (2/3/4) & 6.88 & \textbf{59.13} & \textbf{78.79} & 80.16 & \textbf{82.48} & \underline{76.64} & 87.11 & 64.80 & 26.70 & \underline{69.48} \\
 & MG (3/4/8) & \textbf{6.56} & 56.40 & 74.37 & \underline{81.40} & 81.39 & 75.06 & 89.16 & \textbf{68.30} & 28.70 & 69.35 \\
\rowcolor[HTML]{DCEBFA}  & AR (ours) & \underline{6.63} & 56.14 & \underline{75.00} & 80.90 & \underline{81.94} & \textbf{76.72} & 89.16 & \underline{67.70} & \textbf{29.20} & \textbf{69.59} \\
\midrule
\multirow{3}{*}{6} & AP & \textbf{6.47} & \underline{55.80} & 72.73 & \textbf{81.98} & \underline{81.23} & \textbf{76.80} & \underline{89.92} & \underline{68.50} & \underline{28.50} & \underline{69.43} \\
 & MG (3/4/8) & \textbf{6.47} & \textbf{56.06} & \underline{73.11} & 81.64 & \textbf{81.34} & \textbf{76.80} & \textbf{90.30} & \textbf{69.20} & 27.70 & \textbf{69.52} \\
\rowcolor[HTML]{DCEBFA}  & AR (ours) & \underline{6.49} & 55.63 & \textbf{73.65} & \underline{81.84} & 80.90 & \underline{76.24} & 89.23 & 67.50 & \textbf{29.20} & 69.28 \\
\midrule
\multirow{3}{*}{8} & AP & \textbf{6.46} & \underline{55.89} & 72.81 & \textbf{81.97} & \underline{81.07} & \textbf{76.48} & \underline{89.69} & \underline{68.90} & \underline{28.20} & \underline{69.37} \\
 & MG (3/4/8) & \textbf{6.46} & \textbf{56.23} & \underline{73.06} & 81.68 & 81.01 & \textbf{76.48} & \textbf{89.84} & \textbf{69.10} & 27.80 & \textbf{69.40} \\
\rowcolor[HTML]{DCEBFA}  & AR (ours) & \textbf{6.46} & 55.38 & \textbf{73.15} & \underline{81.81} & \textbf{81.18} & \underline{76.32} & 89.54 & 68.20 & \textbf{28.70} & 69.28 \\
\midrule
16 & FP16 & 6.46 & 55.89 & 72.77 & 81.94 & 81.34 & 76.64 & 89.84 & 69.00 & 28.50 & 69.49 \\
\bottomrule
\end{tabular}
\end{table*}

\clearpage
\section{Comparison with Single-Precision Quantization}  
\label{app:static}

\begin{table*}[t]
\centering
\caption{Single-precision GPTQ: WikiText-2 PPL and accuracy (\%) on eight benchmarks; Avg. is their mean. Each bit width is a separate model. Any-Rate and MG (2/3/4) are in Tables~\ref{tab:full_a} and~\ref{tab:full_b}.}
\label{tab:static}
\footnotesize
\setlength{\tabcolsep}{3.0pt}
\renewcommand{\arraystretch}{1.0}
\begin{tabular}{clc|cccccccc|>{\columncolor[gray]{0.92}}c}
\toprule
Bits & Method & PPL & ARC-c & ARC-e & HellaSwag & PIQA & WinoGrande & GSM8K & TriviaQA & NQ-open & Avg. \\
\midrule
\rowcolor[gray]{0.82} \multicolumn{12}{c}{\textbf{Qwen2.5-3B}} \\
\midrule
 2 & GPTQ & $>$1000 & 27.56 & 25.76 & 26.50 & 50.65 & 48.30 & 1.67 & 0.00 & 0.00 & 22.55 \\
 3 & GPTQ & 10.33 & 41.13 & 64.44 & 69.34 & 76.12 & 62.75 & 32.15 & 29.90 & 10.20 & 48.25 \\
 4 & GPTQ & 8.41 & 49.23 & 75.29 & 72.92 & 78.24 & 67.80 & 70.51 & 46.50 & 16.60 & 59.64 \\
\midrule
16 & FP16 & 8.03 & 47.27 & 73.11 & 73.55 & 78.89 & 68.35 & 75.51 & 50.30 & 18.20 & 60.65 \\
\midrule
\rowcolor[gray]{0.82} \multicolumn{12}{c}{\textbf{Gemma-3-4b-it}} \\
\midrule
 2 & GPTQ & $>$1000 & 22.44 & 27.61 & 27.80 & 50.92 & 51.46 & 0.68 & 0.10 & 0.00 & 22.63 \\
 3 & GPTQ & 17.38 & 46.42 & 72.60 & 67.35 & 74.86 & 64.09 & 22.14 & 33.70 & 11.40 & 49.07 \\
 4 & GPTQ & 17.03 & 56.66 & 77.82 & 73.02 & 77.42 & 67.64 & 70.66 & 46.90 & 15.90 & 60.75 \\
\midrule
16 & FP16 & 16.07 & 57.08 & 77.95 & 74.13 & 76.99 & 69.38 & 77.41 & 49.00 & 18.30 & 62.53 \\
\midrule
\rowcolor[gray]{0.82} \multicolumn{12}{c}{\textbf{Llama-3.1-8B-Instruct}} \\
\midrule
 2 & GPTQ & $>$1000 & 26.19 & 26.30 & 26.38 & 51.52 & 48.38 & 0.61 & 0.00 & 0.00 & 22.42 \\
 3 & GPTQ & 9.42 & 43.69 & 68.64 & 74.98 & 75.90 & 70.48 & 43.06 & 48.00 & 16.40 & 55.14 \\
 4 & GPTQ & 7.55 & 54.61 & 79.34 & 78.83 & 80.47 & 72.93 & 68.61 & 71.00 & 25.70 & 66.44 \\
\midrule
16 & FP16 & 7.21 & 55.20 & 79.59 & 79.26 & 80.96 & 73.88 & 76.50 & 75.30 & 30.30 & 68.87 \\
\midrule
\rowcolor[gray]{0.82} \multicolumn{12}{c}{\textbf{Phi-4 (14B)}} \\
\midrule
 2 & GPTQ & $>$1000 & 24.32 & 31.94 & 31.70 & 55.77 & 51.14 & 2.05 & 0.00 & 0.10 & 24.63 \\
 3 & GPTQ & 7.19 & 54.27 & 70.83 & 79.62 & 80.79 & 71.27 & 84.61 & 59.80 & 23.20 & 65.55 \\
 4 & GPTQ & 6.59 & 55.38 & 71.42 & 81.71 & 81.01 & 74.43 & 89.08 & 67.30 & 28.70 & 68.63 \\
\midrule
16 & FP16 & 6.46 & 55.89 & 72.77 & 81.94 & 81.34 & 76.64 & 89.84 & 69.00 & 28.50 & 69.49 \\
\bottomrule
\end{tabular}
\end{table*}

\subsection{Storage and Precision Switching}

This section compares Any-Rate with single-precision quantization, which quantizes a separate model for each bit width. To serve several precisions, a single-precision method must store one model per bit width, so its storage grows with the number of modes. With single-precision GPTQ, serving the 2--4-bit modes of Llama-3.1-8B-Instruct takes 9.8 bits per weight, and serving every integer mode from 2 to 8 bits takes 36.8 (Figure~\ref{fig:static}(c)). Any-Rate serves the 2--4-bit modes, including two half-bit modes, from its ladder at 4.21 bits per weight, and all nine modes from 2 to 8 bits from one checkpoint at 8.30, about a quarter of the single-precision cost. MG (2/3/4) also stores one checkpoint (4.26 bits per weight), but it serves only the three integer modes from 2 to 4 bits. The advantage extends to runtime, when the precision changes. Single-precision models share no data, so switching the precision discards the weights already held in memory or caches and reads the entire new model. All modes of Any-Rate read the same planes from the first one, so the planes already loaded are reused: a lower mode reads fewer of them, and a higher mode reads only the additional planes. Since each mode reads only its own planes, the memory traffic of decoding also scales with the selected precision (Appendix~\ref{app:efficiency}). The remaining question is how much accuracy this costs.

\subsection{Baselines}

\begin{figure*}[t]
\centering
\includegraphics[width=\textwidth]{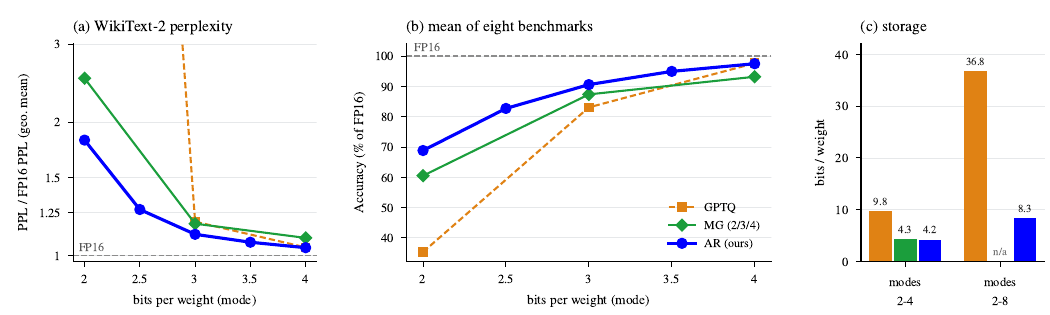}
\caption{Any-Rate versus single-precision GPTQ and MG (2/3/4), averaged over the four models. (a) WikiText-2 PPL relative to FP16 (geometric mean). (b) Mean accuracy of the eight benchmarks relative to FP16. (c) Storage to serve the 2--4-bit and the 2--8-bit modes (Llama-3.1-8B-Instruct); GPTQ stores one model per bit width, and MG (2/3/4) supports only 2--4 bits.}
\label{fig:static}
\end{figure*}

Since Any-Rate builds on the GPTQ framework, using the same Hessian-related Grams and its column-order error feedback, we take GPTQ~\citep{frantar2023gptq} as the single-precision baseline, and we add MG (2/3/4), which nests GPTQ across 2, 3, and 4 bits with MatGPTQ~\citep{kleinegger2026matgptq}. Single-precision GPTQ is thus the single-mode reference for both nested methods: the gap between MG (2/3/4) and GPTQ shows the nesting cost of a shared scalar grid, and the gap between Any-Rate and GPTQ the net effect of coded planes and per-mode targets (Section~\ref{sec:an_nest}). GPTQ uses an asymmetric scale and zero point per group of 128 with \texttt{act\_order}, which add 0.25 bits per weight and are included in the storage of Figure~\ref{fig:static}(c). GPTQ uses the same evaluation code as Any-Rate and quantizes each of the four models at 2, 3, and 4 bits, and MG (2/3/4) is the checkpoint of our main results. Table~\ref{tab:static} lists the per-model results of GPTQ, and Tables~\ref{tab:full_a} and~\ref{tab:full_b} those of MG (2/3/4) and Any-Rate.

\subsection{Accuracy}

In Figure~\ref{fig:static}(a, b), the two nested methods depart from single-precision GPTQ in opposite ways. MG (2/3/4) is above GPTQ at 2 and 3 bits on every model, but at 4 bits it falls below GPTQ on three of the four models, to 93\% of the FP16 accuracy on average against 98\% for GPTQ: this is the nesting cost of a shared scalar grid, which grows with the bit width (Section~\ref{sec:an_nest}). Any-Rate is above both at 2 and 3 bits on average and is never more than 0.7 points below GPTQ at 4 bits, so its ladder avoids most of this cost. GPTQ collapses at 2 bits, whereas both nested methods keep usable accuracy there. The half-bit modes add further room: Any-Rate at 2.5 bits comes close to GPTQ at 3 bits, and at 3.5 bits it exceeds MG (2/3/4) at 4 bits on average. At 2 and 3 bits, the differences among the three methods are more pronounced in the generative tasks (GSM8K, TriviaQA, and NQ-open) than in the multiple-choice tasks. Any-Rate thus matches single-precision GPTQ at 4 bits and exceeds it below, while serving every mode from one checkpoint with about a quarter of the storage.

\section{Sample Problems} 
\label{app:samples}

\subsection{GSM8K, Phi-4 (idx: 177)}

\newlength{\respmargin}\setlength{\respmargin}{2.5em}
\newenvironment{resp}[1]{\begin{list}{}{\setlength{\leftmargin}{\respmargin}\setlength{\rightmargin}{\respmargin}\setlength{\listparindent}{0pt}\setlength{\itemindent}{0pt}\setlength{\topsep}{1pt}\setlength{\partopsep}{0pt}\setlength{\parsep}{0pt}}\item[]{\rmfamily\small\bfseries #1}\par\ttfamily\footnotesize\spaceskip=0.45em plus 0.5em minus 0.1em\relax}{\end{list}}
{\small\sloppy\emergencystretch=3em \textbf{Question:} Zaid spends 1/4 of his salary on rent, 1/3 on car fuel and donates half of the remaining amount to his favorite charity. He gives his daughter 200\$ to use for her weekly expenses and 700\$ to his wife to budget for groceries and other household goods. If Zaid earns 6000\$ per month, how much money will he still have after all these expenses and donations? \textit{(gold: 350)}\par}

\subsubsection{2-bit}
{\small\sloppy\emergencystretch=3em
\begin{resp}{AP: \textcolor{red}{False}}Zaid spends 6000/4 =\allowbreak{} {<}{<}6000/4=\allowbreak{}1500{>}{>}\allowbreak{}1500 on rent. He spends 6000/3 =\allowbreak{} {<}{<}6000/3=\allowbreak{}2000{>}{>}\allowbreak{}2000 on car fuel. He donates 6000 - 1500 - 2000 =\allowbreak{} {<}{<}6000-1500-2000=\allowbreak{}2500{>}{>}\allowbreak{}2500 to his favorite charity. He gives his daughter 200 and his wife 700, so he spends 2500 +\allowbreak{} 700 =\allowbreak{} {<}{<}2500+\allowbreak{}700=\allowbreak{}3200{>}{>}\allowbreak{}3200 in total. So, he has 6000 - 1500 - 2000 - 2500 - 3200 =\allowbreak{} {<}{<}6000-1500-2000-2500-3200=\allowbreak{}100{>}{>}\allowbreak{}100 left. \#\#\#\# \textcolor{red}{100}\end{resp}
\smallskip
\begin{resp}{AB: \textcolor{red}{False}}Zaid spends 6000/4 =\allowbreak{} \${<}{<}6000/4=\allowbreak{}1500{>}{>}\allowbreak{}1500 on rent. He spends 6000/3 =\allowbreak{} \${<}{<}6000/3=\allowbreak{}2000{>}{>}\allowbreak{}2000 on car fuel. So he has 6000 - 1500 - 2000 =\allowbreak{} \${<}{<}6000-1500-2000=\allowbreak{}2500{>}{>}\allowbreak{}2500 left. He gives his daughter 200\$ and his wife 700\$. So he has 2500 - 200 - 700 =\allowbreak{} \${<}{<}2500-200-700=\allowbreak{}1600{>}{>}\allowbreak{}1600 left. \#\#\#\# \textcolor{red}{1600}\end{resp}
\smallskip
\begin{resp}{MG (2/3/4): \textcolor{red}{False}}Zaid spends 6000/4 =\allowbreak{} {<}{<}6000/4=\allowbreak{}1500{>}{>}\allowbreak{}1500\$ on rent. He spends 6000/3 =\allowbreak{} {<}{<}6000/3=\allowbreak{}2000{>}{>}\allowbreak{}2000\$ on car fuel. He has 6000 - 1500 - 2000 =\allowbreak{} {<}{<}6000-1500-2000=\allowbreak{}2500{>}{>}\allowbreak{}2500\$ left. He donates half of the remaining amount, so he donates 2500/2 =\allowbreak{} {<}{<}2500/2=\allowbreak{}1250{>}{>}\allowbreak{}1250\$. He gives his daughter 200\$ and his wife 700\$. So he has 2500 - 1250 - 200 - 700 =\allowbreak{} {<}{<}2500-1250-200-700=\allowbreak{}450{>}{>}\allowbreak{}450\$ left. \#\#\#\# \textcolor{red}{450}\end{resp}
\smallskip
\begin{resp}{AR (ours): \textcolor{blue}{True}}Zaid spends 6000/4 =\allowbreak{} \${<}{<}6000/4=\allowbreak{}1500{>}{>}\allowbreak{}1500 on rent. He spends 6000/3 =\allowbreak{} \${<}{<}6000/3=\allowbreak{}2000{>}{>}\allowbreak{}2000 on car fuel. He has 6000 - 1500 - 2000 =\allowbreak{} \${<}{<}6000-1500-2000=\allowbreak{}2500{>}{>}\allowbreak{}2500 left. He donates 2500/2 =\allowbreak{} \${<}{<}2500/2=\allowbreak{}1250{>}{>}\allowbreak{}1250 to his favorite charity. He gives 200 +\allowbreak{} 700 =\allowbreak{} \${<}{<}200+\allowbreak{}700=\allowbreak{}900{>}{>}\allowbreak{}900 to his family. So he has 2500 - 1250 - 900 =\allowbreak{} \${<}{<}2500-1250-900=\allowbreak{}350{>}{>}\allowbreak{}350 left. \#\#\#\# \textcolor{blue}{350}\end{resp}
}

\subsubsection{2.5-bit}
{\small\sloppy\emergencystretch=3em
\begin{resp}{AR (ours): \textcolor{blue}{True}}Zaid spends 6000/4 =\allowbreak{} \${<}{<}6000/4=\allowbreak{}1500{>}{>}\allowbreak{}1500 on rent. He spends 6000/3 =\allowbreak{} \${<}{<}6000/3=\allowbreak{}2000{>}{>}\allowbreak{}2000 on car fuel. He has 6000 - 1500 - 2000 =\allowbreak{} \${<}{<}6000-1500-2000=\allowbreak{}2500{>}{>}\allowbreak{}2500 left. He donates 2500/2 =\allowbreak{} \${<}{<}2500/2=\allowbreak{}1250{>}{>}\allowbreak{}1250 to his favorite charity. He has 2500 - 1250 =\allowbreak{} \${<}{<}2500-1250=\allowbreak{}1250{>}{>}\allowbreak{}1250 left. He gives his daughter 200\$. He gives his wife 700\$. He has 1250 - 200 - 700 =\allowbreak{} \${<}{<}1250-200-700=\allowbreak{}350{>}{>}\allowbreak{}350 left. \#\#\#\# \textcolor{blue}{350}\end{resp}
}

\subsection{NQ-open, Phi-4 (idx: 949, 78, 162)}

\begin{table}[h]
\caption{NQ-open examples (Phi-4). Correct answers in blue, wrong in red; ``--'': no such mode.}
\label{tab:qual_nqopen}
\centering
\small
\begin{tabular}{lccccc}
\toprule
Method & 2-bit & 2.5-bit & 3-bit & 3.5-bit & 4-bit \\
\midrule
\multicolumn{6}{c}{\textbf{Q:} Who is the guy that jumped from space? \textit{(gold: Felix Baumgartner)}} \\
\midrule
AP & \textcolor{red}{Chris Hadfield} & -- & \textcolor{red}{Gary Powers} & -- & \textcolor{blue}{Felix Baumgartner} \\
AB & \textcolor{red}{Yevgeny Khar} & -- & \textcolor{blue}{Felix Baumgartner} & -- & \textcolor{blue}{Felix Baumgartner} \\
MG (2/3/4) & \textcolor{red}{Felix Baum} & -- & \textcolor{blue}{Felix Baumgartner} & -- & \textcolor{blue}{Felix Baumgartner} \\
\rowcolor[HTML]{DCEBFA}
AR (ours) & \textcolor{blue}{Felix Baumgartner} & \textcolor{blue}{Felix Baumgartner} & \textcolor{blue}{Felix Baumgartner} & \textcolor{blue}{Felix Baumgartner} & \textcolor{blue}{Felix Baumgartner} \\
\midrule
\multicolumn{6}{c}{\textbf{Q:} Who captained the first European ship to sail around the tip of Africa? \textit{(gold: Bartolomeu Dias)}} \\
\midrule
AP & \textcolor{red}{Vasco da Gama} & -- & \textcolor{blue}{Bartolomeu Dias} & -- & \textcolor{blue}{Bartolomeu Dias} \\
AB & \textcolor{red}{Bartolom\'e de L\'eon} & -- & \textcolor{blue}{Bartolomeu Dias} & -- & \textcolor{blue}{Bartolomeu Dias} \\
MG (2/3/4) & \textcolor{red}{Vasco da Gama} & -- & \textcolor{blue}{Bartolomeu Dias} & -- & \textcolor{blue}{Bartolomeu Dias} \\
\rowcolor[HTML]{DCEBFA}
AR (ours) & \textcolor{blue}{Bartolomeu Dias} & \textcolor{blue}{Bartolomeu Dias} & \textcolor{blue}{Bartolomeu Dias} & \textcolor{blue}{Bartolomeu Dias} & \textcolor{blue}{Bartolomeu Dias} \\
\midrule
\multicolumn{6}{c}{\textbf{Q:} Who won the mens single ice skating 2018? \textit{(gold: Yuzuru Hanyu)}} \\
\midrule
AP & \textcolor{red}{Yuzak} & -- & \textcolor{blue}{Yuzuru Hanyu} & -- & \textcolor{blue}{Yuzuru Hanyu} \\
AB & \textcolor{red}{Patrick Chan} & -- & \textcolor{red}{Nathan Chen} & -- & \textcolor{blue}{Yuzuru Hanyu} \\
MG (2/3/4) & \textcolor{red}{Yuzuri} & -- & \textcolor{blue}{Yuzuru Hanyu} & -- & \textcolor{blue}{Yuzuru Hanyu} \\
\rowcolor[HTML]{DCEBFA}
AR (ours) & \textcolor{blue}{Yuzuru Hanyu} & \textcolor{blue}{Yuzuru Hanyu} & \textcolor{blue}{Yuzuru Hanyu} & \textcolor{blue}{Yuzuru Hanyu} & \textcolor{blue}{Yuzuru Hanyu} \\
\bottomrule
\end{tabular}
\end{table}

\section{Method Details} 

\subsection{Layer-wise Objective and Error Feedback}
\label{app:background}

A linear layer computes $WX$, and its quantized version computes $\widehat W X$. With $E = W - \widehat W$, 
the layer-wise reconstruction loss in Eq.~\eqref{eq:layer_loss} is given by
\begin{equation}
\|EX\|_F^2 \;=\; \mathrm{tr}\!\big(E\,\Gamma\,E^{\top}\big), \qquad \Gamma = XX^{\top},
\end{equation}
where $\Gamma$ is the Gram matrix of the inputs, proportional to the Hessian of the reconstruction loss, and is referred to as the Hessian following GPTQ~\citep{frantar2023gptq}. The error is the sum of $E_r \Gamma E_r^{\top}$ over the rows $E_r$ of $E$, so the rows are independent: the encoder chooses the bits and the LUT of each row separately, and only $\Gamma$ is shared across the matrix. Linear layers that take the same input within a decoder layer share $\Gamma$, so each decoder layer has four Hessians (q/k/v, o, gate/up, and down). We compute $\Gamma$ from the inputs of the FP16 model. The calibration data pass through each FP16 layer, which accumulates $\Gamma$ from its inputs, and after encoding, the input of the next layer is the output of the FP16 layer rather than of the quantized one. The quantization error of earlier layers therefore never enters the $\Gamma$ of later layers, and each layer can be encoded independently. Appendix~\ref{app:hyper} describes the calibration data.

As in GPTQ, we process the columns from left to right and pass the error of each processed column to the columns not yet processed (error feedback). The columns keep their original order (no \texttt{act\_order}). We damp $\Gamma$ by adding 1\% of its mean diagonal, after setting any zero diagonal entry to 1. With the upper-triangular Cholesky factor $R$ of $\Gamma^{-1} = R^{\top}R$, the target and the error of column $i$ are
\begin{equation}
\tilde\tau_i \;=\; \tau_i - \sum_{l\in\mathcal{B}(i),\, l<i} e_l\,R_{li}, \qquad e_i \;=\; \frac{\tilde\tau_i - q_i}{R_{ii}},
\end{equation}
where $\mathcal{B}(i)$ is the eight-column block that contains column $i$, $\tau_i$ is the target of column $i$, which already includes the errors of earlier blocks, and $q_i$ is the chosen quantized value; all quantities are per row. We process eight columns as a block and pass the errors of the block to the remaining columns at once when the block ends; Algorithm~\ref{alg:block} gives the choice within a block.

\subsection{Coded Planes and the Hamming Code}
\label{app:coded}

An unconstrained (U) plane stores one bit per weight. A coded plane groups eight consecutive columns of a row into a block and constrains the eight bits $v = (v_0,\dots,v_7)$ that it holds in the block to be a codeword of a code $\mathcal{C}\subset\{0,1\}^8$. The plane is then specified by $\log_2|\mathcal{C}|$ bits per block, so its rate is $\log_2|\mathcal{C}|/8$ bits per weight; a U plane corresponds to $|\mathcal{C}| = 256$. We use the extended Hamming $[8,4,4]$ code~\citep{conway1999splag}, which is also the first-order Reed--Muller code RM(1,3), and call a plane with this code an H plane. Its 16 codewords are the binary linear combinations of
\begin{equation}
(1,1,1,1,1,1,1,1),\quad (0,1,0,1,0,1,0,1),\quad (0,0,1,1,0,0,1,1),\quad (0,0,0,0,1,1,1,1).
\end{equation}
Since only 16 of the 256 bit patterns are allowed, a 4-bit message $(a_0,\dots,a_3)$ per block suffices, and the rate is 0.5 bits per weight. The minimum distance is 4, so two distinct codewords differ in at least four columns; Section~\ref{sec:an_code} and Appendix~\ref{app:an_proof} analyze the gain that this structure brings. The bit of column $j\in\{0,\dots,7\}$ of the block, written in binary as $j = 4j_2 + 2j_1 + j_0$, is $v_j = a_0 \oplus j_0 a_1 \oplus j_1 a_2 \oplus j_2 a_3$, so decoding is an XOR of the message bits selected by the column position; a 16-entry codeword table gives the same result. Other codes such as the single-parity $[8,7,2]$ or the repetition $[8,1,8]$ code fit the same scheme, but we use only the $[8,4,4]$ code, whose rate of exactly 0.5 gives the half-bit modes.

A layout is the string of plane types starting from the most significant bit of the index. The rate of the mode that reads the first $k$ planes is the sum of the rates of those planes (1 for U, 0.5 for H), so UHHHHHH has seven planes and a total rate of 4 bits; Table~\ref{tab:layout_modes} lists the rate of each mode. In one block, UHHHHHH stores the 8 bits of the U plane and a 4-bit message for each of the six H planes, 32 bits in total or 4 bits per weight. In the file, a U plane packs the bits of eight columns into one byte, and an H plane packs the messages of two blocks into one byte; the released code documents the exact bit order. At the same rate, H planes increase the number of planes: two bits take two planes with U only but three with UHH, so the index is one bit longer, and both the LUT entries and the values a single weight can take double. In exchange, the bits of a block are restricted to codewords, so the encoder has to search over them (Appendix~\ref{app:planejoint}).

\begin{table}[h]
\caption{Rate of the mode that reads the first $k$ planes of UHHHHHH.}
\label{tab:layout_modes}
\centering
\small
\begin{tabular}{lccccc}
\toprule
Planes read $k$ & 3 & 4 & 5 & 6 & 7 \\
Planes & UHH & UHHH & UHHHH & UHHHHH & UHHHHHH \\
Rate (bits/weight) & 2 & 2.5 & 3 & 3.5 & 4 \\
\bottomrule
\end{tabular}
\end{table}

\subsection{Encoding Procedure for a Matrix}
\label{app:algorithm}

Encoding a matrix takes the weights $W$, the Hessian $\Gamma$, a layout, a set of modes, and mode weights $\lambda_m$, and produces the index and the shared LUT. Since the rows are independent, every step runs on all rows in parallel, and the codewords and U bits are chosen per row. We first form the damped Cholesky factor $R$ from $\Gamma$ (Appendix~\ref{app:background}) and initialize the LUTs. The LUT of the lowest mode (three planes, eight entries) is obtained per row by k-means~\citep{lloyd1982least} weighted by the diagonal of $\Gamma$, and the LUT of each next mode splits every entry into two, shifted up and down by a quarter of the mean gap between adjacent entries; this follows the incremental upscaling of Any-Precision LLM (AP)~\citep{park2024anyprecision}, but with a fixed shift instead of re-clustering. Because the two children average to their parent, the initial LUTs already satisfy the averaging relation between mode LUTs, and these LUTs only serve as a starting point for the refitting below.

Encoding and LUT refitting then form one round, repeated $N_{\mathrm{ref}}$ times. In a round, the target of every mode starts at $W$, and the columns are processed from the left in blocks of eight; for each block, Algorithm~\ref{alg:block} chooses the index and passes the errors to the targets of the remaining columns, with the LUT held fixed. Once the pass ends and the index is fixed, the shared LUT is refit by closed-form least squares (Appendix~\ref{app:table}), and the round's objective is the $\lambda_m$-weighted sum of the relative proxy losses (Eq.~\eqref{eq:proxy_loss}) of the modes reconstructed with the refit LUT. A new LUT changes the best index, so the next round encodes again with it. Since the objective does not always decrease from round to round, we keep the index and LUT of the round with the lowest objective. Appendix~\ref{app:hyper} lists the mode weights, $N_{\mathrm{ref}}$, the beam width $B$, the number of evaluated candidates $K$, and the damping.

\subsection{Shared Look Up Table}
\label{app:table}

The mode that reads the first $k$ planes looks up its LUT $T_k$ with the first $k$ bits of the index (Eq.~\eqref{eq:prefix_reconstruction}). AP stores a separate LUT $T_k$ for each mode, whereas Any-Rate stores only the LUT $T$ of the top mode and derives the others from it by averaging. Let $W_r\in\mathbb{R}^c$ be the weights of a row (as a column vector), $\iota\in\{0,\dots,2^b-1\}^c$ its index, and $P\in\{0,1\}^{c\times 2^b}$ the one-hot matrix that marks the entry each weight points to, so that the top mode reconstructs $PT$. Mode $m$, which reads the first $k_m$ planes, averages the $2^{b-k_m}$ entries that share the same first $k_m$ bits. With the block-diagonal averaging matrix $M_m$, whose entries are $1/2^{b-k_m}$ between entries sharing the first $k_m$ bits and 0 otherwise, mode $m$ reconstructs $PM_mT$. For a fixed index, the objective of the row,
\begin{equation}
\sum_m \lambda_m\,(W_r - PM_mT)^{\top}\Gamma\,(W_r - PM_mT),
\end{equation}
is quadratic in $T$. With $A = P^{\top}\Gamma P$ and $h = P^{\top}\Gamma W_r$, its minimizer solves the normal equations
\begin{equation}
\Big(\sum_m \lambda_m\,M_m^{\top}A\,M_m\Big)\,T \;=\; \sum_m \lambda_m\,M_m^{\top}h.
\end{equation}
We add $10^{-6}$ times the mean diagonal to the left-hand matrix to keep it nonsingular and solve the $2^b\times 2^b$ systems of all rows at once. Because the errors of all modes enter one equation, the resulting $T$ fits the low and high modes together.

The LUT is stored in 8 bits. For each row, the minimum $\mathrm{lo}$ and the step $s = (\max T - \min T)/255$ are first rounded to FP16, and the codes are then set to $\mathrm{round}\big((T-\mathrm{lo})/s\big)\in\{0,\dots,255\}$, so the stored file reproduces exactly the LUT the encoder used. The index choice of the next round and the objective of the current round are both computed with this 8-bit LUT, so the error of the 8-bit quantization is already accounted for in the encoding. The metadata of a row is $2^b$ 8-bit codes and two FP16 values, $8\cdot 2^b + 32$ bits, or $(8\cdot 2^b + 32)/c$ bits per weight for $c$ columns, whereas separate FP16 LUTs for every mode would take $16\sum_k 2^k/c$. For UHHHHHH ($b=7$, modes $k=3,\dots,7$), this is 1056 against 3968 bits per row, or 0.26 against 0.97 bits per weight at $c=4096$: sharing roughly halves the cost, and 8-bit storage halves it again. On average over the linear layers of Llama-3.1-8B-Instruct, where $c$ is 4096 or 14336, this is 0.21 against 0.78 bits per weight.

\subsection{Residual Planes}
\label{app:residual}

The 4-bit mode of the ladder reconstructs $\widehat w_4$ from the index and the shared LUT. Each weight has a residual $r = w - \widehat w_4$ and stores a separate 4-bit residual index $\rho\in\{0,\dots,15\}$. The planes of the ladder are index bits into the shared LUT, whereas the four residual planes are index bits of an offset added to $\widehat w_4$. They start from the most significant bit, cost 1 bit per weight each, and carry no code, since above 4 bits the modes differ little in quality and half-bit steps are not needed. The $m$-bit mode ($m=5,\dots,8$) looks up the offset table $\delta_m$ of size $2^{m-4}$ with the first $m-4$ bits of $\rho$ (Eq.~\eqref{eq:residual_correction}). We do not extend the ladder with more planes, because the shared LUT doubles with every plane: adding four U planes ($b=11$) also gives 8 bits per weight, but the 11-bit index makes the metadata grow to $8\cdot 2^{11}+32 = 16416$ bits per row, or 3.24 bits per weight on Llama-3.1-8B-Instruct, whereas the residual needs only 30 offsets per row, or 0.09 bits per weight (Table~\ref{tab:residual_cost}). The residual is stored in a file separate from the ladder, so serving only the 2--4-bit modes does not need it.

The residual index is chosen column by column from the left. Without a code, each column has only 16 candidates, so they are compared directly without a search, and columns are not grouped into blocks. Each of the four modes keeps its own target $\tau_m$ that receives its errors (per-mode targets), and at column $j$ we choose
\begin{equation}
\rho_j \;=\; \arg\min_{\rho}\ \sum_{m=5}^{8}\Big(\tau_{m,j} - \widehat w_{4,j} - \delta_m\big[\rho\gg(8-m)\big]\Big)^2,
\end{equation}
after which each mode passes its error $\big(\tau_{m,j} - \widehat w_{m,j}\big)/R_{jj}$ to the targets of the remaining columns (Appendix~\ref{app:background}). The offset tables are set as follows. With $r_{95}$ the 95th percentile of $|r|$ in the row, the centers of $2^{m-4}$ equal-width intervals of $[-r_{95}, r_{95}]$ initialize $\delta_m$. After the index is chosen with these values, each entry of $\delta_m$ is replaced by the mean residual $r$ of the weights assigned to it (kept if no weight is assigned), and the index is chosen once more. This refit uses plain means without Hessian weighting. Each residual plane packs eight columns per byte, and the 30 offsets are stored per row in FP16, which is 480 bits per row or $480/c$ bits per weight. Table~\ref{tab:residual_err} lists the relative proxy loss of each mode.

\begin{table}[t]
\caption{Bits per weight of three ways to support the 5--8-bit modes (Llama-3.1-8B-Instruct, linear layers).}
\label{tab:residual_cost}
\centering
\small
\begin{tabular}{lccccc}
\toprule
Configuration & $b$ & Index planes & Shared LUT & Residual offsets & Total \\
\midrule
Ladder only (2--4 bits) & 7 & 4.00 & 0.21 & -- & 4.21 \\
Ladder extended by four U planes (2--8 bits) & 11 & 8.00 & 3.24 & -- & 11.24 \\
Ladder + residual (2--8 bits) & 7 & 8.00 & 0.21 & 0.09 & 8.30 \\
\bottomrule
\end{tabular}
\end{table}

\begin{table}[h]
\caption{Relative proxy loss (\%) of the residual modes (Qwen2.5-3B, mean over 252 matrices).}
\label{tab:residual_err}
\centering
\small
\begin{tabular}{lccccc}
\toprule
Mode (bits) & 4 & 5 & 6 & 7 & 8 \\
\midrule
Relative proxy loss (\%) & 0.422 & 0.358 & 0.131 & 0.059 & 0.035 \\
\bottomrule
\end{tabular}
\end{table}

\section{Encoder Design Choices} 
\label{app:encoder}

The encoder involves several choices: which planes are U and which are H (the layout), how the planes of a block are encoded together, the width of the candidate search, and the ratio of the mode weights. Encoding a full model and measuring its PPL takes hours, too long to try every combination. We therefore narrow the choices with the proxy loss of Appendix~\ref{app:proxy}, computed on real weight matrices, and check for a few settings that the order of the proxy losses matches the order of the PPL of fully encoded models. All comparisons in this section use Qwen2.5-3B and, except for Appendix~\ref{app:weights}, which compares mode weights, our mode weights 1-1-2-3-4 (for the 2, 2.5, 3, 3.5, and 4-bit modes).

\subsection{Proxy Loss}
\label{app:proxy}

The proxy loss is the output error of a layer, $\mathcal{L}_m^{\mathrm{layer}}=\mathrm{tr}(E_m\Gamma E_m^{\top})$ (Eq.~\eqref{eq:layer_loss}). It is the layer-wise reconstruction error that GPTQ~\citep{frantar2023gptq} and its successors minimize, and for a ladder our encoder minimizes its mode-weighted sum, Eq.~\eqref{eq:joint_objective}. Since the scale of weights and inputs varies across layers, we report the relative proxy loss $L_m$ of Eq.~\eqref{eq:proxy_loss}, normalized by $\mathrm{tr}(W\Gamma W^{\top})$ of the original weights. Layouts support different modes, so their weighted objectives are not comparable, and we record each mode separately. Because the objective separates over rows, the costlier comparisons (Appendix~\ref{app:search}) estimate the relative proxy loss from 128 randomly chosen rows per matrix. The proxy reflects only the output error of one layer, not the accumulation of errors across layers or the nonlinearity between them, so it does not predict the PPL exactly; it does, however, order settings whose proxy losses differ substantially in the same way as the PPL (Appendix~\ref{app:layout}), which is why we use it to compare settings here and in Section~\ref{sec:analysis}.

\subsection{Layout}
\label{app:layout}

We measure the proxy loss on all 252 linear-layer matrices of Qwen2.5-3B (base), 36 layers $\times$ 7 projections. For each layer, the Hessian is computed from the calibration data (Appendix~\ref{app:hyper}), and every row of the matrix is encoded with beam width 64 to obtain the relative proxy loss of each mode. The mode weights are 1-1-2-3-4; modes that a layout does not support are dropped and the remaining weights are kept (e.g., 1-1-2-4 for UHHHHU and 1-2-4 for UUUU). Table~\ref{tab:layout} reports the relative proxy loss of each layout as a ratio to UHHHHHH at the same bit width (geometric mean over the 252 matrices), together with the PPL of the fully encoded model.

\begin{table}[t]
\caption{Relative proxy loss (w.r.t.\ UHHHHHH) and WikiText-2 PPL of each layout (Qwen2.5-3B). Bits/weight includes the LUT metadata. Best in bold, second best underlined; ``--'' means the layout has no such mode.}
\label{tab:layout}
\centering
\small
\setlength{\tabcolsep}{4pt}
\begin{tabular}{lc ccccc ccccc|c}
\toprule
 & & \multicolumn{5}{c}{Proxy loss} & \multicolumn{5}{c|}{WikiText-2 PPL} & \\
\cmidrule(lr){3-7} \cmidrule(lr){8-12}
Layout & Bits/weight & 2 & 2.5 & 3 & 3.5 & 4 & 2 & 2.5 & 3 & 3.5 & 4 & FP16 \\
\midrule
UUUU & 4.06 & 1.510 & -- & 1.591 & -- & 1.741 & 78.46 & -- & 11.32 & -- & 9.25 & \multirow{5}{*}{8.03} \\
UHHHHU & 4.20 & \textbf{0.959} & \textbf{0.960} & \underline{1.003} & -- & 1.131 & \textbf{15.08} & \textbf{10.77} & \underline{9.39} & -- & \underline{8.53} & \\
HHHHHHU & 4.39 & 4.534 & 4.866 & 5.152 & -- & 6.376 & $>$1000 & $>$1000 & 423.22 & -- & 17.96 & \\
UUHHHH & 4.20 & 1.614 & 1.229 & 1.113 & \underline{1.076} & \underline{1.070} & 123.38 & 14.45 & 9.85 & \underline{8.97} & 8.57 & \\
UHHHHHH (ours) & 4.39 & \underline{1.000} & \underline{1.000} & \textbf{1.000} & \textbf{1.000} & \textbf{1.000} & \underline{15.41} & \underline{10.78} & \textbf{9.29} & \textbf{8.73} & \textbf{8.45} & \\
\bottomrule
\end{tabular}
\end{table}

Proxy loss and PPL rank the layouts almost identically. The only pair whose order flips is UHHHHU and UUHHHH at 4 bits, whose PPLs differ by 0.04, so the proxy can separate layouts that differ substantially but cannot decide smaller differences. HHHHHHU, whose first plane is H, has a 4--6 times larger proxy loss and collapses in PPL at low bit widths, so the first plane must be U. UUUU, with U planes only, has no half-bit modes and also falls behind at the integer bit widths, and UUHHHH, with two leading U planes, is much worse at 2 bits; using H planes early favors the low bit widths. The two layouts with one U followed by H planes are close. UHHHHU, whose last plane is U, matches or slightly beats UHHHHHH at 2--2.5 bits but has no 3.5-bit mode and falls behind at 4 bits. UHHHHHH and UUHHHH are the only layouts with all five modes, and UHHHHHH is better at every bit width in both proxy loss and PPL while staying close to the best layout at 2--2.5 bits; we therefore use UHHHHHH.

\subsection{Efficient Plane-Joint Encoding}
\label{app:planejoint}

With U planes only, each column chooses its bits from its own target, which already includes the errors of earlier columns. An H plane, however, ties the bits of the eight columns of a block to one of 16 codewords, so the bits of one column depend on the choices of the other seven, and the codewords are chosen jointly per block (Section~\ref{sec:planejoint}). Methods whose planes are all U, such as Any-Precision LLM (AP)~\citep{park2024anyprecision}, AnyBCQ (AB)~\citep{park2026anybcq}, and MatGPTQ (MG)~\citep{kleinegger2026matgptq}, decompose over columns and do not face this problem. Let the columns of a block be $i=1,\dots,n$ ($n=8$) and its H planes $p=1,\dots,n_{\mathrm{H}}$. A candidate $s=(s_1,\dots,s_{n_{\mathrm{H}}})$ chooses one codeword $s_p\in\mathcal{C}$ for each H plane, giving $16^{n_{\mathrm{H}}}$ candidates. A candidate fixes the H bits $a_i(s)\in\{0,1\}^{n_{\mathrm{H}}}$ of column $i$ and leaves only its U bits $u$. Mode $m$, which reads the first $k_m$ planes, gives column $i$ the value
\begin{equation}
q_{m,i}(a,u)=T_{k_m}\!\left[\iota(a,u)\gg(b-k_m)\right],
\label{eq:pj_value}
\end{equation}
where $\iota(a,u)$ is the $b$-bit index formed from $a$ and $u$. With $\tau_{m,i}$ the target of mode $m$ entering the block, the exact cost of a candidate is computed by processing the columns in order, as in GPTQ:
\begin{align}
\tilde\tau_{m,i}&=\tau_{m,i}-\sum_{l<i}e_{m,l}\,R_{li}, &
u_i&=\arg\min_{u}\sum_m \lambda_m\big(\tilde\tau_{m,i}-q_{m,i}(a_i(s),u)\big)^2,\label{eq:pj_fb}\\
e_{m,i}&=\frac{\tilde\tau_{m,i}-q_{m,i}(a_i(s),u_i)}{R_{ii}}, &
\mathrm{cost}(s)&=\sum_m \lambda_m\sum_{i=1}^{n}e_{m,i}^2.\label{eq:pj_cost}
\end{align}
Because of the error feedback in Eq.~\eqref{eq:pj_fb}, the columns must be processed in order, so evaluating one candidate is costly. U planes of the same storage need only a few comparisons per column, but the $16^{n_{\mathrm{H}}}$ candidates of H planes, about $1.7\times10^7$ per block for UHHHHHH, cannot all be evaluated.

We reduce the candidates with two approximations. First, without the error feedback inside the block, the columns become independent once a candidate is fixed, and the error of column $i$ depends only on its H bits $a$. We compute once, for every $(i,a)$, the error with the best U bits,
\begin{equation}
D[i,a]=\min_{u}\ \frac{1}{R_{ii}^{2}}\sum_m \lambda_m\big(\tau_{m,i}-q_{m,i}(a,u)\big)^2,
\label{eq:pj_D}
\end{equation}
and the approximate cost of a candidate is then a sum of eight entries,
\begin{equation}
\widehat{\mathrm{cost}}(s)=\sum_{i=1}^{n}D\big[i,a_i(s)\big]=\big[\mathrm{vec}(D)^{\top}Z\big]_s,
\label{eq:pj_approx}
\end{equation}
where $Z\in\{0,1\}^{(n\,2^{n_{\mathrm{H}}})\times16^{n_{\mathrm{H}}}}$ is a fixed matrix that marks the eight entries of each candidate, so one matrix product scores all candidates. Only the $K$ candidates with the lowest approximate cost are evaluated exactly, and the best is taken:
\begin{equation}
s^\ast=\arg\min_{s\in\mathcal{S}}\ \mathrm{cost}(s),\qquad |\mathcal{S}|=K.
\label{eq:pj_sel}
\end{equation}
The number of exact evaluations is thus fixed at $K$, but scoring still enumerates all $16^{n_{\mathrm{H}}}$ candidates (Table~\ref{tab:stage_cost}).

Second, we use the prefix structure: the error of a mode depends only on the codewords of the planes it reads. If mode $m$ reads the first $p_m$ H planes, then for a partial candidate $s_{\le p}$ that fixes the first $p$ codewords, the approximate errors of the modes with $p_m\le p$ are already determined. Defining with these modes
\begin{equation}
D_p[i,a]=\min_{u}\ \frac{1}{R_{ii}^{2}}\sum_{m:\,p_m\le p}\lambda_m\big(\tau_{m,i}-q_{m,i}(a,u)\big)^2,\qquad a\in\{0,1\}^{p},
\label{eq:pj_Dj}
\end{equation}
the approximate cost of a partial candidate takes the same form,
\begin{equation}
\widehat{\mathrm{cost}}_p(s_{\le p})=\sum_{i=1}^{n}D_p\big[i,a_i^{(p)}(s_{\le p})\big].
\label{eq:pj_partial}
\end{equation}
At each step, the surviving partial candidates are extended by the 16 codewords of the next H plane, and only the $B$ with the lowest cost are kept:
\begin{equation}
\mathcal{S}_p=\operatorname*{top\text{-}B}_{(s,\,s_p)}\Big\{\widehat{\mathrm{cost}}_p\big((s,s_p)\big)\ :\ s\in\mathcal{S}_{p-1},\ s_p\in\mathcal{C}\Big\},\qquad \mathcal{S}_0=\{\varnothing\}.
\label{eq:pj_beam}
\end{equation}
Before the first mode is complete, all partial candidates have the same cost, so the beam can rank them only from the plane that completes a mode. In UHHHHHH, the 2-bit mode is complete at the second H plane, and the 16 candidates of the first H plane all fit in the beam. The cost at the last step equals Eq.~\eqref{eq:pj_approx}, so the $K$ lowest of $\mathcal{S}_{n_{\mathrm{H}}}$ form the set $\mathcal{S}$ of Eq.~\eqref{eq:pj_sel} and are evaluated exactly. This is Algorithm~\ref{alg:block}; unless noted otherwise, all encodings in this paper use it with $B=K=64$. The number of ranked partial candidates is $16B$ per plane, linear in $n_{\mathrm{H}}$ (Table~\ref{tab:stage_cost}), and the memory scales with $B$. With $B\ge16^{n_{\mathrm{H}}-1}$, the ranking equals that of Eq.~\eqref{eq:pj_approx}, and a smaller $B$ approaches fixing the planes one at a time, so $B$ trades encoding cost for quality (Appendix~\ref{app:search}).

\begin{table}[h]
\caption{Candidates ranked per block by full enumeration and by the prefix beam ($B=64$). Both evaluate $K=64$ candidates exactly.}
\label{tab:stage_cost}
\centering
\small
\begin{tabular}{rrr}
\toprule
$n_{\mathrm{H}}$ & Full enumeration ($16^{n_{\mathrm{H}}}$) & Prefix beam ($\le16Bn_{\mathrm{H}}$) \\
\midrule
4 & 65,536 & 4,096 \\
5 & 1,048,576 & 5,120 \\
6 & 16,777,216 & 6,144 \\
\bottomrule
\end{tabular}
\end{table}

\subsection{Candidate Search}
\label{app:search}

To check that the two approximations of Appendix~\ref{app:planejoint} do not hurt the encoding, we first compare the searches on a layout with four H planes ($n_{\mathrm{H}}=4$, UHHHHU), the largest case in which every candidate can still be evaluated exactly with error feedback. We use 12 matrices of Qwen2.5-3B (the k, o, gate, and down projections of layers 3, 18, and 33) with 128 random rows per matrix, their Hessians, and equal mode weights (Table~\ref{tab:beam_proxy}). Full enumeration followed by exact evaluation of the top $K$ reaches the same loss as evaluating all candidates in every mode. The prefix beam approaches that loss as it widens: at $B=64$, every mode is within 0.7\% of evaluating all candidates in less than half the time, whereas $B=16$ is faster but 2.6--3.0\% worse at 3 and 4 bits. Narrow beams reach a slightly lower loss at 2 bits because the objective is a weighted sum over modes, so sacrificing the higher modes can reduce the error of the lower ones. Encoding the full model with full enumeration and with the beam of width 64 gives nearly the same PPL.

\begin{table}[h]
\caption{Candidate search for UHHHHU ($n_{\mathrm{H}}=4$, Qwen2.5-3B): proxy loss relative to evaluating all candidates (12 matrices, geometric mean) and WikiText-2 PPL. Time is relative to evaluating all candidates.}
\label{tab:beam_proxy}
\centering
\small
\begin{tabular}{l cccc c}
\toprule
 & 2 & 2.5 & 3 & 4 & Time \\
\midrule
\multicolumn{6}{l}{Proxy loss (relative to all candidates)} \\
All candidates & 1.000 & 1.000 & 1.000 & 1.000 & 1.00$\times$ \\
Full enumeration, top-$K$ & 1.000 & 1.000 & 1.000 & 1.000 & 0.99$\times$ \\
Prefix beam, $B=16$  & 0.991 & 1.012 & 1.026 & 1.030 & 0.27$\times$ \\
Prefix beam, $B=32$  & 0.996 & 1.006 & 1.013 & 1.017 & 0.32$\times$ \\
\rowcolor[HTML]{DCEBFA} Prefix beam, $B=64$ (default) & 0.999 & 1.003 & 1.007 & 1.004 & 0.47$\times$ \\
Prefix beam, $B=128$ & 1.000 & 1.000 & 1.002 & 1.003 & 0.78$\times$ \\
Prefix beam, $B=256$ & 1.000 & 1.000 & 1.001 & 1.002 & 1.30$\times$ \\
\midrule
\multicolumn{6}{l}{WikiText-2 PPL} \\
Full enumeration, top-$K$ & 15.01 & 10.74 & 9.37 & 8.50 & \\
\rowcolor[HTML]{DCEBFA} Prefix beam, $B=64$ (default) & 15.08 & 10.77 & 9.39 & 8.53 & \\
\bottomrule
\end{tabular}
\end{table}

With $n_{\mathrm{H}}=4$ the candidates are few, so wide beams take as long as evaluating all candidates (1.30$\times$ at $B=256$); the beam pays off for $n_{\mathrm{H}}\ge5$, where the cost of the other searches grows rapidly (Table~\ref{tab:stage_cost}). For our layout UHHHHHH ($n_{\mathrm{H}}=6$), neither evaluating nor ranking all candidates is feasible, so the beam of width 256 serves as the reference (Table~\ref{tab:beam6}). $B=64$ stays within 2.2\% of it in every mode at 0.29$\times$ the search time, whereas $B=16$ worsens toward the higher modes, by about 10\% at 4 bits. The PPL of fully encoded models differs much less: all beam widths from 16 to 256 are within 0.2 of each other at every bit width. We therefore set both the beam width $B$ and the number of exactly evaluated candidates $K$ to 64.

\begin{table*}[t]
\caption{Beam width for UHHHHHH ($n_{\mathrm{H}}=6$, Qwen2.5-3B): proxy loss relative to $B=256$ (252 matrices), WikiText-2 PPL, and the time of the candidate search on the 12 matrices of Table~\ref{tab:beam_proxy} relative to $B=256$. Best in bold, second best underlined.}
\label{tab:beam6}
\centering
\small
\begin{tabular}{l ccccc ccccc c}
\toprule
 & \multicolumn{5}{c}{Proxy loss} & \multicolumn{5}{c}{WikiText-2 PPL} & \\
\cmidrule(lr){2-6} \cmidrule(lr){7-11}
Beam width & 2 & 2.5 & 3 & 3.5 & 4 & 2 & 2.5 & 3 & 3.5 & 4 & Time \\
\midrule
$B=16$ & \textbf{0.948} & \textbf{0.992} & 1.046 & 1.085 & 1.104 & \textbf{15.29} & 10.94 & 9.38 & 8.78 & 8.50 & 0.17$\times$ \\
\rowcolor[HTML]{DCEBFA} $B=64$ (default) & \underline{0.989} & \underline{0.997} & \underline{1.011} & \underline{1.019} & \underline{1.022} & 15.41 & \textbf{10.78} & \textbf{9.29} & \textbf{8.73} & \textbf{8.45} & 0.29$\times$ \\
$B=256$ & 1.000 & 1.000 & \textbf{1.000} & \textbf{1.000} & \textbf{1.000} & \underline{15.31} & \underline{10.83} & \underline{9.32} & \textbf{8.73} & \underline{8.46} & 1.00$\times$ \\
\bottomrule
\end{tabular}
\end{table*}

\subsection{Mode Weights}
\label{app:weights}

The mode weights $\lambda_m$ set how much the error of each mode counts in the ladder objective $\sum_m \lambda_m\,\mathrm{tr}(E_m \Gamma E_m^{\top})$ (Eq.~\eqref{eq:joint_objective}). We compare our weights 1-1-2-3-4 with seven alternatives: uniform weights and settings that put more weight on the lower or the higher modes (Table~\ref{tab:weights}). We measure the proxy loss on 63 matrices of Qwen2.5-3B (layers 3, 7, \dots, 35) and the PPL after encoding the full model. Shifting weight toward the lower modes improves the 2-bit mode at the cost of the higher modes, and shifting it toward the higher modes does the opposite. No setting improves all modes at once: each of the seven alternatives has a higher proxy loss than the default in at least one mode, and the PPL shows the same trade-off. Our weights 1-1-2-3-4 are therefore not a unique optimum but a middle point that gives up some accuracy at 2 bits to protect 3 bits and above, where they reach the lowest PPL (within 0.02 of 1-2-3-5-7). Since the weights can also be read as how often each mode is used, a deployment that knows how often each precision is used can set them accordingly before encoding.

\begin{table*}[h]
\caption{Mode weights (Qwen2.5-3B): proxy loss relative to the default (63 matrices) and WikiText-2 PPL (FP16 8.03). Best in bold, second best underlined.}
\label{tab:weights}
\centering
\small
\begin{tabular}{l ccccc ccccc}
\toprule
 & \multicolumn{5}{c}{Proxy loss} & \multicolumn{5}{c}{WikiText-2 PPL} \\
\cmidrule(lr){2-6} \cmidrule(lr){7-11}
Weights & 2 & 2.5 & 3 & 3.5 & 4 & 2 & 2.5 & 3 & 3.5 & 4 \\
\midrule
\rowcolor[HTML]{DCEBFA} 1-1-2-3-4 (default) & 1.000 & 1.000 & 1.000 & \underline{1.000} & 1.000 & 15.41 & 10.78 & \textbf{9.29} & \textbf{8.73} & \textbf{8.45} \\
1-1-1-1-1 (uniform) & 0.918 & \underline{0.925} & 1.024 & 1.111 & 1.170 & \textbf{14.60} & \underline{10.73} & 9.45 & 8.88 & 8.56 \\
1-1-2-2-2 & 0.970 & 0.966 & \textbf{0.975} & 1.033 & 1.081 & 15.19 & 10.85 & \underline{9.36} & 8.78 & 8.50 \\
1-2-3-5-7 & 1.069 & 0.974 & \underline{0.994} & \textbf{0.980} & \textbf{0.973} & 15.70 & \textbf{10.69} & \textbf{9.29} & \underline{8.75} & \underline{8.47} \\
1-1-1-1-4 & 0.948 & 0.960 & 1.065 & 1.156 & \underline{0.997} & 15.32 & 10.82 & 9.39 & 8.85 & 8.48 \\
3-2-1-1-1 & \underline{0.862} & 0.928 & 1.118 & 1.245 & 1.324 & 15.17 & 11.05 & 9.75 & 9.01 & 8.67 \\
4-1-1-1-1 & \textbf{0.842} & 0.986 & 1.148 & 1.269 & 1.350 & 15.26 & 11.22 & 9.75 & 9.06 & 8.67 \\
7-5-3-2-1 & 0.864 & \textbf{0.922} & 1.096 & 1.259 & 1.413 & \underline{14.75} & 11.07 & 9.73 & 9.07 & 8.75 \\
\bottomrule
\end{tabular}
\end{table*}

\section{Analysis Details} 
\label{app:analysis}

\subsection{Measurement Setup and the Loss Decomposition}
\label{app:an_setup}
\label{app:an_decomp}

All measurements in the analysis use the linear-layer matrices of Qwen2.5-3B~\citep{qwen2025qwen25}. Its 36 layers have 7 projections each, 252 matrices in total; we exclude 6 MLP matrices of early layers (up in layer 1, gate/up/down in layer 2, and gate/up in layer 3, with layers indexed from 0), for which the LUT refit of the coded ladder became numerically singular for some rows, and use the remaining 246. The Hessians are computed on the FP16 model with the same C4~\citep{raffel2020c4} calibration data as in our experiments (Appendix~\ref{app:setup}) and normalized to a mean diagonal of 1; to bound the computation, we encode the first 256 rows of each matrix. The loss is the relative proxy loss $L_m$ of Eq.~\eqref{eq:proxy_loss}, always evaluated with the full $\Gamma$ regardless of how the matrix was encoded. Across matrices we report means of dB values and medians of ratios.

The encoder uses the same beam search ($B=K=64$) and refit ($N_{\mathrm{ref}}=4$) as in our experiments, with damping $10^{-8}$ instead of 0.01 so that the decomposition below holds almost exactly for the evaluation $\Gamma$. For nested encoding, the mode weights match the integer modes of the two ladders: 1-2-4 for the 2/3/4-bit modes of the scalar ladder (UUUU), and 1-0.02-2-0.02-4 for the coded ladder (UHHHHHH), which nearly switches off the half-bit modes. A single-mode encoding encodes only that bit width; for the 3- and 4-bit coded structures, we read that mode from a ladder whose other modes have weight 0.02. An encoding without error feedback uses only the diagonal of $\Gamma$ during encoding. The nesting cost of MatGPTQ (MG)~\citep{kleinegger2026matgptq} is measured on the MG (2/3/4) checkpoint of our main results, with the same rows and $\Gamma$. This checkpoint gives the three modes equal weights (1-1-1), because 1-2-4, which puts more weight on 4 bits, makes the 2-bit mode collapse (Appendix~\ref{app:an_mg348}); for comparison, we also measure a checkpoint requantized with weights 1-2-4 by the same recipe. The reference is the loss of quantizing only that bit width with standard GPTQ~\citep{frantar2023gptq} (no \texttt{act\_order}, damping 0.01) on the grid of MG (asymmetric uniform grid with group size 128 and clip search).

We now derive Eq.~\eqref{eq:importance}. Consider one row with weights $w\in\mathbb{R}^c$, quantized values $\widehat w$, and error $E_r=w-\widehat w$, and let $R$ be the upper-triangular factor with $\Gamma^{-1}=R^{\top}R$ (Appendix~\ref{app:background}). Error feedback sets the target of column $j$ and its quantization error to
\begin{equation}
\tilde\tau_j \;=\; w_j-\sum_{i<j}\frac{R_{ij}}{R_{ii}}\,\varepsilon_i,
\qquad
\varepsilon_j \;=\; \tilde\tau_j-\widehat w_j ,
\label{eq:an_rule}
\end{equation}
which is the update of Appendix~\ref{app:background} with $e_i=\varepsilon_i/R_{ii}$; passing the errors of a block at once gives the same targets. Solving Eq.~\eqref{eq:an_rule} for $E_r$ gives, in row-vector form,
\begin{equation}
E_r \;=\; \varepsilon\,D_R^{-1}R,
\qquad D_R=\mathrm{diag}(R_{11},\dots,R_{cc}),
\label{eq:an_rowloss}
\end{equation}
since $(D_R^{-1}R)_{ij}=R_{ij}/R_{ii}$ for $i\le j$. With $R\,\Gamma R^{\top}=I$, the row loss becomes
\begin{equation}
E_r\Gamma E_r^{\top}
\;=\; \varepsilon\,D_R^{-1}\big(R\,\Gamma R^{\top}\big)D_R^{-1}\varepsilon^{\top}
\;=\; \sum_j d_j\,\varepsilon_j^2,
\qquad d_j=\frac{1}{R_{jj}^2}.
\label{eq:an_diag}
\end{equation}
Because $\Gamma=R^{-1}R^{-\top}$ with $R^{-1}$ upper triangular, $1/R_{jj}^2$ is the Schur complement of $\Gamma_{>j,>j}$ in $\Gamma_{\ge j,\ge j}$, which is the expression for $d_j$ in Eq.~\eqref{eq:importance}, and hence
\begin{equation}
d_j \;=\; \Gamma_{jj}-\Gamma_{j,>j}\,\Gamma_{>j,>j}^{-1}\,\Gamma_{>j,j} \;\le\; \Gamma_{jj}.
\label{eq:an_schur}
\end{equation}
Eq.~\eqref{eq:an_diag} holds for any choice of $\widehat w$, so whether columns are chosen one at a time or eight at a time as a block, as in an H plane, reducing the layer loss amounts to choosing, in each column or block, points close to the targets $\tilde\tau_j$ weighted by $d_j$. It is exact for the $\Gamma$ used during encoding; with damping, it is approximate for the evaluation $\Gamma$. Viewing $\Gamma$ as the second-moment matrix of the inputs, $d_j$ is the variance of the input of column $j$ that the inputs of later columns do not explain.

Without error feedback, $\tilde\tau_j=w_j$ and $E_r=\varepsilon$, so $E_r\Gamma E_r^{\top}=\sum_{j,k}\Gamma_{jk}\varepsilon_j\varepsilon_k$. If the errors of different columns have zero expected product and the same second moment $\sigma^2$ in every column and in both cases, the expected row losses without and with feedback are
\begin{equation}
\mathbb{E}\big[E_r\Gamma E_r^{\top}\big]_{\mathrm{no\,FB}} \;=\; \sigma^2\sum_j\Gamma_{jj},
\qquad
\mathbb{E}\big[E_r\Gamma E_r^{\top}\big]_{\mathrm{FB}} \;=\; \sigma^2\sum_j d_j,
\label{eq:an_fb_expect}
\end{equation}
whose ratio in dB is Eq.~\eqref{eq:feedback_gain}.

\subsection{Proof of Theorem~\ref{thm:uniform} and the Coded Gain on Real Matrices}
\label{app:an_proof}

Let the values $y = (y_1,\dots,y_8)$ of the 8 columns of a block be exchangeable, and let both structures map an index $\iota$ to the reconstruction value $T(\iota)$ with one LUT $T$ per row. Let $\mathcal I$ be the set of index combinations that a block of the coded structure can choose: the bits of a U plane are free in every column, and the 8 bits of each H plane must form a codeword of the extended Hamming code $\mathcal C = [8,4,4]$~\citep{conway1999splag}, which is the Reed--Muller code RM(1,3). For a LUT $T$ and importance $d$, the weighted distortion of the coded structure and its optimum are
\begin{equation}
D_T(d) = \mathbb{E}_y\Big[\min_{\iota\in\mathcal I} \sum_j d_j\,\big(y_j - T(\iota_j)\big)^2\Big],
\qquad
D_{\mathrm{coded}}^*(d) = \inf_T D_T(d),
\label{eq:an_dist}
\end{equation}
so that $D_{\mathrm{c}}=D_{\mathrm{coded}}^*(\mathbf 1)/8$. Let $\mathcal A=\mathrm{Aut}(\mathcal C)=\mathrm{AGL}(3,2)$ be the automorphism group of the code, acting on $d$ by permuting the column positions.

\begin{proof}[Proof of Theorem~\ref{thm:uniform}]
For fixed $T$ and $y$, the expression inside the expectation in Eq.~\eqref{eq:an_dist} is a minimum of functions linear in $d$, and taking the expectation and the infimum over $T$ preserves concavity, so $D_{\mathrm{coded}}^*$ is concave in $d$. It is also homogeneous of degree one in $d$. Since every H plane uses the same code and the U planes are unconstrained, each $\pi\in\mathcal A$ maps $\mathcal I$ onto itself; as $y$ is exchangeable and the LUT is shared by all columns,
\begin{equation}
D_{\mathrm{coded}}^*(\pi d)=D_{\mathrm{coded}}^*(d)\qquad\text{for all }\pi\in\mathcal A.
\label{eq:an_invariance}
\end{equation}
Because $\mathcal A$ acts transitively on the 8 positions, every entry of $\bar d=|\mathcal A|^{-1}\sum_{\pi\in\mathcal A}\pi d$ equals $\tfrac18\sum_j d_j$. Eq.~\eqref{eq:an_invariance}, Jensen's inequality, and homogeneity then give
\begin{equation}
D_{\mathrm{coded}}^*(d) \;=\; \frac{1}{|\mathcal A|}\sum_{\pi\in\mathcal A}D_{\mathrm{coded}}^*(\pi d) \;\le\; D_{\mathrm{coded}}^*(\bar d) \;=\; \frac{\sum_j d_j}{8}\,D_{\mathrm{coded}}^*(\mathbf 1) \;=\; D_{\mathrm{c}}\sum_j d_j.
\label{eq:an_jensen}
\end{equation}
In the scalar structure, once $T$ is fixed, the nearest point of each column does not depend on $d_j$, so with $Q_T$ the nearest-point quantizer of $T$,
\begin{equation}
D_{\mathrm{scalar}}^*(d) \;=\; \inf_T \sum_j d_j\,\mathbb{E}\big[(y_j-Q_T(y_j))^2\big] \;=\; D_{\mathrm{LM}}\sum_j d_j,
\end{equation}
where $D_{\mathrm{LM}}$ is the Lloyd--Max distortion of the marginal distribution of a column.
\end{proof}

The proof uses neither the bit rate nor the LUT size, so it applies equally to UHHHH and UHHHHHH. It remains to check whether $D_{\mathrm{c}} < D_{\mathrm{LM}}$ at each bit width. For any fixed LUT $T$ and any index choice $\iota(y)\in\mathcal I$,
\begin{equation}
D_{\mathrm{c}} \;\le\; \frac18\,\mathbb{E}_y\Big[\sum_j \big(y_j - T(\iota_j(y))\big)^2\Big],
\label{eq:an_cert}
\end{equation}
so any encoding with a fixed LUT bounds $D_{\mathrm{c}}$ from above. We learn the row LUTs with our encoder on i.i.d.\ Gaussian matrices with 2048 columns, fix them, and quantize 512 freshly drawn rows with an exact index search (the U bits separate across columns, and all combinations of the H planes are enumerated). The resulting distortion is lower than Lloyd--Max by 0.61\,dB at 2 bits and 0.44\,dB at 3 bits, with more than $10^6$ fresh samples and a standard error below $10^{-4}$ (Table~\ref{tab:an_code}). This is the gain of placing the reconstruction points of eight columns jointly, known from vector quantization: applying Construction A to the same code gives the $E_8$ lattice, on which QuIP\#~\citep{tseng2024quipsharp} builds its codebook, and QTIP~\citep{tseng2024qtip} uses a trellis code; here the gain comes without a codebook. At 4 bits (UHHHHHH), the $16^6$ combinations of the H planes make an exact search infeasible, and the in-sample distortion, which converges as the number of columns grows, ends 0.35\,dB above Lloyd--Max; we therefore cannot claim that the coded structure wins at 4 bits under uniform importance. Its 4-bit gain on real matrices comes from allocation, discussed next.

\begin{table}[h]
\caption{Single-mode gain of the coded structure over the scalar one at equal rate (dB). Gaussian: uniform importance, certified against Lloyd--Max. Qwen2.5-3B: mean over matrices (matrices won).}
\label{tab:an_code}
\centering
\small
\begin{tabular}{lccc}
\toprule
 & 2-bit & 3-bit & 4-bit \\
\midrule
Gaussian, uniform importance & $+0.61$ & $+0.44$ & not certified \\
Qwen2.5-3B & $+1.84$ (246) & $+1.69$ (242) & $+1.31$ (210) \\
\bottomrule
\end{tabular}
\end{table}

\begin{figure}[t]
\centering
\includegraphics[width=\textwidth]{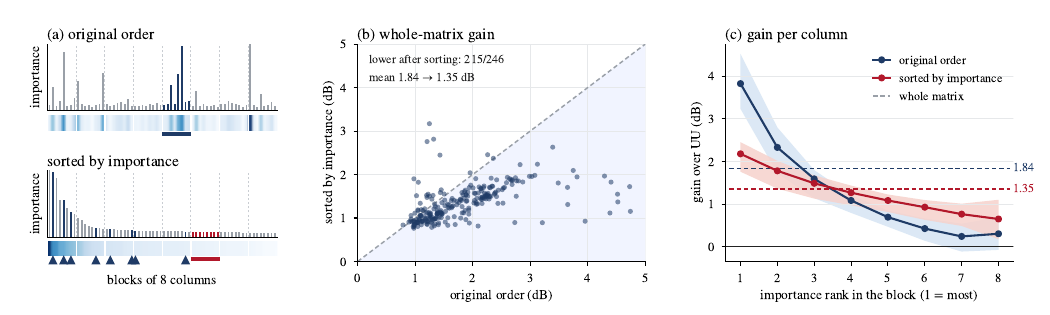}
\caption{Allocation gain of the coded plane (Qwen2.5-3B, 2-bit). (a) Importance $d_j$ of 64 columns, original and sorted. (b) Whole-matrix gain before and after sorting columns by importance. (c) Gain of UHH over UU by in-block importance rank in the original and sorted column order (mean; band: interquartile range); dashed lines: whole matrix.}
\label{fig:an_alloc_full}
\end{figure}

On real matrices, the gain concentrates on the important columns of each block (Figure~\ref{fig:an_alloc_full}c): the most important column gains the most, and the gain shrinks with rank. Sorting the columns by importance, which reduces the differences within a block, flattens the gain across ranks and lowers the overall gain from 1.84\,dB to 1.35\,dB.

\begin{figure}[t]
\centering
\includegraphics[width=\textwidth]{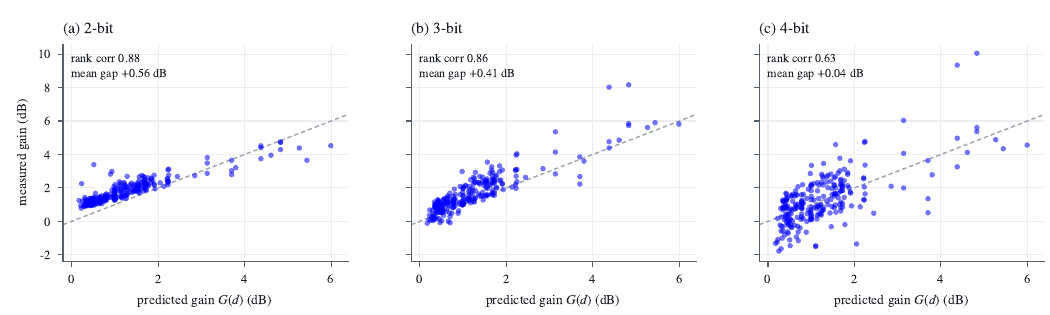}
\caption{Single-mode gain of the coded structure over the scalar one, measured vs.\ predicted by the weighted high-resolution gain $\hat G(d)$ (no fitted parameter), 246 matrices of Qwen2.5-3B. Dashed: $y=x$.}
\label{fig:an_code_pred}
\end{figure}

The allocation gain can be predicted with a high-resolution approximation. With a uniform LUT, the reconstruction points of UHH form the set $\Lambda = \{\mathbf c_1 + 2\mathbf c_2 + 4\mathbf z : \mathbf c_1,\mathbf c_2\in\mathcal C,\ \mathbf z\in\mathbb Z^8\}$. This set, given by the code formula, is a lattice only if the Schur square of $\mathcal C$ lies in $\mathcal C$~\citep{kositwattanarerk2014construction}, but the Schur square of RM(1,3) is the $[8,7]$ code RM(2,3), so $\Lambda$ is not a lattice. It is still periodic with period $4\mathbb Z^8$, so, treating the source at high resolution as uniform over one period, we can compute the per-column distortion $D_\Lambda(d)$ of weighted nearest-point quantization. Comparing it with the per-column distortion $\bar d/3$ of the cubic lattice $2\mathbb Z^8$ (UU) at the same point density, where $\bar d = \tfrac18\sum_j d_j$ here denotes the mean importance of the block, gives the predicted gain
\begin{equation}
\hat G(d) = 10\log_{10}\frac{\bar d/3}{D_\Lambda(d)},
\label{eq:an_G}
\end{equation}
which we compute per block from the importance of the matrix and average with the loss share of each block; nothing is fitted. The prediction tracks the measured gain per matrix with rank correlations 0.88, 0.86, and 0.63 at 2, 3, and 4 bits (Figure~\ref{fig:an_code_pred}). The measured gain exceeds the prediction by 0.56, 0.41, and 0.04\,dB on average, close to the Gaussian values above (0.61 and 0.44\,dB, and none certified at 4 bits), consistent with attributing the difference to the low-rate shape gain that the high-resolution prediction does not capture. Although the prediction is computed once for UHH and reused at 3 and 4 bits, it still orders the matrices well, so allocation explains much of the variation in gain across matrices even at higher bit widths.

\subsection{Feedback Gain, Remark~\ref{rem:nest}, and Proposition~\ref{prop:meanprop}}
\label{app:an_fbgain}
\label{app:an_nestproof}
\label{app:an_meanprop}

We measure the feedback gain by encoding the scalar structure in a single mode twice, once with the full $\Gamma$ (with feedback) and once with only the diagonal of $\Gamma$ (without feedback), and evaluating both with the full $\Gamma$. The gain averages about 2.75\,dB at all three bit widths. At 4 bits, the measured gain per matrix follows the prediction of Eq.~\eqref{eq:feedback_gain} with rank correlation 0.70 (Figure~\ref{fig:an_fb_nest}a), but the prediction is larger by 1.1\,dB on average; since it assumes the same per-column error in both cases (Eq.~\eqref{eq:an_fb_expect}), it should be read as predicting the ordering across matrices rather than the size.

Let $L_m^{\mathrm{single},*}$ denote the loss of the optimal single-mode encoding of mode $m$ with its own LUT.

\begin{remark}
\label{rem:nest}
(i) Without feedback and with exactly nested grids, $c_m = 0$ for every mode. (ii) With or without feedback, $c_m \ge 0$ when $L_m^{\mathrm{single}}=L_m^{\mathrm{single},*}$.
\end{remark}

\begin{proof}
(i) Without feedback, $\tilde\tau_{m,j}=w_j$ for every mode, so with exactly nested grids the nearest point of each mode is a child of the nearest point of the mode below, and the nested encoding attains the single-mode loss of every mode. (ii) Decoding only mode $m$ from a nested encoding is a valid single-mode encoding with the same LUT, even though the error trajectories of the two encodings differ, so
\begin{equation}
L_m^{\mathrm{nest}} \;\ge\; L_m^{\mathrm{single},*}
\quad\Longrightarrow\quad
c_m \;=\; 10\log_{10}\frac{L_m^{\mathrm{nest}}}{L_m^{\mathrm{single},*}} \;\ge\; 0.
\end{equation}
\end{proof}

The measured $c_m$, however, is relative to a single-mode encoding found by the same search with its own LUT rather than to the optimum, so it can be negative on individual matrices; the interquartile range of the 4-bit cost of the scalar ladder without feedback ($-0.47$ to $0.45$\,dB) indicates the size of these deviations.

For Proposition~\ref{prop:meanprop}, fix column $j$ and the targets $\tilde\tau_{m,j}$, and let $\mathcal N$ be the set of value tuples $q=(q_m)_m$ that satisfy the nesting constraint. The weighted column loss is
\begin{equation}
\ell_j(q) \;=\; d_j\sum_m \lambda_m\,\big(\tilde\tau_{m,j}-q_m\big)^2 .
\label{eq:an_colloss}
\end{equation}
Per-mode targets choose $q^{\mathrm{pm}}=\arg\min_{q\in\mathcal N}\ell_j(q)$, whereas mean propagation chooses the nested values $q^{\mathrm{mp}}\in\mathcal N$ nearest to the shared target $\bar\tau_j$.

\begin{proof}[Proof of Proposition~\ref{prop:meanprop}]
(i) Since $q^{\mathrm{mp}}\in\mathcal N$, it is a candidate of the minimization that defines $q^{\mathrm{pm}}$, so $\ell_j(q^{\mathrm{pm}})\le\ell_j(q^{\mathrm{mp}})$.

(ii) Write the error of mode $m$ under mean propagation as
\begin{equation}
\tilde\tau_{m,j}-q^{\mathrm{mp}}_m \;=\; \big(\tilde\tau_{m,j}-\bar\tau_j\big)+\eta_{m,j},
\qquad \eta_{m,j}=\bar\tau_j-q^{\mathrm{mp}}_m .
\label{eq:an_mp_split}
\end{equation}
Since $\eta_{m,j}$ has zero mean and is uncorrelated with $\tilde\tau_{m,j}-\bar\tau_j$, the cross term vanishes in expectation, and
\begin{equation}
\mathbb{E}\big[(\tilde\tau_{m,j}-q^{\mathrm{mp}}_m)^2\big] \;=\; \mathbb{E}\big[(\tilde\tau_{m,j}-\bar\tau_j)^2\big]+\mathbb{E}\big[\eta_{m,j}^2\big],
\label{eq:an_mp_expect}
\end{equation}
where, by assumption, the second term equals the second moment of the error of quantizing mode $m$ at its own target. The expected excess of $\ell_j$ is therefore the weighted sum of the first term, and with $S=\sum_m\lambda_m$ the weighted variance identity gives
\begin{equation}
\Delta_j \;=\; d_j\sum_m \lambda_m\,\mathbb{E}\big[(\tilde\tau_{m,j}-\bar\tau_j)^2\big]
\;=\; \frac{d_j}{S}\sum_{m<m'}\lambda_m\lambda_{m'}\,\mathbb{E}\big[(\tilde\tau_{m,j}-\tilde\tau_{m',j})^2\big],
\label{eq:an_mp_result}
\end{equation}
which is Eq.~\eqref{eq:meanprop_cost}.
\end{proof}

For two modes $m<m'$ with gap $g_j=\tilde\tau_{m',j}-\tilde\tau_{m,j}$, the shared target lies between the two targets,
\begin{equation}
\tilde\tau_{m,j}-\bar\tau_j \;=\; -\frac{\lambda_{m'}}{\lambda_m+\lambda_{m'}}\,g_j,
\qquad
\tilde\tau_{m',j}-\bar\tau_j \;=\; \frac{\lambda_m}{\lambda_m+\lambda_{m'}}\,g_j,
\label{eq:an_mp_dev}
\end{equation}
and Eq.~\eqref{eq:an_mp_result} reduces to $\Delta_j=d_j\,\frac{\lambda_m\lambda_{m'}}{\lambda_m+\lambda_{m'}}\,\mathbb{E}[g_j^2]$: the mode with the smaller weight is quantized farther from its own target.

In actual encoding, the error trajectories of the two rules differ, so the target gaps themselves differ, and we compare whole matrices by measurement. Changing only the target rule in the same encoder on the scalar ladder (weights 1-2-4), the sum over the three modes of the relative excess loss $(L_m^{\mathrm{nest}} - L_m^{\mathrm{single}})/L_m^{\mathrm{single}}$ is 1.41 times larger under mean propagation in the median (10--90\% range 1.32--1.60) and larger on 245 of the 246 matrices; per mode, mean propagation costs more on 246, 236, and 245 matrices at 2, 3, and 4 bits. On the coded ladder (weights 1-0.02-2-0.02-4, the same 1-2-4 on the integer modes) the sum is 1.42 times larger (239 of 246 matrices), but at the 3-bit mode alone mean propagation costs less on 226 matrices.

\subsection{How Large the Nesting Cost Is: MatGPTQ and the Coded Ladder}
\label{app:an_mg348}
\label{app:an_codecost}

\begin{figure}[h]
\centering
\includegraphics[width=\textwidth]{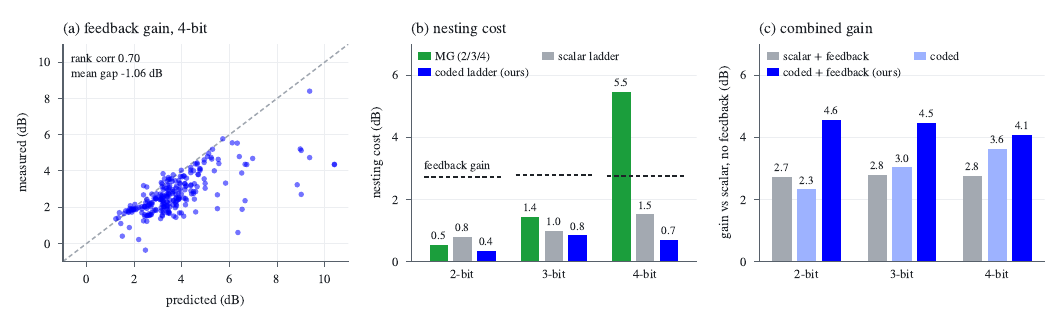}
\caption{Feedback and nesting, 246 matrices of Qwen2.5-3B. (a) Feedback gain at 4 bits, measured vs.\ predicted (dashed: $y=x$). (b) Nesting cost, mean; dashed: feedback gain. (c) Single-mode gain over the scalar structure without feedback, mean.}
\label{fig:an_fb_nest}
\end{figure}

Figure~\ref{fig:an_fb_nest}b compares the nesting costs of the three ladders. The 4-bit cost of MG (2/3/4) exceeds the feedback gain on 245 of the 246 matrices. Measured in the same way, the official ladder MG (3/4/8) has a large relative cost at 8 bits (23\,dB on average), but its 8-bit nested loss is only 0.80 times the 4-bit single-mode GPTQ loss of the same matrix (median), and its 4-bit cost drops to 1.9\,dB (from 5.45\,dB in MG (2/3/4)). Since the single-mode loss shrinks by about a factor of four per bit, the same relative cost means a small absolute loss at high bit widths. The cost has not vanished in this ladder; without a 2-bit mode, the target gaps between modes appear to be smaller.

The MG (2/3/4) checkpoint in our experiments gives the three modes equal weights (1-1-1). Requantizing with the same recipe and weights 1-2-4, which put more weight on 4 bits, lowers the 4-bit nesting cost from 5.45\,dB to 2.73\,dB and improves the 4-bit PPL from 9.58 to 8.83, but raises the 2-bit cost from 0.54\,dB to 1.61\,dB: the 2-bit layer loss grows by 1.07\,dB on average, on all 246 matrices, and the 2-bit PPL degrades from 25.9 to 42.8 (Table~\ref{tab:an_mgweights}). The shared target is pulled toward the heavily weighted 4-bit mode, so the 2-bit mode is quantized far from its own target (Eq.~\eqref{eq:an_mp_dev}). In absolute terms, the increase at 2 bits (relative proxy loss from 0.062 to 0.080) is about three times the decrease at 4 bits (from 0.013 to 0.007). Under a shared target, the weights cannot remove the cost but only shift it between modes, so we use 1-1-1, which balances all three modes including 2 bits, as the baseline. Even at the same weights 1-2-4, MG (2/3/4) costs more than the scalar ladder with per-mode targets at 2 and 4 bits (Table~\ref{tab:an_nest}), and on the same model Any-Rate reaches a lower PPL than both versions at all three bit widths.

\begin{table}[h]
\centering
\caption{MG (2/3/4) with mode weights 1-1-1 (the baseline in our results) and 1-2-4, and Any-Rate, Qwen2.5-3B: WikiText-2 PPL (FP16 8.03) and layer nesting cost (dB, mean over matrices), each against single-mode encoding of its own structure at the same bit width. Best in bold, second best underlined.}
\label{tab:an_mgweights}
\small
\begin{tabular}{@{}lcccccc@{}}
\toprule
 & \multicolumn{3}{c}{WikiText-2 PPL} & \multicolumn{3}{c}{Nesting cost (dB)} \\
\cmidrule(lr){2-4} \cmidrule(lr){5-7}
 & 2-bit & 3-bit & 4-bit & 2-bit & 3-bit & 4-bit \\
\midrule
MG (2/3/4), weights 1-1-1 & \underline{25.89} & 10.75 & 9.58 & \textbf{0.54} & 1.43 & 5.45 \\
MG (2/3/4), weights 1-2-4 & 42.77 & \underline{10.08} & \underline{8.83} & 1.61 & \underline{0.96} & \underline{2.73} \\
\rowcolor[HTML]{DCEBFA} AR (ours) & \textbf{15.41} & \textbf{9.29} & \textbf{8.45} & \underline{0.80} & \textbf{0.59} & \textbf{0.24} \\
\bottomrule
\end{tabular}
\end{table}

The nesting cost splits into a structural part, which remains without feedback, and a conflict part, which arises with feedback. With $c_m^{\mathrm{struct}}$ the nesting cost of mode $m$ measured without feedback, we define the conflict cost $\kappa_m$, in units of the single-mode loss, by
\begin{equation}
10^{c_m/10} \;=\; 10^{c_m^{\mathrm{struct}}/10} + \kappa_m,
\qquad\text{i.e.,}\qquad
\kappa_m \;=\; \frac{L_m^{\mathrm{nest}}}{L_m^{\mathrm{single}}}\bigg|_{\mathrm{FB}} - \frac{L_m^{\mathrm{nest}}}{L_m^{\mathrm{single}}}\bigg|_{\mathrm{no\,FB}} .
\label{eq:an_split}
\end{equation}
The structural cost of the coded ladder averages 0.11, 0.71, and 0.46\,dB at 2, 3, and 4 bits, while that of the scalar ladder is small (0.12, 0.23, and $-0.10$\,dB); it arises because the children of a coded plane do not split the cell of their parent exactly. For two modes $m<m'$ at small gaps, the excess loss from conflicts takes the form
\begin{equation}
\Delta^{\mathrm{conf}} \;\approx\; K\sigma_g^2,
\qquad
\sigma_g^2=\mathbb{E}\big[g^2\big],
\qquad
K \;\le\; \frac{\lambda_m\lambda_{m'}}{\lambda_m+\lambda_{m'}},
\label{eq:an_conf_model}
\end{equation}
where $\sigma_g^2$ is the importance-weighted mean squared gap between the targets of the two modes (Eq.~\eqref{eq:an_mp_dev}), and the bound on $K$ follows from parts (i) and (ii) of Proposition~\ref{prop:meanprop} under the assumptions of part (ii). The coded ladder moves the two factors in opposite directions. Solving the two-mode problem exactly per block for ladders one plane apart (UHH to UHHH for the coded ladder and UU to UUU for the scalar one, with equal weights), with the block targets and LUTs that the encoder actually saw, gives coded-to-scalar ratios of 0.36--0.53 for $\sigma_g^2$, 1.15--1.30 for $K$, and 0.46--0.64 for $\Delta^{\mathrm{conf}}$. Because of its computational cost, this block model was measured on 12 matrices (layers 0, 17, and 35, one projection for each of the four Hessians: q, o, gate, and down). On the 246 matrices, the coded-to-scalar ratio of $\kappa_m$ measured with the actual encoder is 0.33, 0.28, and 0.15 in the median, more favorable to the coded ladder than the block model.

Figure~\ref{fig:an_fb_nest}c shows gains over scalar single-mode encoding without feedback. Feedback alone gives about 2.75\,dB, and coded planes alone give 2.3--3.6\,dB. Together they give the largest gain, 4.1--4.6\,dB, but less than the sum of the two, and the shortfall grows with the bit width, reaching 2.3\,dB at 4 bits. The gain of coded planes also moves in opposite directions: without feedback it grows with the bit width (from 2.3 to 3.6\,dB), but with feedback it shrinks (from 1.84 to 1.31\,dB; Table~\ref{tab:an_code}). Since both mechanisms exploit the differences in importance across columns, they appear to share the same room at higher bit widths.

\begin{figure}[h]
\centering
\includegraphics[width=\textwidth]{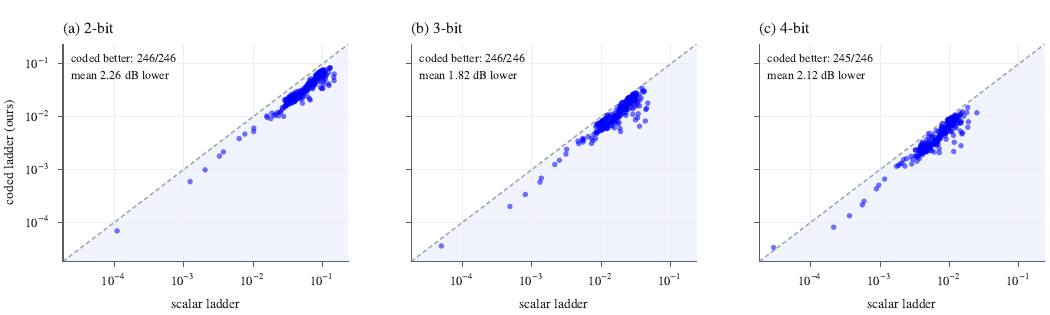}
\caption{Nested loss of the coded ladder (ours) vs.\ the scalar ladder on the same matrix, 246 matrices of Qwen2.5-3B, same encoder and integer-mode weights (log scale). Shaded: coded ladder lower. Dashed: $y=x$.}
\label{fig:an_nested}
\end{figure}

\subsection{Nested Ladders per Matrix and the Paper Ladder}
\label{app:an_paperladder}

What the model sees is the absolute loss of the nested ladder. Without relying on single-mode references, we therefore compare the two nested ladders on the same matrix directly, with the same encoder, rate, and integer-mode weights (Figure~\ref{fig:an_nested}). The coded ladder has a lower loss on almost every matrix, by 2.26, 1.82, and 2.12\,dB on average at 2, 3, and 4 bits. On each matrix, this difference splits as
\begin{equation}
10\log_{10}\frac{L_m^{\mathrm{nest,scalar}}}{L_m^{\mathrm{nest,coded}}}
\;=\; \underbrace{10\log_{10}\frac{L_m^{\mathrm{single,scalar}}}{L_m^{\mathrm{single,coded}}}}_{G_m^{\mathrm{single}}}
\;+\; \big(c_m^{\mathrm{scalar}}-c_m^{\mathrm{coded}}\big),
\label{eq:an_decomp_matrix}
\end{equation}
which follows from the definition of $c_m$ in Eq.~\eqref{eq:nesting_cost} and is Eq.~\eqref{eq:nested_gain} applied to a single matrix.

To compare integer modes only, the coded ladder above nearly switches off the half-bit modes with weight 0.02, whereas the ladder of our experiments uses all five modes (1-1-2-3-4). Since encoding is independent across rows, we read the first 256 rows of the stored paper encoding and compare them with the same single-mode references. With the half-bit modes switched on, the nesting cost changes from 0.35, 0.84, and 0.70\,dB to 0.80, 0.59, and 0.24\,dB at 2, 3, and 4 bits: the 2-bit mode pays more, apparently because the 2.5-bit mode with weight 1 directly above it now shares the conflict, while the 3- and 4-bit modes pay less, so the 4-bit cost of the paper ladder is far below that of MG (2/3/4) (5.45\,dB) and of the scalar ladder (1.51\,dB). At the model level, too, switching off the half-bit modes improves only the 2-bit PPL slightly and degrades the other modes (Table~\ref{tab:ablation}). The stored encoding, however, uses damping 0.01 and the analysis ladder $10^{-8}$, so this comparison also contains the effect of damping. Re-encoding the paper ladder with both dampings on 12 matrices (q/o/gate/down of layers 4, 18, and 33), the cost changes by less than 0.3\,dB on 10 of them; the exceptions are the gate and down projections of layer 4, whose Hessians are close to singular, with changes of 0.7 and 6.7\,dB at 4 bits. With damping fixed at $10^{-8}$, the paper ladder still pays more at 2 bits on all 12 matrices and less at 3 and 4 bits on 10 of them, and its 4-bit cost (0.71\,dB on average) stays well below those of the scalar ladder (1.47\,dB) and MG (2/3/4) (6.51\,dB) on the same matrices; its difference from the analysis ladder (0.75\,dB), however, is smaller than in the comparison with the stored encoding.

\section{Experimental Setup} 
\label{app:setup}
\label{app:hyper}

We quantize all decoder linear layers of Qwen2.5-3B (base)~\citep{qwen2025qwen25}, the language model of Gemma-3-4b-it~\citep{team2025gemma3}, Llama-3.1-8B-Instruct~\citep{grattafiori2024llama3}, and Phi-4~\citep{abdin2024phi4}, keeping the embeddings, \texttt{lm\_head}, and normalization layers in FP16; the fused qkv and gate\_up projections of Phi-4 are each quantized as one matrix. All baselines, Any-Precision LLM (AP)~\citep{park2024anyprecision}, AnyBCQ (AB)~\citep{park2026anybcq}, MatGPTQ (MG)~\citep{kleinegger2026matgptq}, and GPTQ~\citep{frantar2023gptq}, follow the authors' public code and official recipes, with only the calibration data unified to C4~\citep{raffel2020c4} ($100\times512$ tokens for AP, 512 samples for AB, and the same $128\times2048$ for GPTQ, MG, and Any-Rate). The two deviations are that AP lowers its seed precision from 3 to 2 bits to obtain a 2-bit mode, and that MG (2/3/4) is the official MG recipe with only the bit widths changed to 2/3/4 (4-bit master, weights 1-1-1). MG (3/4/8) reads its 2- and 6-bit modes by truncating its 8-bit master, and AB is trained for 2--4 bits only. Any-Rate uses the layout UHHHHHH (modes of 2, 2.5, 3, 3.5, and 4 bits), mode weights 1-1-2-3-4, $B=K=64$, $N_{\mathrm{ref}}=4$, and damping of 0.01 times the mean Hessian diagonal (Appendix~\ref{app:background}); the layout, the mode weights, and the beam width were chosen on the proxy loss of Qwen2.5-3B (Appendix~\ref{app:encoder}), all values are used for all models, and the residual planes for 5--8 bits are encoded after the ladder is fixed (Appendix~\ref{app:residual}).

We evaluate with lm-evaluation-harness~\citep{gao2024lmeval} 0.4.5: WikiText-2~\citep{merity2017wikitext} PPL on non-overlapping 2048-token windows; zero-shot \texttt{acc\_norm} on ARC-Challenge and ARC-Easy~\citep{clark2018arc}, HellaSwag~\citep{zellers2019hellaswag}, and PIQA~\citep{bisk2020piqa}, and \texttt{acc} on WinoGrande~\citep{sakaguchi2020winogrande}; 5-shot greedy flexible-extract accuracy on all 1,319 GSM8K~\citep{cobbe2021gsm8k} problems, with generations that continue after their first answer cut at that answer for every run; and 5-shot greedy exact match on the first 1,000 validation questions of TriviaQA~\citep{joshi2017triviaqa} and NQ-open~\citep{kwiatkowski2019nq}. The per-question outputs and grading of every value are in the released results.

\section{Storage and Decoding Efficiency} 
\label{app:storage}
\label{app:efficiency}

\subsection{Storage}

We measure the storage needed to serve all modes of each method from one checkpoint, in bits per weight over the 224 decoder linear layers of Llama-3.1-8B-Instruct~\citep{grattafiori2024llama3} (Table~\ref{tab:storage}). The count includes the codes and all metadata (scales, offsets, LUTs, and permutations) and excludes the FP16 embeddings, \texttt{lm\_head}, and normalization layers, which are the same for all methods. Each method uses its recommended configuration: Any-Precision LLM (AP)~\citep{park2024anyprecision} its 8-bit parent, MatGPTQ (MG)~\citep{kleinegger2026matgptq} its official recipe, MG (3/4/8), and AnyBCQ (AB)~\citep{park2026anybcq} one set of scales per supported bit width; AB was encoded only for 2--4 bits because of its encoding time, but its storage for 2--8 bits follows exactly from its format. All methods store about 8 bits of codes, but their metadata grows differently. Any-Rate keeps one small LUT per row, so its metadata scales with the number of rows and is independent of the group size and the number of modes. MG (3/4/8) keeps a scale and a zero point per group and shares its 8-bit master among all modes. AP keeps a LUT per row for every bit width, dominated by that of its 8-bit parent, and AB stores $b$ scales per group for every supported bit width $b$, so its metadata grows fastest with the number of modes. Serving 2--8 bits, Any-Rate thus needs about the same storage as MG (3/4/8) while providing nine modes, including two half-bit modes. Since the ladder (2--4 bits) and the residual (5--8 bits) are separate, serving only 2--4 bits needs the ladder alone, 4.21 bits per weight. The 2- and 6-bit modes of MG (3/4/8) are read by truncating its 8-bit master and are not optimized.

\begin{table}[h]
\centering
\caption{Storage of one checkpoint serving all modes of each method (Llama-3.1-8B-Instruct, 224 linear layers; codes and all metadata).}
\label{tab:storage}
\small
\setlength{\tabcolsep}{4pt}
\begin{tabular}{lccc}
\toprule
Method & Optimized bits & Modes & Bits/weight \\
\midrule
AP & 2--8 & 7 & 9.60 \\
AB & 2--8 & 7 & 13.38 \\
MG (3/4/8) & 3, 4, 8 & 7 (sliced) & \textbf{8.26} \\
\rowcolor[HTML]{DCEBFA}
AR (ours) & 2--4 (by 0.5), 5--8 & 9 & \underline{8.30} \\
\bottomrule
\end{tabular}
\end{table}

\subsection{Kernel Latency}

We time the batch-1 decoding GEMV on an RTX 5070 Ti (16\,GB, 48\,MB L2) with FP16 activations. As in actual decoding, where the whole model is read from memory for every token, we create copies of the weights totaling at least 320\,MB and call them in turn inside one CUDA graph, so that the weights come from memory rather than L2. Every combination of method and bit width is timed in five rounds, in a random order per round, and we report the median, which reduces the effect of clock fluctuations near the power limit. Since kernel time does not depend on the weight values, each method uses synthetic weights in its own format. The Any-Rate kernel assigns one lane to each row and keeps the mode LUT of the row in shared memory without bank conflicts. The 2- to 3.5-bit modes read FP16 mode LUTs precomputed from the stored 8-bit LUT, and the 4-bit mode builds its LUT inside the kernel from the 8-bit codes. H planes are stored as 4-bit messages, which the kernel either expands to codewords with a 16-entry codeword table in registers or turns into indices directly by XOR in the message domain. Small matrices fill the GPU with split-K and large ones with a persistent stream-K schedule, which prefetches the LUT of the next tile with \texttt{cp.async}. Like the kernels of the other methods, Any-Rate selects among these variants and their launch settings per matrix shape and mode, using short preliminary runs separate from the measurement rounds.

The baselines use the kernels released by their authors without modification, and we checked that each kernel's output matches its own dense path within a relative error of $3.1\times10^{-3}$. The released AP kernel supports only 3 bits and above, so we time 2 bits with the AP kernel included in the released AB code; at 3 and 4 bits this kernel produces bit-identical outputs to the released AP kernel, and its kernel time, measured with a profiler, agrees within 0.5\%. The released execution path of MG applies its \texttt{act\_order} permutation outside the kernel, so we report times with and without the permutation. Over a whole token (Table~\ref{tab:gemv}), Any-Rate is faster than AB and than MG with the permutation and within 5--14\% of AP. Per layer (Table~\ref{tab:gemv_layer}), it matches AP on down at 2 bits and on q/o at 4 bits and is faster on down at 4 bits, is 4\% slower on down at 3 bits and 8--18\% slower on gate/up and on q/o at 2 and 3 bits, and is 1.3--1.4 times slower on the short-row k/v, where building the LUT of each row is a fixed cost.

\begin{table}[h]
\caption{Linear-layer GEMV time per decoded token (ms) of Llama-3.1-8B-Instruct, batch 1, RTX 5070 Ti. Lower is better; fastest per column in bold, second underlined.}
\label{tab:gemv}
\centering
\small
\setlength{\tabcolsep}{4pt}
\begin{tabular}{lccccc}
\toprule
Method & 2 & 2.5 & 3 & 3.5 & 4 \\
\midrule
AP & \textbf{2.84} & -- & \textbf{3.85} & -- & \textbf{5.14} \\
AB & 3.46 & -- & 4.64 & -- & 5.90 \\
MG (2/3/4) & 3.27 & -- & \underline{4.22} & -- & \underline{5.16} \\
\quad + \texttt{act\_order} & 4.12 & -- & 5.06 & -- & 5.99 \\
\rowcolor[HTML]{DCEBFA} AR (ours) & \underline{3.08} & 3.96 & 4.37 & 4.90 & 5.41 \\
\bottomrule
\end{tabular}
\end{table}

\begin{table}[h]
\centering
\caption{GEMV latency (\textmu s) of the Llama-3.1-8B-Instruct linear layers, batch 1, RTX 5070 Ti. MG: MG (2/3/4); MG$^\dagger$: with the \texttt{act\_order} permutation. FP16 (cuBLAS): 41.4 (q/o), 12.5 (k/v), 140.5 (gate/up), 143.6 (down). Fastest per column in bold, second underlined.}
\label{tab:gemv_layer}
\small
\setlength{\tabcolsep}{2.6pt}
\begin{tabular}{@{}lcccccccccccccccccccc@{}}
\toprule
 & \multicolumn{5}{c}{q/o ($4096\times4096$)} & \multicolumn{5}{c}{k/v ($1024\times4096$)} & \multicolumn{5}{c}{gate/up ($14336\times4096$)} & \multicolumn{5}{c}{down ($4096\times14336$)} \\
\cmidrule(lr){2-6} \cmidrule(lr){7-11} \cmidrule(lr){12-16} \cmidrule(lr){17-21}
Bits & 2 & 2.5 & 3 & 3.5 & 4 & 2 & 2.5 & 3 & 3.5 & 4 & 2 & 2.5 & 3 & 3.5 & 4 & 2 & 2.5 & 3 & 3.5 & 4 \\
\midrule
AP & \textbf{7.9} & -- & \textbf{10.3} & -- & \underline{13.7} & \underline{3.8} & -- & \underline{4.3} & -- & \underline{5.0} & \textbf{21.0} & -- & \textbf{29.5} & -- & \textbf{38.6} & \textbf{23.3} & -- & \underline{31.9} & -- & 46.0 \\
AB & 9.5 & -- & 12.1 & -- & 15.0 & 6.1 & -- & 7.0 & -- & 8.0 & 25.8 & -- & 35.7 & -- & 46.3 & 25.7 & -- & 35.6 & -- & 45.9 \\
MG & 8.7 & -- & \underline{11.0} & -- & \textbf{12.5} & \textbf{3.4} & -- & \textbf{4.0} & -- & \textbf{4.7} & 26.9 & -- & 35.0 & -- & 43.4 & 24.4 & -- & \textbf{31.8} & -- & \textbf{40.3} \\
MG$^\dagger$ & 12.3 & -- & 14.6 & -- & 15.8 & 7.2 & -- & 7.8 & -- & 8.4 & 30.7 & -- & 38.8 & -- & 47.2 & 28.3 & -- & 35.9 & -- & 44.3 \\
\rowcolor[HTML]{DCEBFA}
AR (ours) & \underline{8.5} & 11.4 & 12.2 & 13.1 & 13.8 & 4.9 & 5.4 & 6.0 & 6.6 & 6.6 & \underline{23.0} & 29.9 & \underline{33.6} & 38.0 & \underline{42.5} & \underline{23.6} & 30.4 & 33.2 & 37.7 & \underline{43.4} \\
\bottomrule
\end{tabular}
\end{table}

\subsection{Decoding Throughput}

For the throughput in Table~\ref{tab:tps}, we implement a single decoder with the architecture of Llama-3.1-8B-Instruct (32 layers, GQA, SwiGLU, RoPE, KV cache) and replace only its linear layers with the module of each method. The embeddings, attention, and \texttt{lm\_head} are the same FP16 code for all methods, and q/k/v and gate/up are fused into single projections, as in the generation code of AB and in vLLM~\citep{kwon2023vllm}. One decoding step is captured as a CUDA graph, and we generate 256 tokens continuously from position 128, running the methods in two passes of three rounds each, in alternating order, and reporting the median of the six runs. Since kernel time does not depend on the weight values, the weights are random values in each format. FP16 is not included because the model does not fit in the 16\,GB memory. The cost outside the linear layers is about 4\,ms per token, the same for all methods.

\subsection{Energy Estimate}

LLM accelerators in circuit venues report the system energy per token in two parts: the energy of the computation on the chip, from its measured efficiency, and the energy of external memory access, estimated from the number of bits transferred and the energy per bit of the DRAM~\citep{moon2025trex,qin2025llmaccel,kim2025slimllama,dong2026reram}. We follow the same practice and estimate the energy per token as
\begin{equation}
E_{\mathrm{token}}
\;=\; \underbrace{\big(b_{\mathrm{eff}}\,N + B_{\mathrm{common}}\big)\,e_{\mathrm{DRAM}}}_{\text{external memory access}}
\;+\; \underbrace{\frac{2M}{\eta}}_{\text{computation}},
\label{eq:energy}
\end{equation}
where $N$ is the number of linear-layer weights, $b_{\mathrm{eff}}$ the bits actually read per weight in a mode (codes and metadata; Table~\ref{tab:beff}), $B_{\mathrm{common}}$ the bits of the FP16 \texttt{lm\_head} and of a KV cache of 1024 positions, which are the same for all methods, $M$ the number of MACs per token, $e_{\mathrm{DRAM}}=3.7$\,pJ/bit~\citep{moon2025trex}, and $\eta=88.36$\,TOPS/W~\citep{qin2025llmaccel}. For 4-bit Llama-3.1-8B, Eq.~\eqref{eq:energy} gives 144\,mJ/token including the FP16 \texttt{lm\_head}. Applied to 4-bit LLaMA2-7B, the same equation gives about 120\,mJ/token, on the same scale as the 123\,mJ/token that an accelerator reports for this model with modeled LPDDR3 access~\citep{dong2026reram}.

Batch-1 decoding reads every weight once per token and performs one MAC per weight, so the first term of Eq.~\eqref{eq:energy} dominates: the computation is at most 0.2\% of the energy per token (at 2 bits), and counting each table lookup and each decoding step as one more MAC raises it only to 0.6\%. The energy is therefore proportional to the bits read, and at equal accuracy, reducing the bits read per weight is the main way to reduce it, which is also why LLM accelerators focus on weight compression~\citep{qin2025llmaccel,kim2025slimllama,dong2026reram}. These bits include the metadata as well as the codes (Table~\ref{tab:beff}). AP reads a LUT of $2^b$ FP16 values per row, AB $b$ scales $\alpha$ and one offset $\beta$ per group, and MG a scale and a zero per group; MG (2/3/4) and MG (3/4/8) read the same amount at the same bit width. Any-Rate builds the FP16 LUT of a mode once from the stored 8-bit codes when the mode changes and reads it, and at 4 bits reads the codes directly.

\begin{table}[h]
\centering
\caption{Bits read per weight at each mode (codes and metadata), Llama-3.1-8B-Instruct linear layers, and the metadata each method reads.}
\label{tab:beff}
\small
\begin{tabular}{@{}lccccc l@{}}
\toprule
Method & 2 & 2.5 & 3 & 3.5 & 4 & Metadata read at $b$ bits \\
\midrule
AP & 2.01 & -- & 3.03 & -- & 4.05 & per row: LUT of $2^b$ FP16 values \\
AB & 2.38 & -- & 3.50 & -- & 4.63 & per group of 128: $b$ scales $\alpha$, one offset $\beta$ (FP16) \\
MG & 2.25 & -- & 3.25 & -- & 4.25 & per group of 128: scale and zero (FP16) \\
\rowcolor[HTML]{DCEBFA}
AR (ours) & 2.03 & 2.55 & 3.10 & 3.70 & 4.21 & per row: $2^{2b-1}$ FP16 values; 4 bits: 8-bit $T$, FP16 $\mathrm{lo}$ and $s$ (Appendix~\ref{app:table}) \\
\bottomrule
\end{tabular}
\end{table}

Figure~\ref{fig:energy}b compares, for each accuracy target, the energy of the cheapest mode of each method that reaches the target. Raising the memory access energy threefold or lowering the compute efficiency tenfold changes the energy Any-Rate saves at equal accuracy by less than 0.4 points, since the saving comes from the ratio of the bits read rather than from the two constants.

\section{Mixture-of-Experts Model} 
\label{app:moe}

\subsection{Why Mixture-of-Experts}
\label{app:moe_why}
Mixture-of-experts (MoE) models are a setting in which a checkpoint that serves several precisions is especially useful. In Qwen3-30B-A3B-Instruct-2507 \citep{yang2025qwen3}, the routed experts take 58.0\,GB of the 61.1\,GB BF16 checkpoint (95\%), and attention takes about 1.8\,GB. The model does not fit in the 16\,GB of a consumer GPU, so on-device inference keeps only part of the experts in VRAM and fetches the rest from host memory or SSD, and this loading latency dominates the decoding time \citep{cecar2026}. Expert weights are also known to tolerate low-bit quantization well \citep{kim2023moqe, frantar2024qmoe, li2024quantmoe, huang2024mcmoe}. Moreover, the output of an MoE layer is the gate-weighted sum of the selected experts, so the quantization error of an expert reaches the output scaled by its gate score, and the selected experts of a token need not be read at the same precision. DPA-MoE \citep{kim2026dpamoe} stores expert weights as prefix-decodable chunks and assigns the number of chunks to each token and expert at inference time according to the gate scores. Such runtime allocation needs a format from which several precisions can be read out of one stored checkpoint, and an Any-Rate ladder is such a format: each expert is stored once, and the 2- to 4-bit modes, including the half-bit modes, are read out of the same planes. This appendix reports the accuracy of Qwen3-30B-A3B with only its experts encoded by Any-Rate (Appendix~\ref{app:moe_results}), compares mode allocation by gate score with uniform allocation at the same average bit width (Appendix~\ref{app:moe_alloc}), and measures the decoding speed under a memory budget (Appendix~\ref{app:moe_speed}).

\subsection{Setup}
\label{app:moe_setup}
We encode only the routed experts with Any-Rate. These are the gate, up, and down projections of the 128 experts in each of the 48 layers. Attention, the router, the embeddings, and the normalization layers stay in BF16. They hold about 5\% of the parameters, and all bit widths in this appendix count the expert weights only. The calibration data, the ladder (UHHHHHH with mode weights 1-1-2-3-4), the candidate search ($B=K=64$), and the damping (0.01) follow the dense models (Appendix~\ref{app:setup}). Each expert gets its own Hessian from the calibration tokens routed to it. The gate and up projections share the expert input, and the down projection uses the input it receives inside the expert. Routing selects 8 of the 128 experts, so an expert sees 16k of the 262k calibration tokens on average, but the routing is uneven: the median expert sees about 8k, 970 of the 6,144 experts (16\%) receive fewer than 512 tokens, and one receives none. A Hessian built from a few hundred tokens covers only a few input directions, and the encoder then places the error in the directions it cannot see. For an expert with $n<512$ tokens we add $(512/n)\,(\operatorname{tr}\Gamma/d)\,I$ to its Hessian $\Gamma$, which amounts to adding 512 isotropic tokens with the same mean energy. The expert with no tokens uses an identity Hessian.

We report WikiText-2 perplexity and the eight benchmarks of Section~\ref{sec:setup}, run with the same harness and settings as for the dense models. For a nested baseline we also encode the experts with the 2/3/4-bit version of MatGPTQ (MG)~\citep{kleinegger2026matgptq}, MG (2/3/4), using the same calibration data and the same rule for experts with few tokens, and compare its perplexity with that of Any-Rate.

\begin{figure*}[t]
\centering
\includegraphics[width=\textwidth]{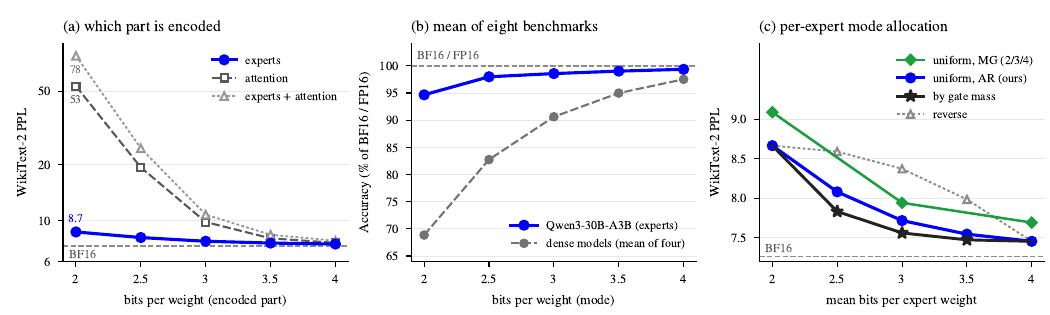}
\caption{Qwen3-30B-A3B-Instruct-2507 with Any-Rate. (a) WikiText-2 perplexity by encoded part. (b) Mean accuracy on the eight benchmarks relative to BF16 / FP16. (c) Per-expert mode allocation at the same mean bits per expert weight, with uniform MG (2/3/4) and Any-Rate for reference.}
\label{fig:moe}
\end{figure*}

\subsection{Results}
\label{app:moe_results}
The uniform rows of Table~\ref{tab:moe_ar} read every expert in the same mode. With only its experts encoded, Qwen3-30B-A3B is close to BF16 at 4 bits and keeps about 95\% of the BF16 mean accuracy at 2 bits, and the half-bit modes fall in order between the integer modes. Only TriviaQA, which tests knowledge recall, drops sharply as the bit width goes down; the other seven tasks lose less than five points even at 2 bits. Figure~\ref{fig:moe}(b) compares this ratio with the mean of the four dense models of Section~\ref{sec:main_results}. The dense models have all their linear layers encoded, so the comparison is not direct, but the gap is large: the dense models fall below 70\% at 2 bits. An MoE model can use a wider range of modes, and reading several precisions from one checkpoint pays off more.

Figure~\ref{fig:moe}(a) shows why we encode only the experts. Encoding only attention, 3\% of the parameters, raises the perplexity sharply below 3 bits and past 50 at 2 bits, while encoding only the experts raises it by 1.4 over BF16 at 2 bits. The two uniform lines in Figure~\ref{fig:moe}(c) are MG (2/3/4) and Any-Rate, encoded with the same calibration data and the same rule for experts with few tokens. Any-Rate has a lower perplexity than MG (2/3/4) at 2, 3, and 4 bits, and it also provides the 2.5- and 3.5-bit modes, which MG (2/3/4) lacks, from the same checkpoint.

\begin{table*}[h]
\centering
\caption{Qwen3-30B-A3B-Instruct-2507 with experts encoded by Any-Rate: WikiText-2 perplexity and accuracy (\%). Bits: mean bits per expert weight. Better allocation in \textbf{bold}.}
\label{tab:moe_ar}
\footnotesize
\setlength{\tabcolsep}{4.2pt}
\renewcommand{\arraystretch}{1.1}
\begin{tabular}{cl|c|cccccccc|>{\columncolor[gray]{0.92}}c}
\toprule
Bits & Allocation & PPL & ARC-c & ARC-e & HellaSwag & PIQA & WinoGrande & GSM8K & TriviaQA & NQ-open & Avg. \\
\midrule
2 & uniform & 8.66 & 58.70 & 82.41 & 75.20 & 79.71 & 73.01 & 90.67 & 54.80 & 17.70 & 66.53 \\
\midrule
\multirow{2}{*}{2.5} & uniform & 8.08 & 61.09 & \textbf{83.42} & 77.37 & \textbf{80.85} & \textbf{74.66} & \textbf{91.96} & 60.70 & 20.80 & 68.86 \\
 & gate mass & \textbf{7.83} & \textbf{61.26} & 82.87 & \textbf{78.42} & 80.58 & 73.95 & 91.05 & \textbf{62.40} & \textbf{22.10} & \textbf{69.08} \\
\midrule
\multirow{2}{*}{3} & uniform & 7.72 & 60.32 & \textbf{82.41} & 78.13 & \textbf{80.63} & \textbf{74.11} & \textbf{92.87} & 63.80 & 21.80 & 69.26 \\
 & gate mass & \textbf{7.56} & \textbf{60.41} & \textbf{82.41} & \textbf{78.73} & 80.25 & 73.56 & 91.96 & \textbf{66.70} & \textbf{22.70} & \textbf{69.59} \\
\midrule
\multirow{2}{*}{3.5} & uniform & 7.55 & 60.92 & 83.00 & 78.62 & \textbf{80.58} & 73.01 & \textbf{92.80} & 65.90 & 21.80 & 69.58 \\
 & gate mass & \textbf{7.47} & \textbf{61.18} & \textbf{83.46} & \textbf{78.99} & 80.03 & \textbf{73.40} & 92.57 & \textbf{66.90} & \textbf{22.80} & \textbf{69.92} \\
\midrule
4 & uniform & 7.45 & 60.58 & 82.83 & 79.00 & 80.30 & 73.72 & 92.87 & 67.20 & 22.00 & 69.81 \\
\midrule
BF16 & - & 7.26 & 62.03 & 83.59 & 79.70 & 80.47 & 73.80 & 92.27 & 68.00 & 22.40 & 70.28 \\
\bottomrule
\end{tabular}
\end{table*}

\subsection{Mode Allocation by Gate Mass}
\label{app:moe_alloc}
Since the quantization error of an expert reaches the output scaled by its gate score (Appendix~\ref{app:moe_why}), we compute for each expert its gate mass, the sum of its gate scores over the calibration tokens, and the relative proxy loss (Eq.~\eqref{eq:proxy_loss}) of each mode, averaged over its three matrices. In each layer we then raise expert modes half a bit at a time, greedily, to minimize the sum of gate mass times error at a fixed mean expert bit width. The allocation is set once per expert, offline, and changes only how many planes are read from each expert, with no re-encoding. As a control we also evaluate the reverse allocation, which keeps the same bit histogram but assigns it in reverse gate-mass order. In Figure~\ref{fig:moe}(c), allocation by gate mass gives a lower perplexity than uniform allocation at all three mean bit widths, and the reverse allocation is worse than uniform, so the gain comes from giving bits in gate-mass order. In Table~\ref{tab:moe_ar}, allocation by gate mass also has a higher mean accuracy than uniform allocation, and at 3.5 bits it is above uniform 4 bits. We evaluate only this offline per-expert allocation; a per-token allocation such as that of DPA-MoE only changes the prefix length read from each expert, so the same checkpoint can serve it without re-encoding.

\begin{figure*}[h]
\centering
\includegraphics[width=\textwidth]{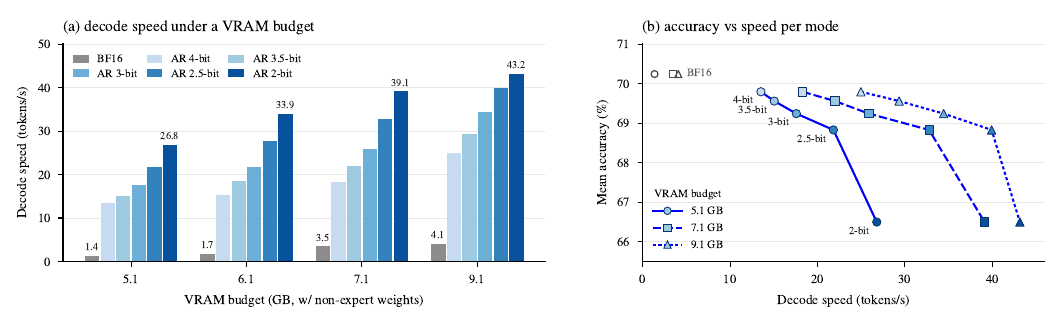}
\caption{Decoding Qwen3-30B-A3B-Instruct-2507 on an RTX 5070 Ti (16\,GB) with an expert cache in VRAM. (a) Decode speed per VRAM budget. (b) Mean accuracy vs.\ decode speed of each mode.}
\label{fig:moe_speed}
\end{figure*}

\subsection{Decoding Speed under a Memory Budget}
\label{app:moe_speed}
We measure the decoding speed of Qwen3-30B-A3B on a 16\,GB consumer GPU (RTX 5070 Ti). The non-expert weights (3.1\,GB) stay in VRAM in BF16, and the rest of the VRAM budget holds a per-layer LRU cache of experts. An expert that misses the cache is read from an NVMe SSD with O\_DIRECT, so the page cache in host memory does not act as a second cache. Any-Rate stores the planes of an expert from the lowest up, so the mode that reads the first $k$ planes is a single read. BF16 experts are read the same way from the original safetensors files. We decode with batch size 1, generating 64 tokens for each of three prompts, and report the mean speed.

In Figure~\ref{fig:moe_speed}(a), Any-Rate decodes 5 to 20 times faster than BF16 under the same VRAM budget. Its experts are smaller, so more of them fit in the cache and a miss reads fewer bytes. At 9.1\,GB, BF16 reads almost half of the requested experts from the SSD and spends 85\% of the time on reads, while Any-Rate at 3 bits serves 95\% of the requests from the cache. Lower modes are faster until, at 2 bits, nearly every expert fits in the cache and computation becomes the bottleneck, so the speed levels off at about 43 tokens/s (43.4 and 43.0 tokens/s at 11.1 and 13.1\,GB). Figure~\ref{fig:moe_speed}(b) shows the accuracy and speed of each mode under the same budget. At each budget in Figure~\ref{fig:moe_speed}(b), going from 4 to 2.5 bits lowers the mean accuracy by less than one point while raising the speed 1.6--1.8 times, whereas going from 2.5 to 2 bits adds only 8--23\% speed and loses 2.3 points. One checkpoint covers all these operating points, the mode is chosen to fit the available VRAM, and the half-bit modes make the choice finer. Figure~\ref{fig:demo_moe} in Appendix~\ref{app:demo} shows this setup running as a chat demo on the same GPU, with the mode switched during generation.

\section{Beyond Language Models} 
\label{app:beyond_lm}

Any-Rate needs only the weights of each linear layer and the Hessian of its inputs, so it is not limited to language models and applies to other networks whose weights lie mostly in linear layers. Models beyond language models also have to run at different levels of quality depending on user needs and the execution environment: the same model is often run at high quality when memory or power is plentiful and at a lower bit width, giving up some quality, when they are scarce. Keeping a separately quantized checkpoint for each bit width multiplies the storage and maintenance cost by the number of bit widths, whereas Any-Rate reads integer and half-bit modes from one checkpoint and meets these needs with a single model file. In this appendix, we build checkpoints of a vision-language model, diffusion models (image and video), and a speech recognition model with the same procedure and examine how their outputs change across bit widths.

The goal of this appendix is to show that Any-Rate works across model types and that quality changes gradually with the bit width, not to claim the best result at every bit width. For the speech recognition, diffusion, and video generation models we compare with MatGPTQ (MG)~\citep{kleinegger2026matgptq} applied with the same calibration data, using its 2/3/4-bit version (4-bit master) as in our main results. Since MG targets only language models, we built it ourselves by applying its quantization algorithm to the Hessians computed from the linear-layer inputs of each model. Any-Precision LLM (AP)~\citep{park2024anyprecision} obtains per-weight sensitivities from the gradient of the language-model loss, and AnyBCQ (AB)~\citep{park2026anybcq} relies on a training procedure that assumes the block structure of decoder-only language models; applying them would require a new loss function and training code for every model, so we do not include them. We release the encoding code for the models in this appendix, the Any-Rate checkpoints of each model in quantization repositories linked to the original models, and an environment in which users can run them while switching the bit width.

\subsection{Common Procedure}
\label{app:beyond_lm_setup}

In every model, we encode only the linear layers inside the transformer blocks with Any-Rate and keep the embeddings, normalization layers, convolutions, and input and output projections at their original precision. Since Any-Rate needs only the input Hessian of each linear layer, we attach forward hooks to the original model, pass the calibration inputs through it once to collect the Hessians of all target linear layers at the same time, and then encode each layer with the same procedure and settings as in the main text (the ladder UHHHHHH with mode weights 1-1-2-3-4, and the residual planes).

\begin{table*}[h]
\caption{Qualitative comparison of Gemma-3-4b-it with Any-Rate (language model and vision tower in the same bit mode). Responses to ``Describe this image in one sentence.'' for two photos (cat, left; dog, right), at each bit width (greedy decoding). Only the introductory phrase that the model sometimes prepends (e.g., ``Here's a one-sentence description of the image:'') is removed.}
\label{tab:vlm_qualitative}
\centering
\footnotesize
\setlength{\tabcolsep}{6pt}
\renewcommand{\arraystretch}{1.2}
\setlength{\aboverulesep}{5pt}
\setlength{\belowrulesep}{4pt}
\begin{tabular}{@{}l p{0.43\textwidth} p{0.43\textwidth}@{}}
\toprule
 & \multicolumn{1}{c}{\includegraphics[width=0.43\textwidth]{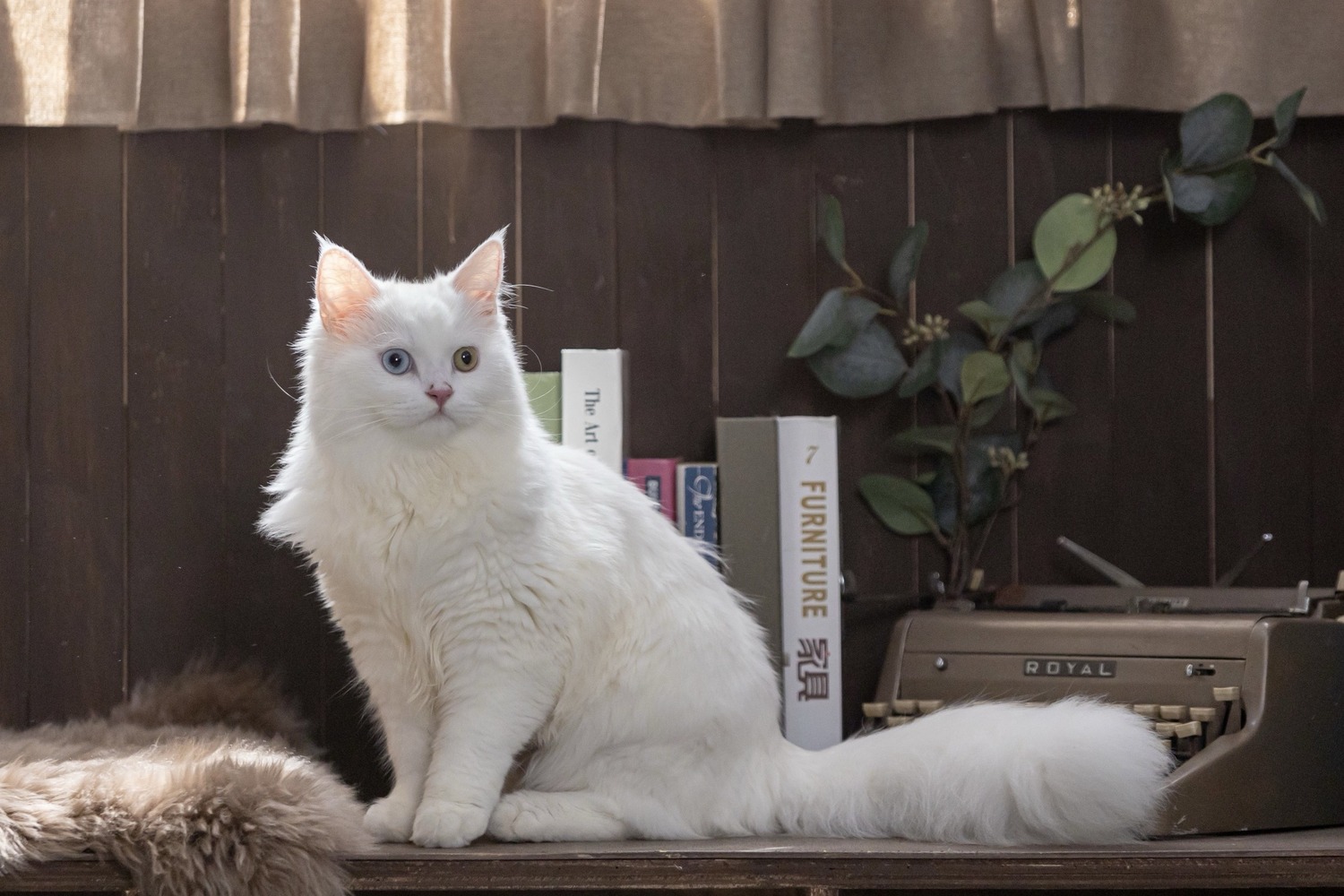}} & \multicolumn{1}{c}{\includegraphics[width=0.43\textwidth]{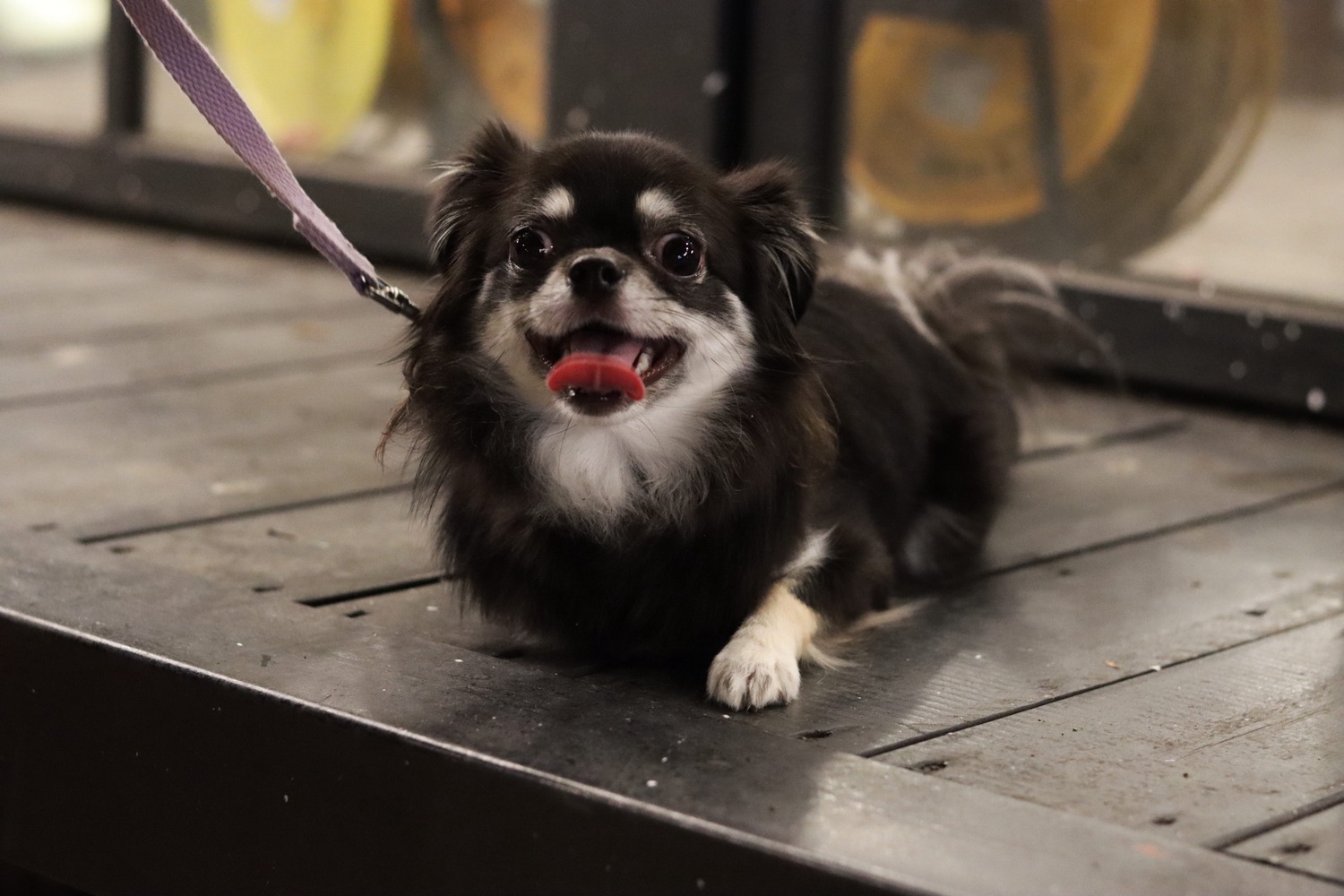}} \\
\midrule
\textbf{FP16} & A fluffy, white cat with striking blue eyes sits regally on a wooden surface next to a vintage typewriter and stacks of books, creating a charming and slightly vintage scene. & A happy, dark-colored chihuahua with white markings and a visible tongue is lying on a dark surface with a leash attached to its collar, looking directly at the camera with a joyful expression. \\
\cmidrule[0.3pt]{1-3}
\textbf{4-bit} & A fluffy, white cat with striking blue eyes sits regally on a wooden bench, surrounded by books and vintage typewriter, in a warmly lit room. & A happy, black and white chihuahua with its tongue out is lying on a dark wooden floor, wearing a leash and looking directly at the camera. \\
\cmidrule[0.3pt]{1-3}
\textbf{3.5-bit} & A fluffy, white cat with striking blue eyes sits regally on a wooden surface, surrounded by books and a vintage typewriter, gazing intently into the distance. & A happy, dark-and-white chihuahua with its tongue out is playfully lying down on a wooden floor, wearing a leash and looking towards the camera. \\
\cmidrule[0.3pt]{1-3}
\textbf{3-bit} & A fluffy white cat with striking blue eyes sits regally on a wooden bench next to a vintage Royal typewriter, framed by a rustic wooden fence and a leafy plant. & A cute, dark-colored chihuahua with white markings sits happily with its tongue out, wearing a leash, on a wooden floor with a window in the background. \\
\cmidrule[0.3pt]{1-3}
\textbf{2.5-bit} & A fluffy white cat sits perched on a wooden bench, gazing at a brown wooden fence. & A cute, small dog with a fluffy coat is lying on a dark floor, looking at the camera with a playful expression. \\
\cmidrule[0.3pt]{1-3}
\textbf{2-bit} & A fluffy white cat sits on a wooden fence, surrounded by books and plants, looking contemplative. & A cute, black and white chihuahua sits on a wooden deck, with its mouth open, showing teeth. \\
\bottomrule
\end{tabular}
\end{table*}

\subsection{Vision-Language Model}
\label{app:beyond_lm_vlm}

The Gemma-3-4b-it~\citep{team2025gemma3} of the main text was evaluated on the language model taken from its multimodal checkpoint. Here we also encode the SigLIP~\citep{zhai2023siglip} vision tower of the same checkpoint with Any-Rate, so that the whole model, which takes images as input, is read at several bit widths from one checkpoint. The language model uses the same encoding as in the main text, and the linear layers of the vision tower are encoded at the same bit width. We do not apply a baseline to this model and show only FP16 and Any-Rate at each bit width, comparing the responses to two photos qualitatively (Table~\ref{tab:vlm_qualitative}). The question is ``Describe this image in one sentence.'' at every bit width, and the responses are obtained with greedy decoding. As the bit width decreases, the model does not collapse abruptly; its responses lose details gradually. From 4 to 3 bits, the responses name the main objects as FP16 does, such as the cat and the typewriter or the chihuahua and its leash. At 2.5 bits the scene remains, but details disappear first: the breed and the leash of the dog, and the books and the typewriter next to the cat. Even at 2 bits the responses are natural sentences about the photo, but they contain statements that do not match it, such as the cat sitting on the wooden fence of the background.

\begin{table}[h]
\centering
\caption{WER (\%, lower is better) of Whisper-large-v3. Mean: unweighted mean of the three sets. MG (2/3/4) has no 2.5-, 3.5-, 6-, or 8-bit modes.}
\label{tab:whisper}
\small
\setlength{\tabcolsep}{4pt}
\begin{tabular}{lccccccccccc}
\toprule
 & FP16 & \multicolumn{7}{c}{AR (ours)} & \multicolumn{3}{c}{MG (2/3/4)} \\
\cmidrule(lr){3-9} \cmidrule(lr){10-12}
Bits & 16 & 8 & 6 & 4 & 3.5 & 3 & 2.5 & 2 & 4 & 3 & 2 \\
\midrule
test-clean  & 2.03  & 2.04  & 2.03  & 2.10  & 2.11  & 2.20  & 2.24  & 2.39  & 2.03  & 2.20  & 69.36 \\
test-other  & 3.90  & 3.92  & 3.97  & 4.06  & 4.16  & 4.24  & 4.48  & 5.31  & 4.40  & 4.84  & 93.79 \\
Earnings-22 & 11.34 & 11.29 & 11.17 & 11.24 & 11.36 & 11.45 & 11.46 & 12.44 & 11.37 & 12.02 & 105.61 \\
\midrule
Mean        & 5.75  & 5.75  & 5.72  & 5.80  & 5.87  & 5.96  & 6.06  & 6.71  & 5.93  & 6.35  & 89.59 \\
\bottomrule
\end{tabular}
\end{table}

\subsection{Speech Recognition}
\label{app:beyond_lm_speech}

We encode all linear layers in the encoder and decoder of Whisper-large-v3~\citep{radford2023whisper} (attention projections and feed-forward layers), keeping the convolutional front end, embeddings, and normalization layers at their original precision. The Hessians are collected from 128 utterances of LibriSpeech~\citep{panayotov2015librispeech} dev-clean, with log-mel features fed to the encoder and the reference transcripts to the decoder with teacher forcing. We report the word error rate (WER) on LibriSpeech test-clean and test-other and on Earnings-22~\citep{delrio2022earnings22}, with greedy decoding in English transcription mode and Whisper's text normalization; the baseline is MG (2/3/4), as described at the beginning of this appendix.

Any-Rate stays within 0.05 points of FP16 down to 4 bits and rises gradually as the bit width decreases, to 6.71\% at 2 bits, with the half-bit modes between the integer modes (Table~\ref{tab:whisper}). MG (2/3/4) is slightly higher at 4 and 3 bits, which also holds sentence by sentence: among the 8,300 utterances, Any-Rate alone transcribes 398 exactly at 3 bits and MG (2/3/4) alone 226. At 2 bits, MG (2/3/4) collapses to a mean WER of 89.6\%, transcribing 14\% of the test-clean utterances as ``Thank you.'' and shortening 409 of the 2,536 utterances with at least five words to two words or fewer, which happens once for Any-Rate. Table~\ref{tab:whisper_ex} shows two utterances in which each method makes a single error at 3 bits, while MG (2/3/4) loses the content of both at 2 bits.

\begin{table}[h]
\centering
\caption{Two utterances of LibriSpeech test-other transcribed by Whisper-large-v3 at each bit width (greedy decoding).}
\label{tab:whisper_ex}
\small
\begin{tabular}{@{}ll p{0.34\linewidth} p{0.34\linewidth}@{}}
\toprule
 & Bits & Utterance 1 & Utterance 2 \\
\midrule
\multicolumn{2}{@{}l}{Reference} & hence a resort to cider and beer & toward dark a man went in to buy a lamp \\
\midrule
FP16 & 16 & Hence a resort to cider and beer. & Toward dark, a man went in to buy a lamp. \\
\midrule
\multirow{5}{*}{AR (ours)} & 4 & Hence a resort to cider and beer. & Toward dark, a man went in to buy a lamp. \\
 & 3.5 & Hence a resort to cider and beer. & Toward dark, a man went in to buy a lamp. \\
 & 3 & Hence a resort to cider and beer. & To our dark, a man went in to buy a lamp. \\
 & 2.5 & Hence a resort to cider and beer. & To a dark, a man went in to buy a lamp. \\
 & 2 & Hence a resort to cider and beer. & To a dark, a man went in to buy a lamp. \\
\midrule
\multirow{3}{*}{MG (2/3/4)} & 4 & Hence a resort to cider and beer. & Toward dark, a man went in to buy a lamp. \\
 & 3 & Hansen Resort to Cider and Beer & Toward dark, a man went in to buy a lamp. \\
 & 2 & Here's to be so excited and beer. & word dog. \\
\bottomrule
\end{tabular}
\end{table}

\subsection{Diffusion Model}
\label{app:beyond_lm_diffusion}

In diffusion models, Any-Rate follows the same procedure layer by layer. The text-to-image model PixArt-$\Sigma$~\citep{chen2024pixartsigma} has two components to encode, each for a different reason. The T5 text encoder~\citep{raffel2020c4}, which turns the prompt into embeddings, has many parameters and takes a large share of memory but runs only once per prompt. The diffusion transformer (DiT)~\citep{peebles2023dit}, which generates the image from these embeddings, is smaller but runs at every diffusion step, so most of the computation for an image comes from it. Both consist of linear layers and are encoded with the same procedure, while the mostly convolutional VAE stays at its original precision. The Hessian of each linear layer is collected while encoding the prompts for T5 and over the whole diffusion process of generating images for the DiT. All images are generated with the same prompt, seed, and generation settings at every bit width and compared side by side qualitatively; we do not report quality metrics or speed. MG (2/3/4) is built as described at the beginning of this appendix, with a 4-bit master from which the 2- and 3-bit modes are read. Because T5 and the DiT affect the result differently, we show the cases in which only one of them or both are quantized.

\begin{figure}[h]
\centering
\includegraphics[width=0.84\linewidth]{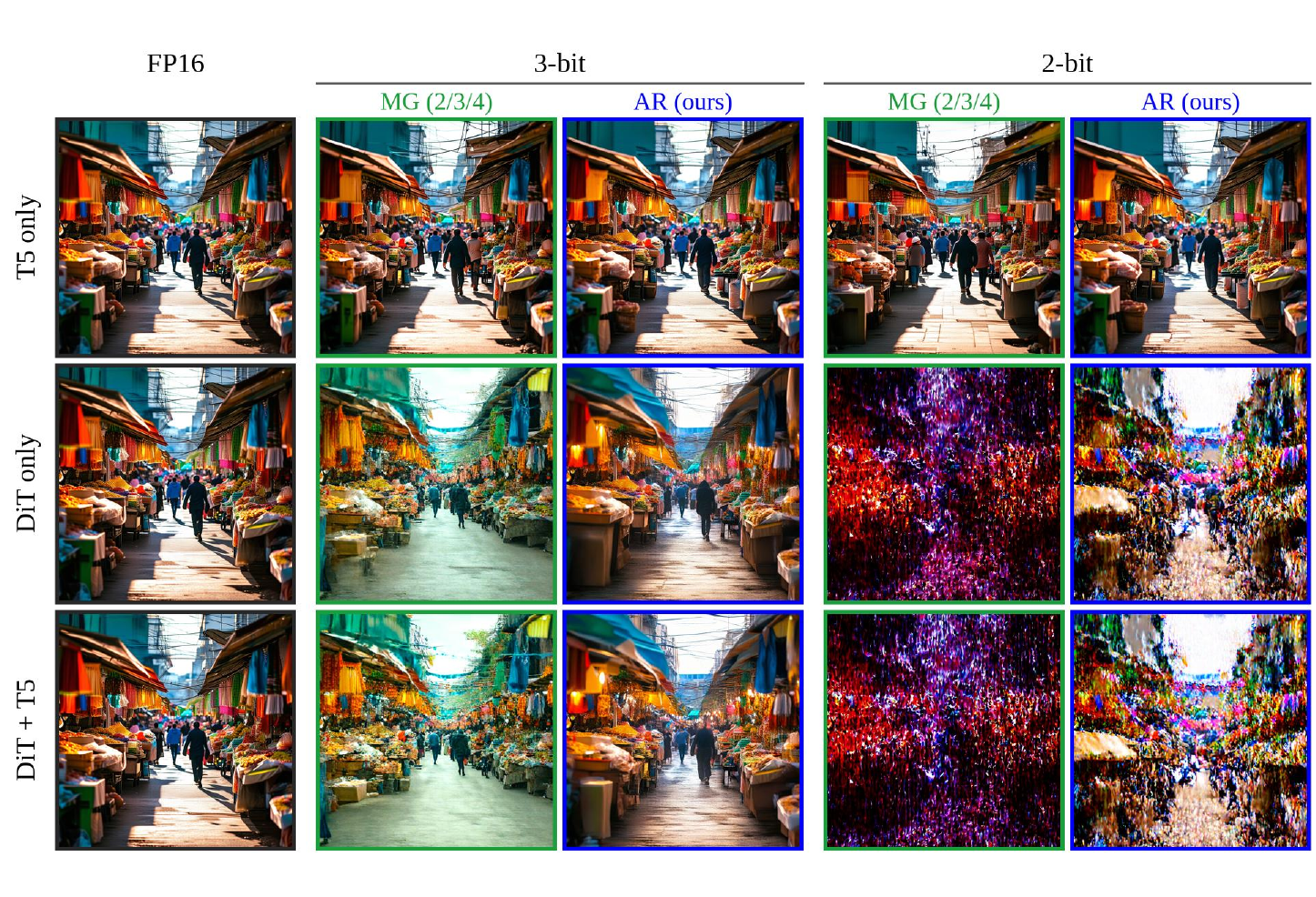}
\caption{Effect of quantizing each component of PixArt-$\Sigma$ for the prompt ``A busy street market with many people and colorful stalls'': the T5 text encoder only, the DiT only, and both, at 3 and 2 bits with MG (2/3/4) and Any-Rate. Same prompt and seed in every image.}
\label{fig:diffusion_parts}
\end{figure}

\begin{figure}[h]
\centering
\includegraphics[width=0.84\linewidth]{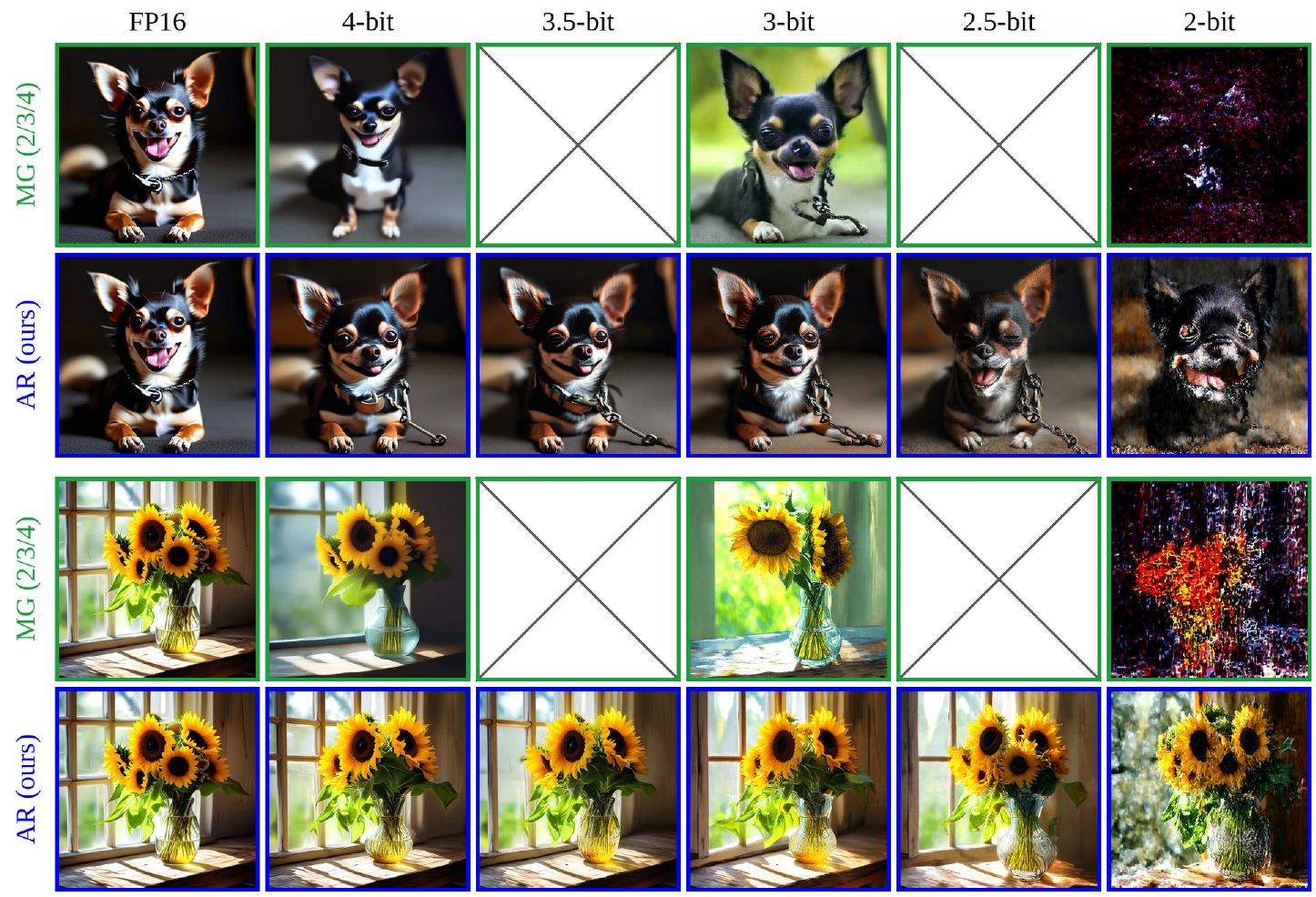}
\caption{PixArt-$\Sigma$ images at each bit width for two prompts, with the DiT and T5 quantized together. For each prompt, MG (2/3/4) (top) and Any-Rate (bottom); MG (2/3/4) has no 2.5- and 3.5-bit modes.}
\label{fig:diffusion_overview}
\end{figure}

Figure~\ref{fig:diffusion_parts} shows one prompt with the two components quantized separately and together. With only T5 quantized, both methods produce images almost identical to FP16 at 3 and 2 bits. Quantizing the DiT changes the image. At 3 bits, Any-Rate keeps the same street scene with a slightly different viewpoint and lighting, and at 2 bits the stalls smear into streaks while the street remains. MG (2/3/4) changes the composition at 3 bits and turns into streaked noise on a dark background at 2 bits. Quantizing both components gives almost the same result as quantizing the DiT alone, so in this example the DiT sets the limit of image quality. Figure~\ref{fig:diffusion_overview} shows two prompts at each bit width with the DiT and T5 quantized together. Any-Rate keeps the same scene and objects as FP16 down to 2.5 bits for both prompts, and at 2 bits the result depends on the prompt: the sunflowers remain recognizable, whereas the facial features of the dog blur. MG (2/3/4) already changes the composition at 4 bits, keeps the main object with a different scene at 3 bits, and at 2 bits turns into noise for both prompts, with only the colors of the sunflowers faintly left, the same tendency as in the street example.

Since Any-Rate reads T5 and the DiT independently from one checkpoint, it can also serve them at different bit widths without additional checkpoints, for example T5 at a high bit width when the prompt must be followed closely and the DiT at a low bit width when image detail matters less; we did not evaluate such mixed combinations.

\subsection{Video Generation}
\label{app:beyond_lm_video}

\begin{figure}[p]
\centering
\includegraphics[width=0.95\textwidth]{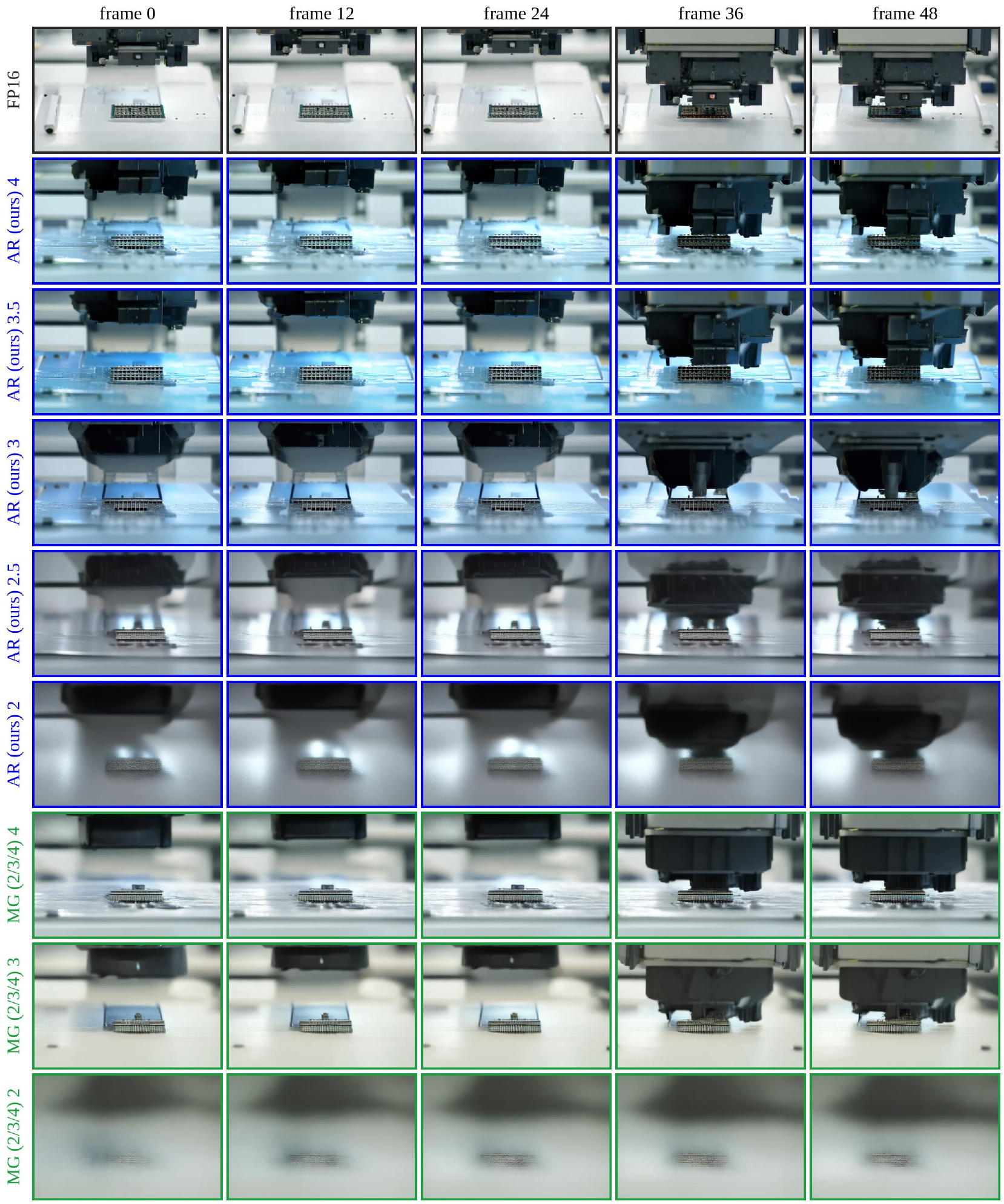}
\caption{CogVideoX-2B frames 0, 12, 24, 36, and 48. Rows: FP16, AR (ours) at 4 to 2 bits, and MG (2/3/4) at 4, 3, and 2 bits; the DiT and T5 are quantized together, with the same prompt and seed. Prompt: ``A robotic arm picks up a tiny chip and places it onto a circuit board in a clean factory, camera slowly panning.''}
\label{fig:video}
\end{figure}

For the text-to-video model CogVideoX-2B \citep{yang2025cogvideox}, we encode the 180 linear layers of the diffusion transformer (DiT) and the 168 linear layers of the T5 text encoder together with Any-Rate and keep the 3D VAE at its original precision. The T5 encoder has the same weights as that of PixArt-$\Sigma$, so we reuse its encoding, and the DiT Hessians are collected at the inputs of the linear layers while running the diffusion process (50 steps) on calibration prompts. Every video is generated with the same prompt, seed, and sampling settings (50 steps, guidance 6, 49 frames), and the comparison is again with MG (2/3/4). We do not report quality metrics or speed; instead, Figure~\ref{fig:video} places five frames taken in temporal order from one video (frames 0, 12, 24, 36, and 48) side by side for each bit width. In the FP16 video, the robot head stays above the chip and comes down onto it at frame 36. Any-Rate keeps this motion in every mode from 4 to 2 bits, while the details fade as the bit width decreases: at 4 and 3.5 bits the grid of the chip and the shape of the head are visible, but the background turns bluish instead of the white of FP16; at 3 bits the shape of the head departs from FP16; at 2.5 bits the scene blurs; and at 2 bits only the outlines of the head and the chip remain, although the head still comes down at frame 36. MG (2/3/4) keeps the white background and the descending motion at 4 and 3 bits, so at these two bit widths its colors are closer to FP16 than those of Any-Rate. At 2 bits, all five frames show only a blurred smear and a faint outline of the chip, and the motion of the head is gone.

\section{Demo}
\label{app:demo}
\begin{figure}[p]
\centering
\includegraphics[height=0.955\textheight]{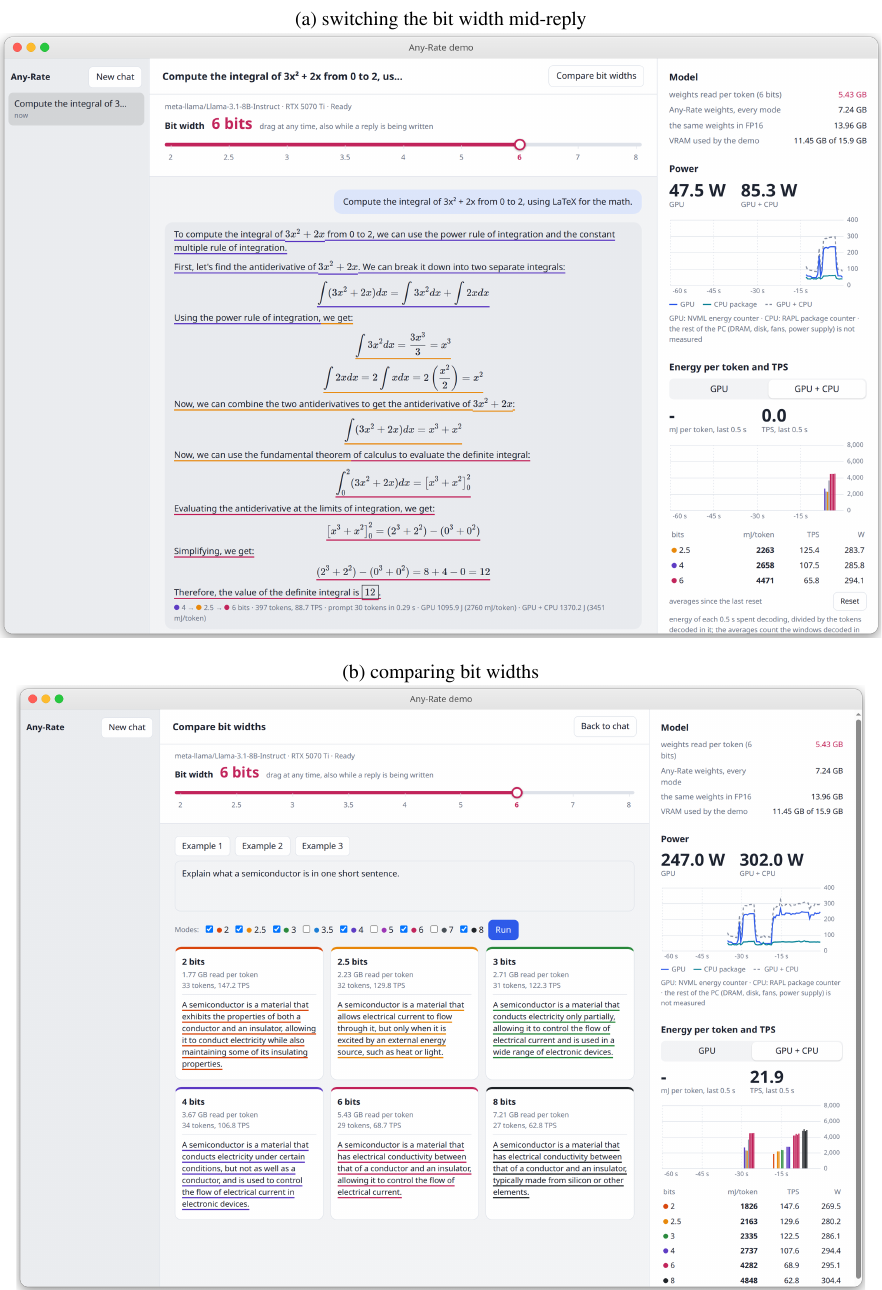}
\caption{Demo on Llama-3.1-8B-Instruct.}
\label{fig:demo_llama}
\end{figure}

\begin{figure}[p]
\centering
\includegraphics[height=0.955\textheight]{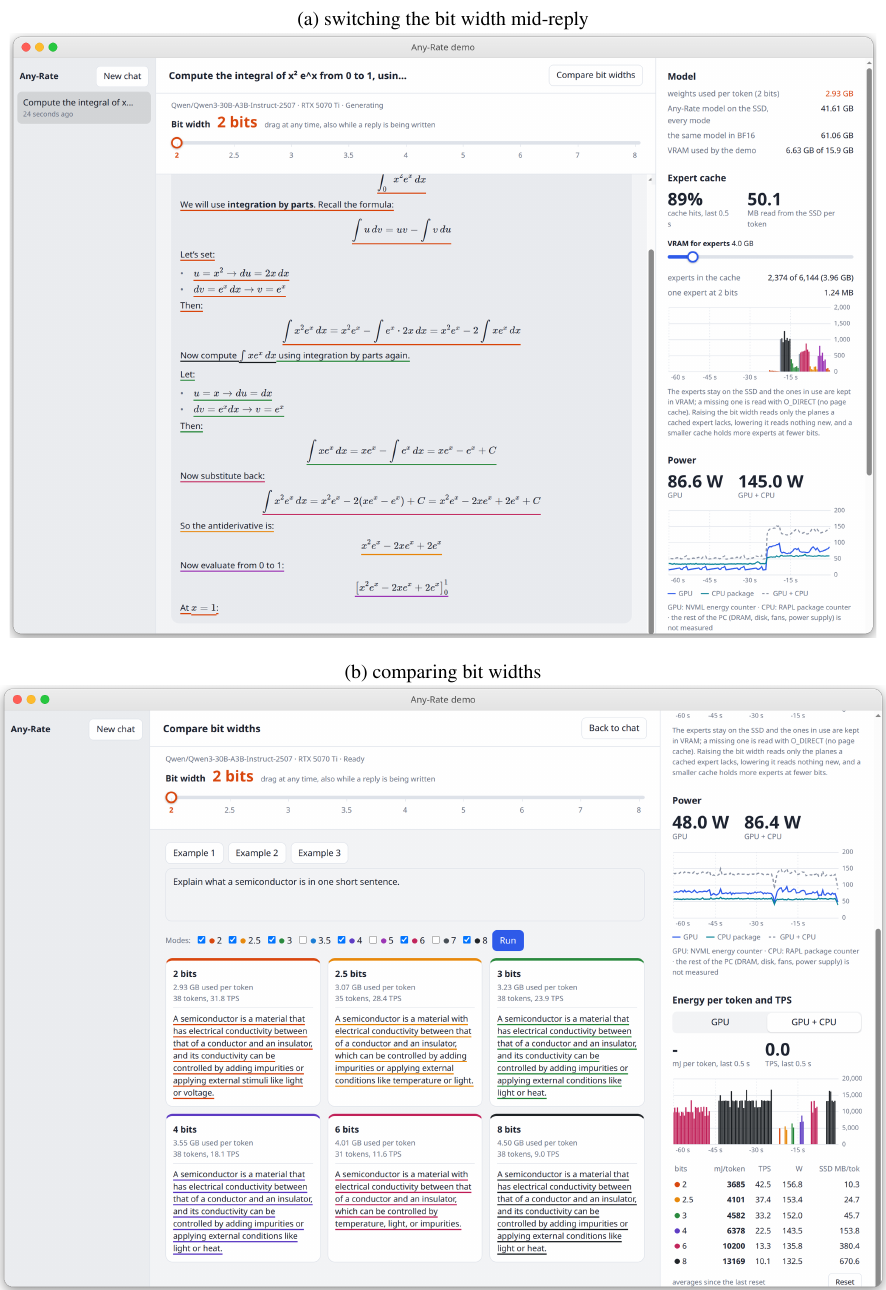}
\caption{Demo on Qwen3-30B-A3B-Instruct-2507.}
\label{fig:demo_moe}
\end{figure}

Figures~\ref{fig:demo_llama} and~\ref{fig:demo_moe} show a chat demo that runs from a single Any-Rate checkpoint on an RTX 5070 Ti (16\,GB). The bit width can be switched among the nine modes from 2 to 8 bits at any time; when it is switched during a reply, decoding continues in the new mode from the next token. In (a), the underline color of each part of the reply marks the mode that generated it, and (b) shows the answer of each mode to the same prompt together with the decoding speed and the energy per token of each mode, measured from the energy counters of the GPU (NVML) and the CPU package (RAPL); the rest of the PC is not measured. For Qwen3-30B-A3B, the experts stay on the SSD and only those in use are kept in a VRAM cache (4\,GB); raising the mode reads only the planes that a cached expert lacks.

\end{document}